\PassOptionsToPackage{warn}{textcomp}

\documentclass [11pt, proquest] {uwthesis}[2018/07/01]
 
\usepackage{alltt}  %

\usepackage{times}
\usepackage{latexsym}
\usepackage{amssymb}
\usepackage{graphicx}
\usepackage{multirow}
\usepackage{textcomp}
\usepackage{array}
\usepackage{url}
\usepackage{xcolor}
\usepackage{comment}
\usepackage[caption=false]{subfig}
\usepackage{enumitem}
\usepackage{booktabs}
\usepackage{amsmath}
\usepackage{rotating}
\usepackage{setspace}
\usepackage{url}
\usepackage{algorithm,algorithmic}
\usepackage{cite}
\usepackage{hyperref}
\usepackage{cleveref}
\usepackage{comment}
\usepackage[htt]{hyphenat}
\usepackage[normalem]{ulem} 
\usepackage{enumitem}
\setlist{nolistsep}
\usepackage{makecell}
\usepackage{bm}

\usepackage{todonotes}
\newcounter{hfang}
\newcommand{\hfang}[1]{%
\refstepcounter{hfang}%
{%
\todo[color=orange, size=\footnotesize]{%
[\textbf{hfang:\thehfang}] #1}%
}}%
\newcounter{done}
\newcommand{\tabitem}{~\llap{\textbullet}~}
\newcommand{\sos}{\scalebox{0.85}{\ensuremath{<\!\!{\rm s}\!\!>}}}
\newcommand{\sep}{\scalebox{0.85}{\ensuremath{<\!\!{\rm sep}\!\!>}}}

\begin{document}
 
%
%

\prelimpages
 
%
%
\Title{Building A User-Centric and Content-Driven Socialbot}
\Author{Hao Fang}
\Year{2019}
\Program{Electrical and Computer Engineering}

\Chair{Mari Ostendorf}{Prof.}{Department of Electrical and Computer Engineering}
\Signature{Yejin Choi}
\Signature{Hannaneh Hajishirzi}
\Signature{Geoffrey Zweig}

\copyrightpage

\titlepage

%
%

%
%

\setcounter{page}{-1}
\abstract{%
Researchers in artificial intelligence (AI) have long been interested in the
challenge of developing a system that can have a coherent conversation with
humans.
The Loebner Prize is a form of Turing test that has challenged researchers since
1990.
More recently, several commercial products of conversational AI have emerged in
the market,
such as Amazon Alexa, Microsoft Cortana, Google Assistant, and Apple Siri.
Research in both task-oriented systems that aim at accomplishing a well-defined
task and chatbots that engage users in chit-chat interactions have made
considerable advancement.
The recent Alexa Prize sets forth a new challenge: creating a {\bf socialbot}
that can hold a coherent and engaging conversation on current events and popular
topics such as sports, politics, entertainment, fashion, and technology.

This thesis was born out of the Alexa Prize. 
Our socialbot, Sounding Board, demonstrated that it is feasible to build a
system that can engage in long conversations when backed by rich content crawled
from the web and knowledge of the user obtained through interaction.
This user-centric and content-driven design helped Sounding Board win the
inaugural Alexa Prize with an average score of 3.17 on a 5-point
scale and an average conversation duration of 10:22, evaluated by a panel of
independent judges.


To build Sounding Board, we develop a system architecture that is capable of
accommodating dialog strategies that we designed for socialbot conversations.
The architecture consists of a multi-dimensional language understanding module
for analyzing user utterances, a hierarchical dialog management framework for
dialog context tracking and complex dialog control, and a language generation
process that realizes the response plan and makes adjustments for speech
synthesis.
Additionally, we construct a new knowledge base to power the socialbot by
collecting social chat content from a variety of sources.
An important contribution of the system is the synergy between the
knowledge base and the dialog management, i.e., the use of a graph structure to
organize the knowledge base that makes dialog control very efficient in bringing
related content to the discussion.

The Alexa Prize offers a new and unique platform for researchers to build and
test socialbots by allowing the systems to interact with millions of real users
through Alexa-enabled devices.
Using the data collected from Sounding Board during the competition, we carry
out in-depth analyses of socialbot conversations and user ratings which provide
valuable insights in evaluation methods for socialbots.
We additionally investigate a new approach for system evaluation and diagnosis
that allows scoring individual dialog segments in the conversation.

Finally, observing that socialbots suffer from the issue of shallow
conversations about topics associated with unstructured data, we study the
problem of enabling extended socialbot conversations grounded on a document.
To bring together machine reading and dialog control techniques, a graph-based
document representation is proposed, together with methods for automatically
constructing the graph.
Using the graph-based representation, dialog control can be carried out by
retrieving nodes or moving along edges in the graph.
To illustrate the usage, a mixed-initiative dialog strategy is designed for
socialbot conversations on news articles.
A new socialbot prototype is developed for user studies demonstrating the
benefits of the proposed document representation and dialog strategies.
}

%
%
\tableofcontents
\listoffigures
\listoftables  
 
%
%

%
%
\acknowledgments{
	First and foremost, I would like to express my sincere gratitude to my advisor
	Mari Ostendorf.
	During my PhD studies, I have learned a lot from her insightful advice on
	improving my research, writings, and public speaking skills.
	Moreover, she has always been supportive, letting me pursue my interest
	in different directions including the Alexa Prize.

	I would also like to thank my reading committee members, Yejin Choi, Hannaneh
	Hajishirzi, and Geoffrey Zweig, for your time, your guidance, and help in
	improving my thesis.
	Thanks to Leah M. Ceccarelli for serving the graduate school representative
	and Eve Riskin for gracefully agreeing to sit on my committee at the last
	minute. 

	Many thanks to Sounding Board team members, Hao Cheng, Elizabeth Clark, Ari
	Holtzman, and Maarten Sap, for all the hard-working days we had together on
	developing the bot and our great memories of the award-winning moment. 
	Also thanks to both Yejin Choi and Noah A. Smith for advising the team
	together with Mari.

	I have also been fortunate to be part of a vibrant and supportive research lab.
	Thank you to all TIAL lab members and alumni: Amittai Axelrod, Sangyun Hahn,
	Ji He, Jingyong Hou, Brian Hutchinson, Aaron Jaech, Yuzong Liu, Roy Lu, Yi
	Luan, Kevin Lybarger, Alex Marin, Julie Medero, Farah Nadeem, Nicole Nichols,
	Sining Sun, Trang Tran, Ellen Wu, Victoria Zayats.
	Thanks to our lab's system administrator, Lee Damon, who has maintained a
	stable filesystem and network for the lab.
	Special thanks to my best friend, Hao Cheng, who spent many long days and
	sleepless nights working on papers and experiments with me and cheered me up
	when I was depressed and anxious.

	I have had several great internships at Microsoft Research.
	Many thanks to Geoffrey Zweig who hosted my first internship and kindly served
	my reading committee.
	Also thanks to my other mentors and collaborators I met at Microsoft Research:
	Yoram Bachrach,
	David Carter,
	Xinlei Chen, 
	Li Deng,
	Piotr Doll\'{a}r,
	Jianfeng Gao,
	Saurabh Gupta, 
	Xiaodong He, 
	Forrest Iandola, 
	Tsung-Yi Lin, 
	Margaret Mitchell,
	John Platt,
	Rupesh K. Srivastava, 
	Ryota Tomioka, 
	Ramakrishna Vedantam,
	Jason D. Williams,
	and C. Lawrence Zitnick.
	I would also like to express thanks to my collaborators outside the University
	of Washington for their help and feedback during my early PhD studies: Peter
	Baumann, Eric Fosler-Lussier, Yanzhang He, Janet Pierrehumbert, and
	collaborators from the International Computer Science Institute.
	 
	My thesis was born out of the Amazon Alexa Prize.
	I would like to express my sincere gratitude to Alexa Prize organizers.
	This work could not have been done without their support.
	Also thanks to the Amazon Alexa Fellowship which funded me after the
	competition.

	Finally, I am grateful beyond words for the support of my parents: Hong Fang
	and Xiaoling Wu.
	It is their love and understanding that makes it possible for me to start and
	complete my PhD journey.
}

%
%

%
 
%
%

\textpages
 
\chapter{Introduction}
People have long been interested in having conversational interactions with
artificial intelligence (AI) systems.
Tremendous progress has been made in the area of conversational AI, including
research in speech technologies, natural language processing, information
extraction, machine learning, etc.
Now, there are a variety of intelligent personal assistants that have
participated in our daily life, such as Amazon Alexa, Apple Siri, Google
Assistant, and Microsoft Cortana.

As an early and influential conversational AI system, ELIZA
\cite{Weizenbaum1966Comm} responds to human users roughly as a psychotherapist
who encourages the human to talk as one would to a psychiatrist.
The conversations for this specific type of psychiatric interview can assume
the system playing the role of the psychiatrist knows almost nothing about the
real world.
Commonly called {\bf chatbots}, such systems attempted to simulate human
behaviors in conversations, and therefore, were evaluated in various forms of
the Turing Test \cite{Turing1950Mind}.
For example, PARRY \cite{Colby1971Parry}, which was developed a few years after ELIZA,
focused on clinical psychology and was used to simulate paranoia, defined as a
mode of thinking characterized by ``malevolence delusions.''
It passed the Turing test in that psychiatrists could not distinguish transcripts
of interviews with PARRY from transcripts of interviews with real paranoids
\cite{Colby1972Turing}.
A more recent influential system, ALICE \cite{Wallace2009ALICE}, is known for
winning the Loebner Prize \cite{LeobnerPrize} multiple times as ``the most human
computer.''
As AI systems become more accessible to the general public, there is growing
interest in using chatbots for entertainment and companionship.
For example, the recent Microsoft XiaoIce is a commercial product for carrying
out entertaining social chit-chat with human users.
It has broad exposure in the Asian market with over 100 million users and
billions of conversation turns \cite{Shum2018arXiv}.
With the recent success of deep learning, a few neural network models have
been proposed for chit-chat response
generation~\cite{Vinyals2015ICML,Shang2015ACL,Sordoni2015NAACL,Li2016EMNLP,Serban2017AAAI}.

Another type of conversational AI systems, i.e., {\bf task-oriented dialog
systems}, puts more emphasis on accomplishing specific tasks.
One of the earliest commercial task-oriented systems is the AT\&T How May I Help
You (HMIHY) system deployed for routing customers calls to the correct
destination \cite{Gorin1997SpeechComm}.
Considerable advancement has been made over the past two decades, notably, with
DARPA's initiatives on the Airline Travel Information System (ATIS)
\cite{Price1990ATIS} and the Communicator project \cite{Walker2002ICSLP}.
Many research prototypes are developed for a single task or a limited set of
tasks with access to a relatively small structured database that contains all
information about the tasks.
Recently, there is great interest in moving towards open-domain systems
that can converse about anything in a large knowledge
base~\cite{Tur2016Interspeech,Mrksic2015ACL,Gasic2015ASRU,Wen2015NAACL}, as well
as systems that can use unstructured data~\cite{Lowe2015SIGDIAL}.

In this thesis, we study a new type of conversational AI, a {\bf socialbot},
as represented in the systems designed for
the Amazon Alexa Prize \cite{AlexaPrize2017}.
The Alexa Prize socialbots are asked to converse with human users about popular
topics and recent news through voice-based interactions.
The conversation is configured to be initiated by the invocation
utterance ``{\it let's chat}'' or its variants (e.g., ``{\it talk to me}'',
``{\it I want to have a conversation}'').
Sometimes, the user may have a topic in mind (e.g., ``{\it let's chat about
sports}''), but in many cases, the user does not have a specific goal initially.
Nevertheless, user goals may emerge as the conversation evolves.
This makes the socialbot significantly different from task-oriented systems
which usually have a specific and well-defined task to accomplish.
On the other hand, the socialbot is not the same as the chit-chat systems which
mostly respond to users via simulating human behaviors and let the user lead
the conversation.
In order to engage users in chatting about popular topics or recent news, the
socialbot needs to strategically push the conversation forward
while handling all kinds of user requests and casual chit-chat interactions.
Therefore, the socialbot brings many new challenges to the area of conversational AI.

The nature of socialbot conversations greatly impacts the technologies and
approaches for building a socialbot that differs from existing task-oriented and
chit-chat systems.
First, the socialbot needs to develop dialog strategies that bring together a
variety of skills, including chit chat, question answering, information seeking,
topic recommendation, handling speech recognition errors, etc.
Furthermore, a large knowledge base is needed to cover the wide variety of
topics the user may be interested in.
Additionally, the socialbot has to explore unstructured data such as news
articles and online discussions to gain knowledge of recent events and
interesting content that can be used in the conversation.
Lastly, the socialbot presents a problem for evaluating dialog success since
there is no clear definition for task completion or engagement. 

The remaining of the chapter is organized as follows.
We first discuss the scenario motivated our dialog strategies in
\S\ref{chap1:sec:gateway}.
The general approach and contributions are presented in
\S\ref{chap1:sec:contributions}.
Finally, we provide an overview of the dissertation in
\S\ref{chap1:sec:overview}.

\section{The Socialbot as a Conversational Gateway to Online Content}
\label{chap1:sec:gateway}
Socialbots can have many applications such as entertainment, education,
healthcare, companionship, assisting visually impaired users, etc.
Besides these applications, the work in this thesis is also motivated by our
vision of using a socialbot as a conversational gateway to online content.
In other words, the socialbot can act as a voice-enabled conversational user
interface between the user and the large amount of online content.

The explosive growth and variety of information online can be overwhelming.
This information overload issue appears not only in e-business services such
as buying products, but also in discovering social media news and discussions.
Classical systems such as databases and search engines alleviate the problem by
allowing users to use descriptive queries for information retrieval.
In recent years, with the capability of effectively suggesting information to
users, recommender systems have played an important role in popular websites
including Amazon, YouTube, Netflix, and TripAdvisor.
Research efforts are made to leverage context and user information to improve
the recommender systems.
Nevertheless, the information acquisition process in most existing information
retrieval and recommender systems rarely preserve the interactive nature of 
the way humans exchange information with each other.
Moreover, most systems are focused on text-based interactions rather than
voice-based interactions.

As the conversational interface becomes more mature and prevalent in our life,
we envision that the socialbot can serve as a conversational gateway to online
content for users to learn about and discuss information tailored to them.
The socialbot can establish a channel between the user and content contributors,
augmenting existing information acquisition methods such as using search engines
(e.g., Google, Microsoft Bing) and browsing social media platforms (e.g.,
Twitter, Reddit).
It intelligently routes content to users by understanding their information
request, assessing their interests in particular topics, and bringing relevant
information from one article while discussing another.
A socialbot like this can become an information acquisition tool in our daily
life.

A demanding type of conversation activity for socialbots is interactive news
discussion.
A recent study on news consumption behaviors observes a trend that most people
appear to share an article online without ever reading
it~\cite{Gabielkov2016SIGMETRICS}.
It is not totally surprising since it takes considerably less time and effort to
share an article than it does to actually read it.
While there are some benefits from the headline-based social exchange of
content, it may discourage readers from being well informed and publishers from
writing in-depth articles.
With a socialbot that can chat about news in an engaging and efficient way,
users may be more willing to learn about a news story, which can benefit both
readers and publishers.

\section{General Approach and Contributions}
\label{chap1:sec:contributions}
The work presented in this thesis is centered around the Alexa Prize socialbot.
Historically, chatbots in the literature have adopted approaches different from
task-oriented systems including both system frameworks and evaluation methods,
perhaps because of the differences in the user's goal and application scenarios.
For both types of systems, there are several existing system frameworks and
datasets that facilitate the research.
However, since the Alexa Prize socialbot is a new type of conversational AI,
there is no available system framework or public dataset that can be easily
used.

In order to study socialbot conversations, a socialbot system is needed in the
first place.
In this thesis, we develop a socialbot using user-centric and content-driven
dialog strategies.
The user-centric strategy allows users to control the topic of conversation
and requires the bot to adapt its responses based on their reactions, topic
interests, and personality.
The content-driven strategy pushes the conversation forward by supplying
interesting and relevant information.
A collection of conversation activities are designed for the socialbot.

The socialbot architecture developed in this thesis uses a novel hierarchical
dialog management framework and a new knowledge base with a rich content
collection.
The proposed hierarchical dialog management framework allows efficient dialog
control of a wide variety of conversation activities and maintains smooth
transitions between them.
It decouples modules for efficient system development and maintenance.
The knowledge base consists of social chat content that is updated daily.
In particular, we represent the knowledge base as a graph consisting of content
nodes and topic nodes.
Edges between a content node and a topic node are labeled by their relations.
Several types of edges are designed based on available information that comes
with the content and that can be extracted using natural language processing
tools.
This graph structure facilitates controlling the dialog flow and bringing new
information to the conversation.

One of the core research problems on dialog systems is system evaluation,
which requires understanding aspects that influence user satisfaction and the
underlying structure of the conversation.
Using the unique data collected from our socialbot during the Alexa Prize
competition, we carry out in-depth analysis on socialbot conversations by
studying correlations between user ratings and a set of conversation acts
proposed for characterizing user and bot turns.
Moreover, a hierarchical dialog structure model is developed for socialbot
conversations that facilitates dividing a conversation into coherent segments.
The segmentation makes it possible to assess quality of individual segments in a
conversation which helps system diagnosis, since a conversation can involve both
good and bad segments.
We further investigate two multi-level scoring models that automatically predict
both conversation-level and segment-level scores, using two different hypotheses
for modeling the scoring process. 
Using the resulting segment scores, we provide case studies to illustrate how
segment scores can be used to diagnose dialog policies and content sources.

Finally, to address the challenge of discussing topics associated with
unstructured data, we propose a graph-structured document representation that
facilitates dialog controls in document-grounded conversations.
Using relation information encoded in the graph, a mixed-initiative dialog
strategy is designed for extended conversations on news articles based on moves
and searches on the graph.
Methods for automatically constructing the graph are developed.
A new socialbot prototype is implemented to carry out user studies 
demonstrating the benefits of the document representation and dialog strategies.

This thesis impacts future research on both socialbots and task-oriented
systems.
Being one of the pioneers in socialbot research, the development of the
socialbot sets a foundation for further studies in this area.
The findings from our in-depth analysis on socialbot conversations provide
valuable insights for future socialbot development and evaluation.
Our work on enabling conversations grounded on news articles greatly
improves the socialbot's capability of discussing the latest events and enlarge its
knowledge base.
While the hierarchical dialog management framework, the multi-level evaluation
methods, and the graph-structured knowledge representation are developed for
socialbots in this thesis, they all can be applied to task-oriented systems.
In particular, for task-oriented systems that involve multiple tasks and
interact with a large-scale knowledge base, the dialog control can be very
complex.
The hierarchical dialog management framework and the use of a graph structure to
organize the knowledge base can accommodate complex dialog control for such
systems.
The multi-level evaluation approaches can also be useful for segmenting a
conversation involving multiple subtasks and identifying the subtasks that are
handled less successfully by the system.

\section{Dissertation Overview}
\label{chap1:sec:overview}
In Chapter~\ref{chap2}, we provide a general overview of conversational AI
research.
Recent work on socialbots are discussed from three aspects: socialbot
architecture, social chat knowledge management, and socialbot evaluation.
We also discuss limitations of existing work on socialbots, much of which was
developed in parallel with this thesis.

Chapter~\ref{chap3} presents the Sounding Board system, which won the inaugural
Alexa Prize in 2017.
The dialog strategies designed for Sounding Board are discussed.
We also describe the complete system architecture and individual components,
including multi-dimensional language understanding, hierarchical dialog
management, language generation, and the social chat knowledge graph.

In-depth analyses on socialbot conversations are carried out in
Chapter~\ref{chap4}.
A hierarchical dialog structure model is developed for segmenting the
conversation.
Two multi-level scoring methods are investigated for automatically predicting
scores at both conversation and segment levels.
We further illustrate the use of segment scores for system diagnosis.

In Chapter~\ref{chap5}, we present a novel approach for the socialbot to carry
out conversations grounded on unstructured documents.
The graph-based document representation and its construction methods are
described.
Dialog strategies are designed based on moves and search on the graph.
Experiments are carried out to evaluate the performance of the construction
methods for the document representation and the dialog strategies.

Finally, Chapter~\ref{chap6} conclude the thesis with a summary of approaches
and models developed in this thesis, as well as the impacts of this thesis.
We also discuss future directions that can advance research on socialbots and
the field of conversational AI.

\chapter{Background}
\label{chap2}
While there are many types of conversational AI systems in the literature and
commercial markets, the focus of this thesis is on open-domain socialbots.
Since it is a new application of conversation AI, system development and
research studies are mostly carried out in the past two years.
In this chapter, we first review types of conversational AI systems in
\S\ref{chap2:sec:conv_ai}.
Then we describe the Alexa Prize challenge in \S\ref{chap2:sec:competition}.
It is this competition that initiates and supports research on socialbots.
In \S\ref{chap2:sec:architecture}, we review socialbot architectures and common
techniques used in existing socialbot systems, which were developed in parallel
with or after our system.
Approaches used for acquiring social chat knowledge are discussed in
\S\ref{chap2:sec:knowledge}.
We review studies on socialbot evaluations in \S\ref{chap2:sec:evaluation}.
Finally, we conclude this chapter with discussion of the limitations of existing
work in \S\ref{chap2:sec:limitation}.

\section{Types of Conversational AI}
\label{chap2:sec:conv_ai}
In the literature, the two main types of conversational AI systems are task-oriented
(sometimes called goal-oriented) and non-task-oriented (sometimes called
non-goal-oriented) systems.
{\bf Task-oriented systems} primarily interact with users to accomplish
specific tasks that range from simple and well-defined tasks (e.g., hotel or
flight booking, restaurant reservation) to complex tasks involving sophisticated
planning and reasoning (e.g., holiday planning, contract negotiation).
In contrast, {\bf non-task-oriented systems} usually engage users in
conversations that do not necessarily involve a task to be accomplished.
Such systems have been used for entertainment and companionship and are usually
referred to as {\bf chatbots}.
Naturally, a conversation can involve a mixture of task-oriented and
non-task-oriented interactions with smooth transitions.
Thus, in practice, there may be some overlap between these two types of systems.

Another common way to categorize conversational AI systems is based on the
dialog
initiative design, i.e., who has the control of the conversation.
Early task-oriented systems (e.g., the AT\&T How May I Help You
system~\cite{Gorin1997SpeechComm}) mostly involve {\bf system initiative}.
The conversation resembles a form-filling process and is completely controlled
by the system.
The user simply responds to provide requested information.
In this design, the system needs to manage how the conversation should move
forward, but the management is relatively easy since the user actions are
constrained and only allowed at certain points in the dialog.
{\bf User-initiative} systems (e.g., \cite{Tomko2004ICSLP},
\cite{Leuski2006SIGDIAL}), on the other hand, rely on the user to control the
conversation flow.
In this case, the user directs the system to provide information or complete a
task.
The system passively responds to user requests but does not proactively push the
conversation forward.
For these systems, dialog control is mainly related to task clarification and/or
error correction.
Nevertheless, it poses challenges to language understanding to cover a wide
variety of user utterances.
Conversational AI systems using a {\bf mixed-initiative} design (e.g., the AT\&T
Communicator system~\cite{Narayanan2000ICSLP}) share the dialog control with the
user, i.e., the initiative shifts back and forth between the system and user.
This type of interaction is usually viewed as a more natural way of
communication, but it brings challenges to both language understanding and
dialog control.

A third way to categorize conversation AI systems is based on the coverage of
domains.
For task-oriented dialog systems, the domain is essentially equivalent to the
task (e.g., restaurant search, flight booking).
They typically have access to a database containing all available information
about the task.
The domain can thus be described by the underlying ontology, which is
essentially a structured representation of the database.
Recently, there is great interest in moving from {\bf single-domain}
systems that operate on a small ontology to {\bf multi-domain} or 
{\bf open-domain} dialog systems that can converse about anything in a large
knowledge
base~\cite{Tur2016Interspeech,Mrksic2015ACL,Gasic2015ASRU,Wen2015NAACL}.
For chatbots, the domain usually means the topic of the conversation (e.g.,
sports, movies) 
or is reflected in the dataset that they are trained on (e.g., the Ubuntu dialog
corpus~\cite{Lowe2015SIGDIAL}, the Twitter corpus~\cite{Ritter2010NAACL}).
Due to the nature of chit-chat conversations, such chatbots are usually viewed
as open-domain systems.
However, most existing systems either do not rely on web-scale knowledge (e.g.,
ALICE~\cite{Wallace2009ALICE}) or use off-the-shelf APIs for question
answering which may have access to an external knowledge base (e.g.,
\cite{Higashinaka2014COLING}). 

Along the line of research on task-oriented systems, one of the earliest
commercial task-oriented systems is the AT\&T How May I Help You system deployed
for routing customers calls to the correct
destination~\cite{Gorin1997SpeechComm}, which is a single-domain
system-initiative system.
DARPA's efforts on the Airline Travel Information System~\cite{Price1990ATIS}
and the Communicator project~\cite{Walker2002ICSLP} has contributed to the
advance of this research area by moving towards somewhat more complex domains
and mixed-initiative dialog design.
A series of research has been carried out on spoken language understanding (see
reviews in~\cite{Wang2005IEEE,Mori2008IEEE}).
Efforts have also been made on developing statistical frameworks for dialog
management.
The Dialog State Tracking Challenge (DSTC) has provided a common test bed and
evaluation suite~\cite{DSTC123,DSTC4,DSTC5}.
The partially observable Markov decision process (POMDP)
framework~\cite{WilliamsPhDThesis,Young2013IEEE} has been effective in
several systems such travel planning, appointment scheduling, car navigation,
etc.
Most recent work has proposed end-to-end neural networks for task-oriented
systems~\cite{Wen2017EACL,Bordes2017ICLR,Zhao2017SIGDIAL,Li2017IJCNLP,BingLiuPhDThesis}.
For these neural dialog systems, the notions of dialog initiative and domain are
mostly defined by the data used to train the neural networks.

%

Research on chatbots, on the other hand, is mostly conducted in parallel to the
task-oriented systems.
Historically, task-oriented systems are primarily spoken dialog systems, and
chatbots are dominated by text-based systems including the three influential
systems ELIZA \cite{Weizenbaum1966Comm}, PARRY \cite{Colby1971Parry}, and ALICE
\cite{Wallace2009ALICE}.
These chatbots mostly originated from the idea of simulating human
conversations and passing the Turing test \cite{Turing1950Mind}.
Chatbots are usually mixed-initiative and the user and the bot switch roles
between a primary speaker and an active listener.
The notion of domain for chatbots is loosely defined since there is no
well-defined task.
In the literature, many chatbots are viewed as open-domain though they generally
cannot handle discussing evolving current events.
Similar to task-oriented systems, statistical response generation models have
also been proposed for chatbots.
In~\cite{Ritter2011EMNLP}, the problem is formulated as a statistical machine
translation task where a two-turn conversation is viewed as translating from the
first turn to the second turn.
Recent work on response generation mostly views it as a sequence-to-sequence
generation problem that can be addressed using neural conversation
models~\cite{Vinyals2015ICML,Shang2015ACL,Sordoni2015NAACL,Li2016EMNLP,Serban2017AAAI}.
In addition to generating the response, these neural models are also used for
retrieving responses from a corpus of human conversations, and thus, they are
sometimes referred to as information-retrieval-based or response-retrieval-based
chatbots.
With a similar motivation as the DSTC, the recent Conversational Intelligence
Challenge (ConvAI) aims to establish a concrete scenario and a standard
evaluation for testing chatbots that engage users on
chit-chat~\cite{ConvAI,ConvAI2017}.

The Alexa Prize {\bf socialbot} is a novel application of conversational
AI~\cite{AlexaPrize2017}.
As acknowledged by other Alexa Prize teams, the goal of discussing popular
topics and recent events makes it significantly different from traditional
task-oriented systems and chatbots that focus on chit-chat.
Despite the interchangeable use of ``task-oriented'' and ``goal-oriented'' in
the literature, we think Alexa Prize socialbots are non-task-oriented but they
are still goal-oriented.
On the one hand, they do not have a specific and well-defined task to carry out
as task-oriented systems.
Nevertheless, socialbots need to elicit the user's goal (e.g., topic interests)
and establish their own goals (e.g., discussing a movie) for the conversation.
On the other hand, they are different from chit-chat systems designed for
passing the Turing test.
Actually, socialbot conversations involve a lot of commands (e.g., 
``{\it continue}'', ``{\it help}'', ``{\it switch topic}'') which users use to
control the conversation flow.

The nature of the socialbot conversations greatly impacts the technologies
needed for developing socialbot systems.
In this thesis, we aim to develop a socialbot that is mixed-initiative and
open-domain.
In most cases, the socialbot needs to proactively lead the conversation in order
to engage the user, but the user has the freedom to interrupt the conversation
flow and switch topics at any time.
The socialbot also needs to operate on a large scale of knowledge base to cover
a wide range of topics the user might be interested in talking about.
A critical issue raised by this is the need to build a social chat knowledge
base, which requires the system to acquire social chat content from unstructured
data and merge knowledge from different sources that are updated frequently.

\section{The Alexa Prize Challenge}
\label{chap2:sec:competition}
With the goal of advancing research in conversational AI, Amazon announced
the Alexa Prize on September 26, 2016 and set forth a new challenge for the
research community: creating a socialbot system that can hold a coherent
and engaging conversation on current events and popular topics such as sports,
politics, entertainment, fashion and technology \cite{AlexaPrize2017}.
The grand prize challenge is to engage a user in such a conversation for 20
minutes.

{\bf Competition Setup}:
With the infrastructure supported by Alexa Prize organizers, university teams
participating in the competition can release their socialbots to a large user
base consisting of millions of Alexa users.
The conversation is always initiated by the user with the invocation phrase 
{\it ``let's chat''} or a few variants (e.g., ``{\it let's talk}'', 
``{\it talk to me}''),
and the user is routed to a randomly selected competing socialbot system. 
The user and the socialbot then take turns to carry out the conversation.
Specifically, the socialbot needs to wait for a user turn before proceeding to
its next turn.
The user can terminate the conversation at any time using an explicit 
{\it ``stop''} command.
Afterwards, the user can optionally rate the system by responding to the
question: 
{\it ``On a scale of one to five, how much would you like to speak to this
socialbot again?''}
The teams can access the optional user ratings of their own socialbots for
system development.
To reduce bias, the competing socialbots are not allowed to reveal the team's
identity during the conversation.

{\bf Finals Evaluation}:
In the 2017 Alexa Prize competition, three finalist socialbot systems
(Alana \cite{HWUAlana2017}, Alquist~\cite{CTUAlquist2017}, Sounding
Board~\cite{UWSoundingBoard2017}) were evaluated by a panel of three independent
judges.
The judges listened to the live conversation carried out by an human interactor
and a finalist socialbot, and the conversation was rated by each judge on a
scale of 1 to 5 (higher is better).
The interactors were instructed to keep the conversation going until two of
three judges have pressed the stop button.
Each finalist socialbot carried out two conversations with the same interactor,
and there were in total three interactors participating the finals evaluation. 
Sounding Board won the finals with an average score of 3.17 and an average
conversation duration of 10:22.
The 2018 Alexa Prize\footnote{Sounding Board did not participate in the 2018 Alexa
Prize.} finals were carried out in a similar setting but with more judges and
interactors.
Among three finalist socialbots (Alana~\cite{HWUAlana2018},
Alquist~\cite{CTUAlquist2018}, Gunrock~\cite{UCDavisGunrock2018}), 
Gunrock won with an average score of 3.1 and an average duration of 9:59.

{\bf Challenges and Opportunities}:
Because of the large and dynamic space of conversation topics, there was no
prior dialog system that could easily be used for this socialbot task, nor were
there corpora available for data-driven analysis and model development.
Therefore, the socialbot brings new challenges and research opportunities to the
conversational AI community.
We categorize them into three aspects:
system architecture, social chat knowledge management, and socialbot evaluation.

In terms of the architecture, the system framework needs to accommodate the
specific nature of socialbot conversations, which are voice-based,
mixed-initiative, and open-domain.
The voice-based interaction nature requires the system to take
into account issues that are not present in text-based chatbots, including
speech recognition errors, lack of sentence segmentation, disfluencies, and
language generation considerations for speech synthesis.
Meanwhile, the lack of conversation corpora makes it difficult to adopt
purely statistical frameworks such as the POMDP framework
and end-to-end neural networks.
While there are several existing dialog datasets and some of them are
open-domain conversations on casual topics (see a comprehensive survey in
\cite{Serban2018DD}), they hardly cover the full spectrum of topics and implicit
goals in socialbot interactions.
For example, the Switchboard dataset~\cite{SwitchboardCorpus} has human-human
phone conversations on specific topics, but commands are rarely observed in this
corpus.

From the perspective of the social chat knowledge management, the system
needs to collect and dynamically update its knowledge of interesting content
and recent events to carry out engaging conversations with users.
There are two open questions: 1) what types of knowledge sources are useful for
socialbot conversations, and 2) how can the knowledge be acquired at scale
and organized efficiently for socialbots to use in the conversation.
For the first question, while there are a variety of knowledge bases such as
Freebase~\cite{Freebase2008} and Wikidata~\cite{DBpedia2015} that provide
factual knowledge (e.g., the birthday of a celebrity), they are not
sufficient for socialbot conversations as they lack social chat content
(e.g., interesting facts, amusing thoughts).
For the second question, the two popular approaches used for
chatbot systems, i.e., scripting responses with template-matching rules (e.g.,
ELIZA~\cite{Weizenbaum1966Comm}, ALICE~\cite{Wallace2009ALICE}) and retrieving
responses from a large scale conversation repository (e.g.,
\cite{Jafarpour2009NIPS,Wang2013EMNLP,Sordoni2015NAACL,Yan2016ACL}), are not
by themselves adequate to efficiently acquire and use social chat knowledge
about current events.
It is not practical to use the response scripting approach at scale by manually
searching for social chat knowledge and incorporating them in the responses.
Meanwhile, the response retrieval approach requires a way to enrich the
response repository or edit retrieved responses to inject new social chat
knowledge.

General methodology for evaluating dialog systems is still an
active research area.
A comprehensive review can be found in \cite[Chapter~6]{Jokinen2010SDS}.
In this thesis, we focus on methods that can be applied to assessing the
performance of socialbots, which had not been studied before the Alexa Prize.
In the past, dialog systems are tested either in the laboratory or in real usage
conditions at a relatively small scale.
While real situations are generally regarded as providing the best
conditions for collecting data, in practice only near-final
products in companies can proceed to extensive testing with real users.
The Alexa Prize setup provides a unique test bed for studying socialbot
conversations by allowing the socialbots to interact with millions of Alexa
users and receive their subjective ratings on the conversations.
Nevertheless, it remains an open question as to how this large scale real
user conversation data can be exploited in socialbot development, considering
that the user ratings vary greatly from person to person and cannot be applied
uniformly to all segments of a conversation.

\section{Socialbot Architectures}
\label{chap2:sec:architecture}
\begin{figure}[!t]
	\centering
	\includegraphics[width=0.9\textwidth,trim=0cm 0cm 0cm 0cm,clip]{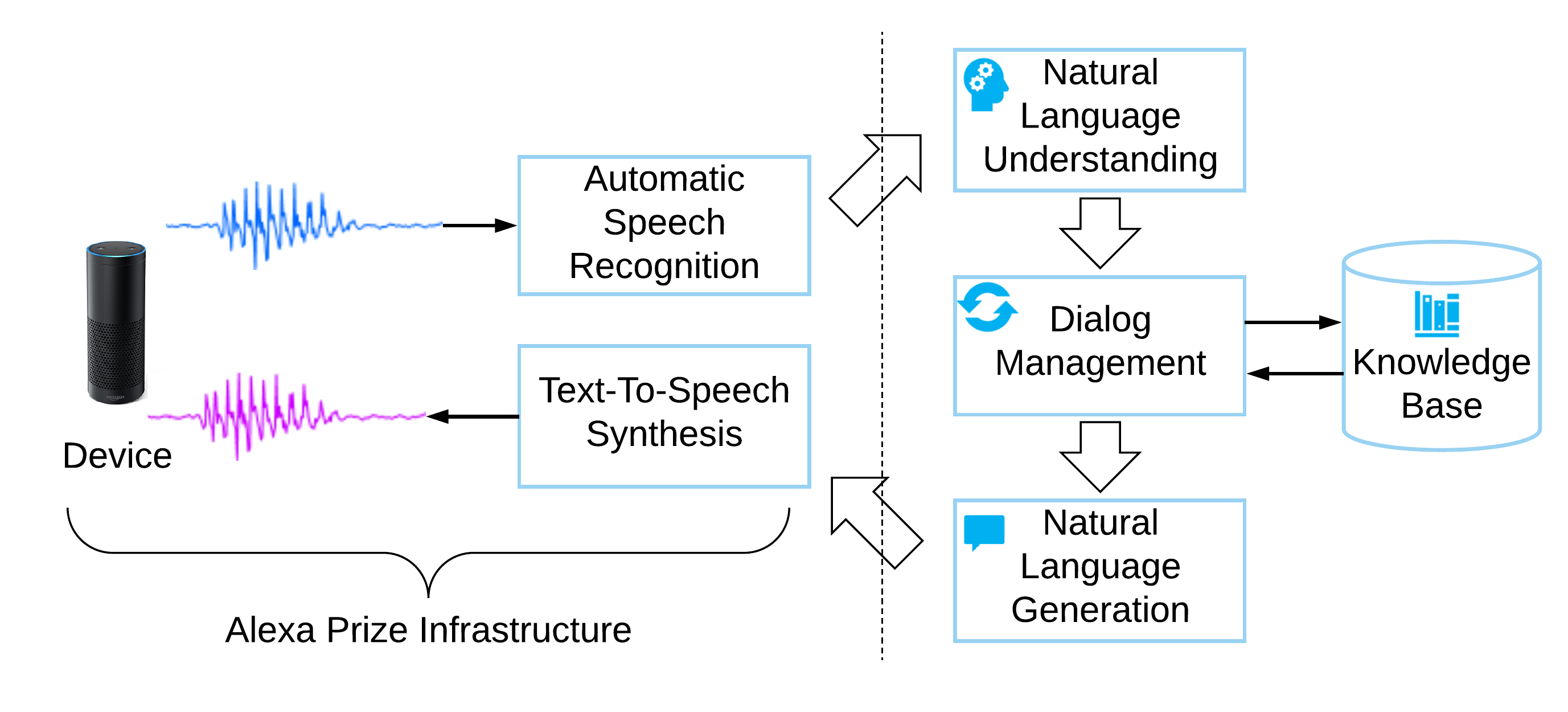}
	\caption{A generic architecture for socialbots.}
	\label{chap2:fig:generic_architecture}
\end{figure}

In the 2017 and 2018 Alexa Prize competitions, university teams have explored
a variety of system architectures for socialbots.
While these architectures vary from each other, most systems follow the generic
architecture as illustrated in Fig.\,\ref{chap2:fig:generic_architecture}, which
is also a widely adopted architecture for task-oriented dialog
systems.
This pipeline architecture include six major blocks:
automatic speech recognition (ASR), natural language understanding (NLU), dialog
management (DM), natural language generation (NLG), text-to-speech (TTS)
synthesis, and a knowledge base (KB).
The workflow through these blocks can be described by the following 5 steps:
\begin{itemize}
	\item The ASR block transcribes the user utterance from audio signals to
		textual hypotheses.
	\item The hypotheses are sent to the NLU block to carry out a series of
		language analyses, creating a semantic representation of the user utterance.
	\item The DM block carries out the dialog control logic using the semantic
		representation from the NLU block and the context information stored in the
		DM block.
		A response plan, i.e., a semantic representation of the response, is
		assembled by the DM block.
		The DM block also communicates with the KB to obtain content for the
		response.
	\item The NLG block realizes the response plan, converting the semantic
		representation to the natural language in the textual form.
	\item The TTS block converts the textual bot response into audio signals.
\end{itemize}

In practice, each block can consist of a collection of modules.
There are many variations on how individual blocks are configured and
implemented, as well as the types of data communicated between them.
Further, there is no clear cut distinction between blocks and modules and not
all socialbots follow this specific organization.
Since the Alexa Prize infrastructure provides the ASR and TTS support,
university teams are thus focused on intermediate steps for managing the
conversation context and generating appropriate responses.

In this section, we first briefly describe the speech processing in the Alexa
Prize infrastructure in \S\ref{chap2:ssec:speech_processing}.
Then we discuss representative components and techniques used by socialbots for
the NLU, DM, and NLG blocks in \S\ref{chap2:ssec:nlu}--\ref{chap2:ssec:nlg},
respectively.

\subsection{Speech processing}
\label{chap2:ssec:speech_processing}
The underlying algorithms for ASR and TTS used by the Alexa Prize infrastructure
is not revealed.
Here, we describe the Alexa Prize infrastructure setup in terms of the
speech-language interface.
For a comprehensive description of ASR and TTS, please refer to the
book~\cite{Jurafsky2009slp}.

{\bf ASR}:
Raw audio signals of user speech are not accessible to Alexa Prize socialbots
for reasons such as privacy concerns.
Instead, the socialbots receive the user utterance hypotheses from the ASR.
The ASR output format is the widely used N-best list, which consists of $N$ most
likely hypotheses of the user utterance.
The output also includes confidence scores for individual words in the
hypothesis.
In the literature, a common way to obtain word confidence scores is to use the
word lattice or confusion network~\cite{Mangu2000findingconsensus}.
Having multiple hypotheses and word confidence is useful for spoken dialog
systems to automatically recover ASR errors and decide whether to ask the user
to repeat or rephrase their utterance.
At the time of the inaugural competition in 2017, the Alexa Prize infrastructure
did not provide information for word timing, sentence segmentation or pauses.
In the 2018 competition, the start and end timing of individual tokens are
provided.

{\bf TTS}:
In the Alexa Prize infrastructure, the TTS converts the textual response to
Alexa's female voice.
A certain level of control on the synthesis can be achieved using the Speech
Synthesis Markup Language (SSML)~\cite{SSML}, which can specify word
pronunciation, prosody, speaking rate, etc.
These controls can help the socialbot to deliver the intended message correctly
and effectively,
e.g., pronouncing unusual words appropriately (e.g., ``Timon''),
improving the naturalness of concatenated sentences (e.g., control the length of
the break between sentences),
and emphasizing certain words in the sentence (e.g., the topic choices).

\subsection{Language understanding}
\label{chap2:ssec:nlu}
Language understanding provides necessary information for socialbots to come up
with an appropriate response.
Most socialbots carry out a series of analyses on the N-best hypotheses of user
utterances as part of the language understanding process.
Although there is no consensus among socialbot developers on the definition and
organization of these analyses, the three most common analyses examine intents,
entities, and sentiments, respectively. 


{\bf Intent Analyses}:
Most socialbots carry out intent classification as a multi-class classification
problem using a customized set of intents defined according to system needs and
capabilities.
Some socialbots also use a set of domain-independent intents.
Gunrock~\cite{UCDavisGunrock2018}
performs dialog act tagging using a subset of dialog acts defined for the
Switchboard corpus \cite{Stolcke2000CL}.
Roving Mind~\cite{UTrentoRovingMind2017} defines the intent as a dialog act
structure and its qualifiers, following an ISO standard dialog act annotation
schema \cite{Bunt2012ISO}.
The dialog act structure consists of a semantic dimension (e.g., social
obligation) and a communicative function (e.g., thanking), whereas the
qualifiers include sentiment polarity, factual information type, and subject
information type.
Following the Phoenix semantic grammar~\cite{Ward1994HLT},
Tartan~\cite{CMUTartan2018} represents the intent in the form of a semantic
grammar which tags individual segments of the utterance with pre-defined
concepts and identifies slots carried by the concept.

{\bf Entity Analyses}:
Identifying entities is important for socialbots since the conversation often
centers on them.
Entities include both named entities (e.g., persons, locations, organizations)
and key noun phrases (e.g., cars, artificial intelligence).
Named entities are usually extracted via a named entity recognizer from toolkits
such as Stanford CoreNLP~\cite{CoreNLP}, Google NLP~\cite{GoogleNLP}, Alexa
Skill Kits~\cite{Kumar2017NIPS}, SpaCy~\cite{SpaCy}.
Noun phrase extraction can rely on regular expression, noun phrase chunker,
constituency parsing, or dependency parsing.
In addition to entity extraction, several
socialbots~\cite{CTUAlquist2017,CTUAlquist2018,CMURubyStar2017,UTrentoRovingMind2017,HWUAlana2018,UCSCSlugbot2018,UCDavisGunrock2018}
carry out entity linking or disambiguation by linking the entity to a node in
the knowledge base.
Another common entity analysis is coreference/anaphora resolution which
identifies expressions referring to the previously mentioned entities.

{\bf Sentiment Analysis}:
Many socialbots carry out sentiment analysis on user utterances, which is useful
for DM such as turn-taking strategies (e.g., acknowledging user) and dialog
control (e.g., switching topics).
They usually use off-the-shelf sentiment analyzers, such as the Stanford
CoreNLP~\cite{CoreNLP}, Google NLP~\cite{GoogleNLP}, and the VADER sentiment
analyzer~\cite{Hutto2014ICWSM} in NLTK~\cite{NLTK}.
Roving Mind uses a lexical-based analyzer following the method
in~\cite{Kennedy2015}.
However, most off-the-shelf sentiment analyzers are developed for other
domains.
For example, Stanford CoreNLP uses movie reviews and VADER uses social media
text.
Considering this domain mismatch issue, Sounding Board develops its own
lexical-based sentiment classifier for social chat, which is included as part of
the multi-dimensional user reaction analysis~\cite{UWSoundingBoard2017}.
While most socialbots analyze the sentiment at the utterance level,
Alquist~\cite{CTUAlquist2018} proposes an entity sentiment analysis in order to
form the socialbot's own opinion on an entity mentioned in the user utterance.
In particular, they search recent tweets containing the entity via Twitter API
and perform sentiment analysis on gathered tweets using a neural network model
trained on a movie review dataset~\cite{Mass2011ACL}.

\subsection{Dialog management}
\label{chap2:ssec:dm}
The DM is the central part of a socialbot, in charge of dialog control, dialog
context tracking, and response planning.
Considering the nature of the open-domain conversation, a common management
strategy is to break down the problem into a set of interaction modes.
Different ways of breakdown have been used: 
by topics (e.g., \cite{CTUAlquist2017,UCDavisGunrock2018,EmUIrisBot2018,CMUTartan2018}), 
by conversation activities (e.g., \cite{UWSoundingBoard2017,UCSCSlugbot2018}),
and by response generation methods (e.g., \cite{UMontrealMILA2017,UEdinburghEdina2017,CMURubyStar2017}).
Individual interaction modes are handled by corresponding components.
A master component is usually used to choose the target interaction modes
\cite{UWSoundingBoard2017,CMUTartan2018,UCDavisGunrock2018}.
Several representative techniques are used for implementing DM components.

{\bf AIML}:
Many
socialbots~\cite{HWUAlana2017,HWUAlana2018,UMontrealMILA2017,EmUIrisBot2018,PrincetonPixie2017,RPIWiseMacaw2017,UCBEigen2017}
use the Artificial Intelligence Markup Language
(AIML)~\cite{Wallace2003AIML,AIML} to code the dialog control rules and bot
responses, usually extending those from ALICE~\cite{Wallace2009ALICE}.
In this method, dialog control is handled by the AIML interpreter, which is
loaded with AIML files that contain a collection of knowledge units.
Each knowledge unit defines a pattern to match the user utterance, a list of
possible bot responses, and conditions that help the interpreter to select the
response to the matched user utterance.
While most knowledge units define two-turn conversations, multi-turn control can
be achieved using long-term context variables.
The AIML-based method is favored for its simplicity, but scalability might be an
issue to handle all kinds of user requests.
Also, it is less flexible for executing complex actions such as querying back-end
databases and APIs.

{\bf Response Retrieval}:
Several socialbots implement methods to retrieve human-written responses.
In this case, they directly obtain well-formed responses without the need of
realization or generation, and both dialog control and dialog context tracking
are heavily integrated into the retrieval process.
It should be noted that while ``response retrieval'' and ``information
retrieval'' are used interchangeably in the literature, we only use the former
to refer to methods described below and the latter (or the term ``content
retrieval'') to methods that retrieve social chat content to fill arguments in
templated responses. 
Both MILA~\cite{UMontrealMILA2017} and Pixie~\cite{PrincetonPixie2017} train
retrieval neural networks to select responses mined from social media and/or the
Cornell movie dialog corpus~\cite{Danescu2011CMCL}.
By encoding utterances using the StarSpace embeddings~\cite{Wu2018AAAI},
Tartan~\cite{CMUTartan2018} finds similar utterances from a question-answering
subset of the DailyDialog dataset~\cite{Li2017DailyDialog} for phatic user
utterances (e.g., ``{\it how are you}'') and returns the corresponding response.
RubyStar~\cite{CMURubyStar2017} uses entities detected in the user utterance to
search for replies via the Twitter search API.
Slugbot~\cite{UCSCSlugbot2018} uses Elasticsearch~\cite{Elasticsearch} to
search for responses from a repository consisting of mined Reddit posts,
hand-crafted responses, crowdsourced conversations, etc.
The search queries are built based on topics, entities, keywords, etc.
Due to the scarcity of publicly available corpora for social chat
conversations, Edina~\cite{UEdinburghEdina2017} crowdsources self-dialogs,
i.e., asking workers to create a conversation on the given topic by self-playing
both participants of the conversation.
They implement a matching score component to retrieve the most appropriate
response from the bank of self-dialogs using a hand-crafted scoring function.
Fantom~\cite{KTHFantom2018} crowdsources responses by asking the worker to
create a bot response given a dialog history.
The collected responses are stored in a dialog graph, where each node is either
a bot utterance or a user utterance.
An edge always starts from a user utterance node to a bot utterance node,
indicating that the bot utterance is an appropriate response to the user
utterance.
In this way, a path from one node to the other encodes the dialog context.
The dialog graph is periodically enriched by automatically creating
crowdsourcing units from new socialbot conversations.
A graph search algorithm is proposed to retrieve bot responses using a learned
semantic similarity function with hand-crafted features.

{\bf Finite State Machine}:
Several socialbots use a finite-state-based dialog control
mechanism~\cite{CMUMagnus2017,CMUTartan2018,SNUChattyChat2017,CTUAlquist2017,UCSCSlugbot2017,UCSCSlugbot2018,UWSoundingBoard2017}.
The dialog state can be defined based on the progress of a specific conversation
activity, and it constitutes a portion of the overall dialog context.
The state transitions rely on the dialog context maintained by the DM.
Although the DM can be automatically learned from conversation data, the state
transitions in current socialbots are mostly hand-crafted 
with some randomness allowed among multiple possible transitions.
In the hand-crafted setting, this approach becomes similar to the AIML-based
approach since both are rule-based.
Nevertheless, it usually gives much more flexibility than AIML for customizing
complex actions at individual dialog states.
Traditionally, finite-state-based dialog control with hand-crafted transitions is
used primarily for simple interactions as the number of states and transitions
increases rapidly with the interaction complexity.
Researchers have proposed several higher-level frameworks to help developers to
build the dialog control logic more
efficiently~\cite{Larsson1998NLE,Seneff2000,Bohus2009CSL}.
In order to handle complex and mixed-initiative interaction paradigms, Sounding
Board~\cite{UWSoundingBoard2017} proposes a hierarchical DM framework in
2017, where each specific conversation activity (called ``miniskill'' in
the framework) is managed by the corresponding DM that controls the state
transitions within the subdialog, and a master DM manages the transitions among
miniskills and dialog control logics that shared among different miniskills.
Several 2018 socialbots use a similar hierarchical approach for dialog state
management, e.g., Alquist~\cite{CTUAlquist2017},
Gunrock~\cite{UCDavisGunrock2018}, and Tartan~\cite{CMUTartan2018}.

\subsection{Response generation}
\label{chap2:ssec:nlg}
Here, we review three representative NLG processes in socialbots.

{\bf Template Realization}:
Some socialbots specify the template and associated arguments for the response
in the DM, and the NLG module simply performs template realization with some
variations in the expression conveying the target dialog act.
Templates are mostly pre-written by developers.
Using the SSML to adjust the prosody and speaking rate is found to be helpful
for better delivering the response during the speech-based
interaction~\cite{UWSoundingBoard2017,UCDavisGunrock2018}.
Note that in the NLG literature, sentence realization can also use grammar rules
rather than templates (see a comprehensive review
in~\cite[Chapter~6]{Reiter2000NLG}).  
Current socialbots have not explored this approach yet.

{\bf Neural Response Generation}:
With the recent advances in deep learning, the neural sequence-to-sequence
approaches have shown promising results for response
generation~\cite{Vinyals2015ICML,Sordoni2015NAACL,Li2016ACL}.
Several socialbots include neural response generation models in NLG.
For example, MILA~\cite{UMontrealMILA2017} extends the hierarchical latent
variable encoder-decoder model~\cite{Serban2017AAAI};
Alquist~\cite{CTUAlquist2017} uses the sequence-to-sequence model
\cite{Sutskever2014NIPS};
Tartan~\cite{CMUTartan2018} explores the ParlAI \cite{ParlAI2017} framework
which uses a memory network~\cite{Sukhbaatar2015NIPS}.
While this approach seems to be promising as it can take advantage of
large-scale conversation data, further efforts are needed to achieve satisfying
conversation quality for socialbots~\cite{CMUTartan2018},
and most systems also include template-based response realization.

{\bf Response Ranking}:
For socialbots that produce multiple candidate responses, a ranker is needed to
select the best one as the ultimate response.
The ranking task has been formulated as a classification
problem~\cite{CMURubyStar2017,HWUAlana2017} and a reinforcement learning
problem~\cite{UMontrealMILA2017}.
Alana~\cite{HWUAlana2017} also experiments with a hand-crafted ranking function without
training.

\section{Social Chat Knowledge Management}
\label{chap2:sec:knowledge}
As discussed earlier in \S\ref{chap2:sec:competition}, it remains open question
that what types of knowledge are useful for socialbots.
Here we group social chat knowledge into four categories: 
backstory (e.g., name, age, hobbies),
factual knowledge (e.g., the birthday of a celebrity), 
social chat content (e.g., recent news, interesting facts),
and conversation responses.

{\bf Backstory}:
Manual curation of a backstory is straightforward and guarantees a consistent
bot persona.
Several socialbots use AIML to customize the backstory
responses~\cite{HWUAlana2017,HWUAlana2018,UMontrealMILA2017,PrincetonPixie2017}.
Although not used in Alexa Prize socialbots, there have been studies on
maintaining consistent persona for data-driven neural response generation
models~\cite{Li2016ACL}.

{\bf Factual Knowledge}:
Socialbots have used public KBs to acquire factual knowledge.
Popular KBs include Google Knowledge Graph~\cite{GoogleKnowledgeGraph},
Microsoft Concept Graph~\cite{MicrosoftConceptGraph},
DBpeida~\cite{DBpedia2015}, Freebase~\cite{Freebase2008},
Wikidata~\cite{Wikidata2014}, ConceptNet\cite{ConceptNet2017}, etc.
A typical usage of factual knowledge in socialbot conversations is providing
a short summary of the entity using attributes associated with the entity
(e.g., \cite{PrincetonPixie2017,CMURubyStar2017}).
In Sounding Board, we use Amazon Evi~\cite{Evi} to get the definition of a
person or an organization that is mentioned in the current bot turn.
As a turn-taking strategy, the bot asks the user that whether they know the
entity or not. 
Depending on the user answer, the bot can decide to explain the entity in
the next turn.
Leveraging SPARQL queries, Alana~\cite{HWUAlana2018} discovers novel and
interesting connections in the Wikidata graph that relate an entity with
another entity in the KB.
Then it uses templated-based response generator to render this discovered
information to the user.
Gunrock~\cite{UCDavisGunrock2018} uses the ConceptNet to find entities that are
relevant to the current topic and extends the discussion on a relevant entity.
Slugbot~\cite{UCSCSlugbot2018} compiles a large-scale ontology from multiple
sources and explores its usage for tour recommendations in the travel domain.

{\bf Social Chat Content}:
As acknowledged by several teams, collecting social chat content is one of the
greatest challenges for Alexa Prize socialbots.
A wide variety of content sources are needed to cover a wide range of topics of
interest.
Early in the competition, we found interesting facts to be good social chat
content and collected a large scale of TodayILearned submissions from
Reddit\footnote{\url{https://www.reddit.com/r/todayilearned}}.
Other socialbots subsequently adopted this type of social chat content.
Types of social chat content that have also been explored include debate
opinions, amusing thoughts, jokes, news headlines, etc.
They are usually crawled from social media platforms (e.g., Reddit, Twitter) or
using public APIs (e.g, Goodreads API~\cite{GoodreadsAPI}, News
API~\cite{NewsAPI}).

Additional language analyses are usually carried out to process and organize the
collected content.
Sounding Board~\cite{UWSoundingBoard2017} indexes content by noun phrase and
filters content that is social inappropriate or not suitable for voice-based
interactions.
Gunrock~\cite{UCDavisGunrock2018} uses the OpenIE tool~\cite{Angeli2015ACL} to
extract relations between two entities from the plain text.
Roving Mind~\cite{UTrentoRovingMind2017} adapts the ClausIE
framework~\cite{ClausIE} to extract $\langle$subject, predicate, object$\rangle$
triplets from news articles and links the article to specific
key phrases.
It also explores automatic extraction of opinion quotes from news articles.
Magnus~\cite{CMUMagnus2017} uses an opinion mining system that retrieves and
summarizes opinions relevant to user's opinion-seeking questions.

{\bf Conversation Responses}:
As discussed in \S\ref{chap2:ssec:dm}, the response-retrieval-based methods
directly select well-formed utterances from a bank of conversation responses.
Compared with factual knowledge and social chat content, these conversation
responses are ready to use without further realization.
They are usually mined from social media (e.g., Twitter, Reddit) or existing
public conversation datasets (e.g., the Cornell movie dialog
corpus~\cite{Danescu2011CMCL}, the DailyDialog
dataset~\cite{Li2017DailyDialog}).
To obtain conversations that are close to the socialbot setting,
Alana~\cite{UEdinburghEdina2017} proposes a self-dialog crowdsourcing task and
Fantom~\cite{KTHFantom2018} develops an automatic pipeline for creating
crowdsourcing units from new socialbot conversations under their dialog graph
framework.



\section{Socialbot Evaluation}
\label{chap2:sec:evaluation}
Currently, socialbots are evaluated primarily by Alexa users who give a rating
upon finishing their conversations with a socialbot.
University teams mostly use user ratings for assessing the system performance
and perform A/B testing for system diagnosis.
Besides the conversation-level user ratings, teams also use conversation
duration and number of turns to assess conversation quality.
Several teams find the number of turns positively correlated with user
ratings~\cite{UTrentoRovingMind2017,RPIWiseMacaw2017,CMURubyStar2017}, although
the correlation is relatively weak~\cite{HWUAlana2018}.
In this section, we review current research on socialbot evaluation from three
aspects: proxy metrics for user ratings, user characteristics that affect user
ratings, and automatic user rating prediction.

{\bf Proxy Metrics for User Ratings}:
One type of analysis looks for proxy metrics to approximate user ratings.
The user sentiment is found to be slightly correlated with user ratings in
several studies~\cite{UTrentoRovingMind2017,CMUTartan2018}.
The Alana team~\cite{HWUAlana2018} studies the percentage of user turns with
positive/negative reactions, which are identified by pre-defined
key phrases and automatically derived sentiment polarity.
By computing the correlation between the percentages and user ratings, their finding
suggests that the percentage of positive user turns is positively correlated
with user ratings, although the correlation is as low as the number of total
turns. 
The percentage of negative user turns shows a much lower
correlation.
They further investigate using either the user rating or the number of turns as
the optimization objective for their response ranker, and find that the user
ratings are not stable enough to be an optimization objective in this setup.
Venkastech et al.\ invesitage a collection of evaluation metrics for socialbots
in addition to user ratings~\cite{Venkastech2017NIPS}.
In contrast to user ratings which rate conversations based on overall
experience, they recruit a set of Alexa users to rate their
conversations based on engagement and interestingness, which turns out to be
generally lower than normal user ratings.
They also annotate socialbot responses in terms of coherence and compute a
response error rate for each socialbot.
Several other automatically computed metrics are studied, including domain
coverage, conversation depth, and topical diversity.
Their correlation analysis indicates that the response error rate, average
engagement ratings, median conversation duration, and conversation depth are
four metrics with highest correlations with user ratings.
Guo et al.\ focus on topic-based metrics for evaluating socialbots and correlate
them with user ratings~\cite{Guo2017NIPS}.
Specifically, they train a classifier to detect the topic of each utterance and
identify topic coherent sub-conversations defined as consecutive turns on the
same topic.
It should be noted that in both \cite{Venkastech2017NIPS} and
\cite{Guo2017NIPS}, the correlation analysis is carried out at the system level,
i.e., computed for each socialbot rather than for individual conversations.
Therefore, the correlation is much higher than those computed over
conversations.
Although not studied in these two papers, some of these metrics (e.g.,
conversation depth, topical diversity) can be calculated for individual
conversations and used as proxy user ratings.

{\bf User Characteristics vs.\ User Ratings}:
Due to their subjectivity, user ratings are known to have high variance.
Several studies investigate the potential effect of user characteristics on user
ratings.
The Tartan team~\cite{CMUTartan2018} finds that user's mood affects their
ratings, e.g., users classified as in a great mood rate conversations on average
1.4 point higher than those classified as unhappy.
Ji et al.\ find that users who curse more tend to rate the conversation lower
than normal users~\cite{RPIWiseMacaw2017}.
The studies in~\cite{Venkastech2017NIPS} indicate that frequent users who have
had at least two conversations with a particular socialbot give lower ratings
than general users.
Our initial analysis in~\cite{Fang2018NAACL} suggests that user personality
traits are correlated with user ratings, where users that are more extraverted,
agreeable, or open to experience tend to rate the conversation higher.

{\bf User Ratings Prediction}:
Another line of research studies automatic user rating predictions for socialbot
conversations.
With a collection of socialbot conversations and associated conversation-level
user ratings, Venkastech et al.~\cite{Venkastech2017NIPS} train deep neural networks and ensemble models
using a variety of features (n-grams of user-bot turns, token overlap between
user utterance and socialbot response, conversation duration, number of turns,
and mean response time) to evaluate socialbot
conversations.
Under a reinforcement learning setting, Serban et al.~\cite{Serban2017arXiv}
learn an ensemble of
linear regression models to predict user ratings as reward signals using 23
features including dialog length, features that characterize the bot utterance
(appropriateness, genericness, length), and features that characterize the user
utterance (sentiment, genericness, length, confusion).







%



\section{Limitations of Existing Work}
\label{chap2:sec:limitation}
At the time of the first Alexa Prize competition, no prior dialog system can be
easily used for socialbots.
We think that developing a system with rule-based dialog control and careful
dialog design is a necessary step towards data-driven and statistical
frameworks.
We also emphasize the importance of maintaining large-scale and real-time social
chat content.
Chapter~\ref{chap3} describes the socialbot system we developed with these
considerations in mind during the first competition in 2017.
Specifically, we develop a socialbot architecture that has a hierarchical DM
framework and a knowledge back-end with dynamically updated social chat content.
We believe these are important steps for current socialbot research.
In fact, several competing socialbots in 2018 Alexa Prize have developed similar
hierarchical DM frameworks~\cite{CTUAlquist2018,UCDavisGunrock2018,CMUTartan2018}
and there is a trend of exploiting much more types of social chat content in
2018 socialbots.

For socialbot evaluation, some metrics have been studied as an automatic measure
of the conversation quality as discussed in \S\ref{chap2:sec:evaluation}.
However, none of these metrics achieves a higher correlation with
conversation-level user ratings than the number of total turns, nor there are
extensive analysis of socialbot conversations in terms of aspects that may
influence the user ratings.
In Chapter~\ref{chap4}, using 68k rated conversations collected from a stable
version of Sounding Board, we carry out in-depth analysis on a variety of
metrics that may reflect the conversation quality from different aspects.
Through this analysis, we successfully identify automatic metrics that have
significantly higher correlation with user ratings than the number of total
turns.

While automatic conversation rating prediction models have been investigated for
socialbot evaluation, an issue neglected by previous work is dialog segment
evaluation.
A conversation can contain both good and bad segments.
Although conversation-level ratings provide a measure for the overall quality
of the conversation, they do not uniformly apply to individual segments.
Automatic dialog segment evaluation is useful for system diagnosis, which is an
important step in the system development cycle.
To address the dialog segment evaluation problem, Chapter~\ref{chap4} also
studies two multi-level models for predicting scores of dialog segments at
different granularity. 
Dialog segment scores can help developers identify successful and problematic
dialog segments in a conversation.
They can also be used for diagnosing dialog policies and content sources as we
will illustrate in Chapter~\ref{chap4}.
Potentially, they can benefit data-driven dialog policy learning in a
statistical dialog system by using the scores as dialog policy rewards, which we
leave as future work.

Lastly, a critical issue faced by all socialbots is the shallow
conversation.
To address this issue, some socialbots have designed structured subdialogs.
For example, Alquist~\cite{CTUAlquist2018} authors topic-specific subdialogs.
In Chapter~\ref{chap3}, we describe a miniskill for movie discussions that uses
structured KB about movies.
However, the challenge remains for in-depth discussions on unstructured
information such as news articles.
In response to this challenge, we propose a document representation for news
articles in Chapter~\ref{chap5}. 
We also develop methods for constructing the document representation
automatically.
A miniskill is designed and implemented in a socialbot to demonstrate the
usability of the proposed document representation and dialog strategies.


\chapter{The Sounding Board System}
\label{chap3}
In this chapter, we describe the Sounding Board system, which competed in the
2017 Alexa Prize finals.\footnote{%
	The Sounding Board system is developed by the University of
	Washington team competing in the 2017 Alexa Prize.
	The team includes five student members: Hao Cheng, Elizabeth Clark, Hao Fang
	(team lead), Ari Holtzman, and Maarten Sap.
	The team is advised by three faculty members: 
	Prof.\ Yejin Choi, Prof.\ Mari Ostendorf (faculty lead), and Prof.\ Noah
	Smith.
	While this chapter describes the whole Sounding Board system, my primary
	contributions are system architecture design and implementation including the
	multi-dimensional language understanding, hierarchical dialog management, and
	social chat knowledge graph construction.
	It is the joint efforts from all team members and faculty advisors that make
	the Sounding Board a success.
	For example, the movie discussion and the integration of movie knowledge into
	the social chat knowledge graph are primarily designed by Hao Cheng.
	The personality assessment component is primarily designed by Maarten Sap.
	A variety of language analyses and dialog acts are developed by Elizabeth
	Clark and Ari Holtzman.
	A portion of this chapter has been published in
	\cite{UWSoundingBoard2017,Fang2018NAACL}.
}
We first describe the high-level system design of Sounding Board in
\S\ref{chap3:sec:dialog_strategy}.
Then, the overall system architecture is presented in
\S\ref{chap3:sec:system_arch}.
The language understanding, dialog management, and language generation
modules are discussed in \S\ref{chap3:sec:nlu}--\ref{chap3:sec:nlg},
respectively, followed with the social chat knowledge graph described in
\S\ref{chap3:sec:kg}.
The details of implemented miniskills are provided in
\S\ref{chap3:sec:miniskills}.
Finally, a sample conversation is provided in
\S\ref{chap3:sec:sample_conversation} to illustrate system capabilities.

\section{System Design}
\label{chap3:sec:dialog_strategy}
Our goal is to build a socialbot that is capable of mixed-initiative and
open-domain social chat.
To be mixed-initiative, the dialog strategy needs to specify ways to lead the
conversation in order to engage the user, while granting the user the freedom to
interrupt the conversation flow and switch topics at any time.
Open-domain means that the dialog strategy needs to be backed with a variety of
conversation activities and content to cover the wide range of user interests.

{\bf Design Objectives}:
Sounding Board's system design has two key objectives, i.e., to be
user-centric and content-driven.
The bot is {\it user-centric} in that users can control the conversation flow at
any time and bot responses are adapted to acknowledge user reactions.
The dialog control further takes into account the user's topic interests gauged
by the bot during the conversation.
The bot is also {\it content-driven}, as it continually supplies interesting and
relevant information to continue the conversation, enabled by a rich content
collection that it updates daily.
It is this content that allows the bot to drive the conversation forward and
engage users for a long period of time.

\begin{table}[t]
	\centering
	\begin{tabular}{l|p{11cm}}
		\hline\hline
		{\bf Miniskill}  & {\bf Description} \\
		\hline
		{\sl Greet} 									
		& Greets the user and acknowledges the answer accordingly. \\
		{\sl Exit} 									
		& Instructs the user to explicitly say ``{\it stop}'' to exit Sounding Board. \\
		{\sl List Activities}
		& Lists available conversation activities. \\
		{\sl List Topics}
		& Lists available discussion topics. \\
		\hline
		{\sl Ask Trivia Questions}
		& Asks a trivia question about a topic. \\
		{\sl Tell Fun Facts}
		& Tells a fun fact about a topic. \\
		{\sl Tell Amusing Thoughts}
		& Tells an amusing thoughts about a topic. \\
		{\sl Tell Life Pro Tips}
		& Tells a life pro tip about a topic. \\
		{\sl Tell Jokes}
		& Tells a joke about a topic. \\
		{\sl Discuss News Headlines}
		& Discuss a news headline about a topic. \\
		{\sl Discuss Movies}
		& Discuss a movie. \\
		\hline
		{\sl Ask Personality Questions}
		& Asks a sequence of questions for probing the user's personality. \\
		\hline\hline
	\end{tabular}
	\caption{Sounding Board's conversation miniskills.}
	\label{tab:miniskills}
\end{table}

{\bf Conversation Miniskills}:
Several conversation activities are designed for Sounding Board.
These activities are implemented as {\it miniskills} of Sounding Board and
summarized in Table\,\ref{tab:miniskills}.
The first four miniskills are functional activities, whereas the next seven
miniskills are content-oriented.
The last miniskill, {\sl Ask Personality Questions}, is a special miniskill designed to calibrate
user personality so that the bot can adjust its topic suggestions accordingly.
Details about these miniskills will be provided in \S\ref{chap3:sec:miniskills}.

{\bf Miniskill Dialog Control}:
In Sounding Board, each miniskill manages a short dialog segment and the dialog
flow can be viewed as a sequence of activity states.
Thus, the dialog control for each miniskill is formulated as a finite-state
machine.
The inventory of states is organized based on the primary dialog act of the bot
at a turn in the envisioned dialog flow of the miniskill.
For example, the inventory for the {\sl Discuss Movies} miniskill includes a
state to confirm with the user about discussing a specific movie,
a state to inform the release year, genre, and director(s) of the movie, 
a state to elicit user comments about the movie if they have already watched it,
a state to inform the plot, a state to inform the review, 
and several states to discuss the director(s) and main actor(s) of the movie.
While the bot has a default dialog flow for the miniskill, the dialog control
still allows the user to take the initiative at any time.
For example, when the bot confirms to discuss a specific movie, the user can
directly ask for the review (e.g., ``{\it was it good?}'') or the plot (e.g.,
``{\it what is it about?}'').
In this case, the miniskill transits to the corresponding state based on the
user's interest, rather than following its default dialog flow of transitioning 
to the state of informing the release year, genre, and director(s) of the movie.

{\bf Dialog Control Principles}:
Sounding Board has three general principles for dialog control.
The {\it coherence} principle requires the bot to be coherent,
i.e., the content should be on the same or a related topic within a dialog
segment.
The {\it user experience} principle requires the bot to properly
acknowledge the user, explicitly or implicitly explain the bot action, 
and instruct the user with available options when they are confused.
The {\it engagement} principle encourages the bot to be diverse in terms
of the topics, conversation activities, and its language use in the response.
It should avoid presenting topics or content that have already been used
in the conversation. 
It also biases the bot to actively suggest relevant topics or conversation
activities.
For example, when the user initiates a topic which the bot does not have any
content about, the bot will suggest a related topic to continue the
conversation.
Similarly, when the user asks a question, the bot can first give the answer
and then suggest some relevant content about a topic extracted from the
question.

\section{System Architecture}
\label{chap3:sec:system_arch}
\begin{figure}[t]
	\centering 
	\includegraphics[width=\textwidth,trim=0cm 1cm 0cm 1cm]{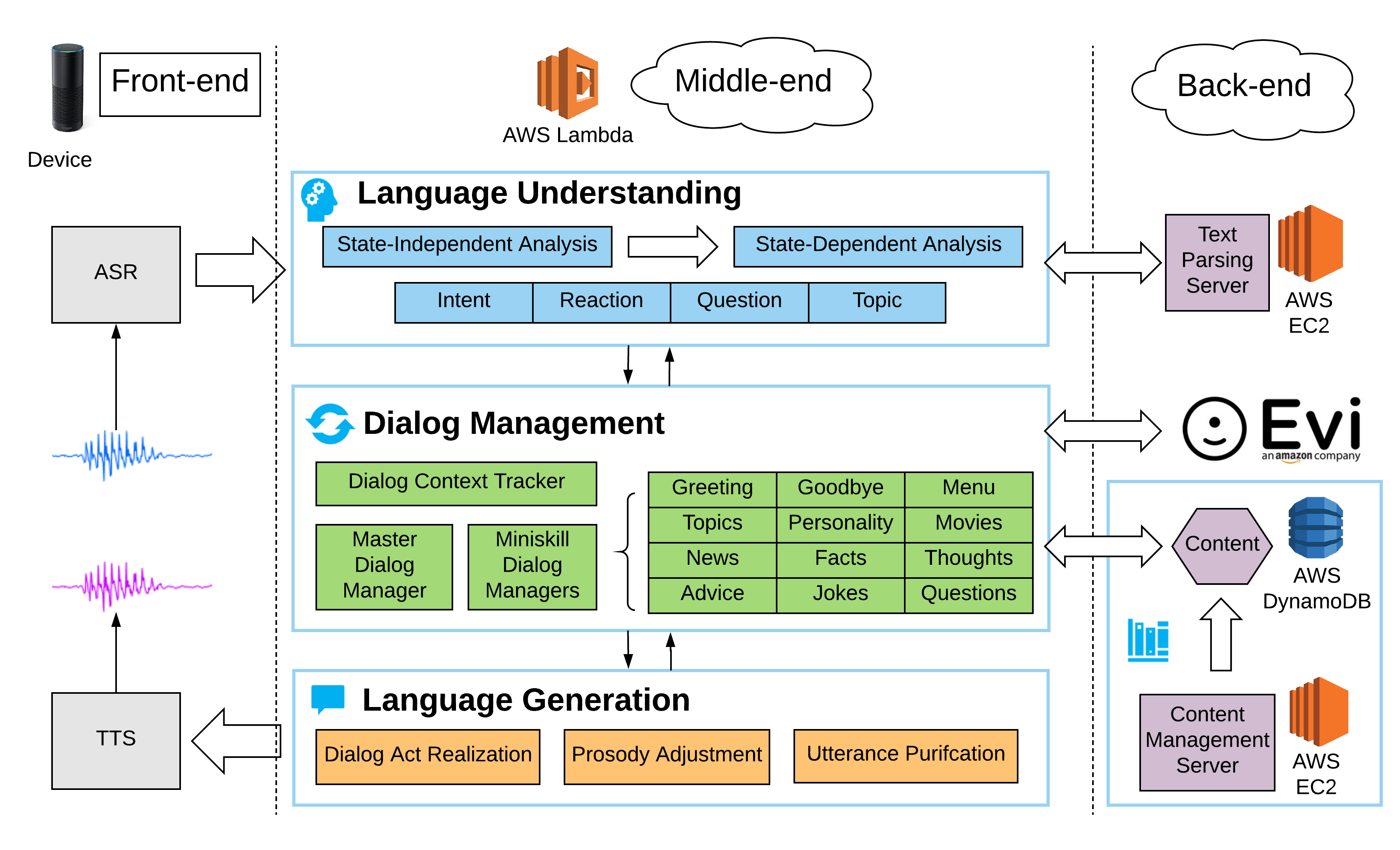}
	\caption{The system architecture of Sounding Board. 
		{\bf Front-end}: an Alexa-enabled device and Amazon's APIs for speech
		processing.
		{\bf Middle-end}: an AWS Lambda endpoint hosting core system modules of
		Sounding Board.
		{\bf Back-end}: External services and AWS DynamoDB tables for storing the
	knowledge graph.}
	\label{fig:system_arch}
\end{figure}

Fig.\,\ref{fig:system_arch} outlines the system architecture of Sounding
Board.
For clarity, we divide the whole system architecture into three parts: 
1)~the front-end, which could be any Alexa-enabled device and include Amazon's
APIs for speech processing,
2)~the middle-end, which is an AWS Lambda endpoint that hosts core system
modules of Sounding Board,
and 3)~the back-end, which includes a collection of supportive services for
Sounding Board.

{\bf Front-end}:
At the front-end, user utterances are transcribed from audio signals to textual
hypotheses through Amazon's automatic speech recognition (ASR) API. 
For each user utterance, multiple hypotheses are provided, together with
confidence scores for individual tokens in each hypothesis.
The Alexa Prize platform has adapted the ASR module, which was
originally designed for commands, to better handle social chat speech. 
The front-end also converts socialbot utterances from text to voices using
Amazon's text-to-speech (TTS) API.
The prosody of bot utterances can be adjusted using the speech synthesis markup
language (SSML).

{\bf Middle-end}:
The middle-end hosts three core system modules of Sounding Board: 
1)~the natural language understanding (NLU) module that carries out
multi-dimensional language analysis on user utterances,
2)~the dialog management (DM) module that tracks the dialog context and
controls the dialog flow,
and 3)~the natural language generation (NLG) module that realizes the response
plan and adjusts the prosody and wording of the bot utterance.
Details of these three modules are described in subsequent sections
(\S\ref{chap3:sec:nlu}--\ref{chap3:sec:nlg}) in this chapter.
These system modules are implemented as a whole with a serverless architecture
so that the bot can benefit from AWS Lambda's continuous scaling for handling
incoming traffic.

{\bf Back-end}:
At the back-end, we deploy a Stanford CoreNLP server \cite{CoreNLP} on an AWS
EC2 instance to run part-of-speech tagging and constituency parsing for user
utterances. 
Since ASR transcripts are case-insensitive, we configure the CoreNLP server to
use its pre-trained caseless models.
Amazon Evi~\cite{Evi} is used for answering factual questions and certain
personal questions.
The majority of the social chat knowledge graph for Sounding Board is stored in
AWS DynamoDB tables.
Part of these tables are updated daily by a content management server, which
will be described in \S\ref{chap3:sec:kg}.
The conversation logs and tracked dialog context attributes
are also stored in corresponding AWS DynamoDB tables.

\section{Multi-Dimensional Language Understanding}
\label{chap3:sec:nlu}
\begin{table}[t]
	\centering
	\begin{tabular}{l|l}
		\hline\hline
		{\bf Frame}  & {\bf Description} \\
		\hline
		\texttt{IntentFrame} 		& the user's intent and its associated slots \\
		\texttt{QuestionFrame}	& the user's question and its type \\
		\texttt{ReactionFrame}  & a multi-dimensional representation of user reactions \\
		\texttt{TopicFrame} 		& all possible topics extracted from the user utterance \\
		\hline\hline
	\end{tabular}
	\caption{High-level frames in the meaning representation of user utterances.}
	\label{tab:nlu_frame}
\end{table}

To characterize user utterances, a meaning representation is designed for
socialbot conversations.
The high-level frames in the meaning representation are listed in
Table\,\ref{tab:nlu_frame}.
Note that a user utterance may specify multiple attributes in these frames in
one turn.
For example, the user utterance 
``{\it that's boring let's talk about science}'' 
involves an attribute in the \texttt{ReactionFrame} 
indicating the negative interest (``{\it that's boring}'')
and an attribute in the \texttt{IntentFrame}
indicating the intent of setting a topic (``{\it let's talk about science}'').
However, the current ASR hypotheses do not provide punctuation or segmentation
information.
Thus, the meaning representation is at the utterance level and these frames are
constructed independently.

The NLU module carries out four sets of language analyses to
build individual frames in Table\,\ref{tab:nlu_frame}.
The analysis usually involves a stage that does not depend on the state of the
dialog flow, followed by a second stage that uses the state information to
refine the corresponding frame.
The dialog state information is useful for handling ellipsis and anaphora
resolution in user utterances.
For example, ``{\it yes}'' can mean acceptance or agreement depending on the
question asked by the bot, which could be either 
``{\it do you want to talk about technology}''
or ``{\it do you think that's true}''.

In this section, we first describe the intent analysis for \texttt{IntentFrame}
in \S\ref{chap3:ssec:intent_analysis}.
The question analysis for \texttt{QuestionFrame} is presented in
\S\ref{chap3:ssec:question_analysis}.
The user reaction analysis for building the \texttt{ReactionFrame} is
illustrated in \S\ref{chap3:ssec:user_reaction_analysis}.
Finally, the topic analysis used for constructing the \texttt{TopicFrame} is
described in \S\ref{chap3:ssec:topic_analysis}.

\subsection{Intent analysis}
\label{chap3:ssec:intent_analysis}
The \texttt{IntentFrame} is a structured representation for user utterances that
stores the user's intent and associated arguments (a.k.a.\ slots).
Table\,\ref{tab:nlu_intents} summarizes the inventory of 17 intents we developed
for Sounding Board, which is potentially generalizable for other socialbots.
These intents are categorized into five broad groups:
menu commands,
dialog control commands, 
topic commands,
other commands,
and non-command intents.
Note that the command names below are intent labels -- the user is not
constraint to use these specific phrases.

\begin{table}[t]
	\centering
{\small
	\begin{tabular}{l|l|l}
		\hline\hline
		{\bf Intent}  & {\bf Description} & {\bf Example} \\
		\hline
		Menu Commands & & \\
		\tabitem \texttt{ListActivities}
		& request available activities
		& {\it go to the menu} \\
		\tabitem \texttt{ListTopics}
		& request available topics
		& {\it new topics} \\
		\tabitem \texttt{Help}
		& request help message
		& {\it help} \\
		\hline
		Dialog Control Commands & & \\
		\tabitem \texttt{Continue}
		& request to continue the activity
		& {\it keep going} \\
		\tabitem \texttt{Repeat}
		& request to repeat the last utterance
		& {\it what did you say} \\
		\tabitem \texttt{Cancel}
		& request to switch activity/topic 
		& {\it enough news} \\
		\tabitem \texttt{Exit}
		& request to exit
		& {\it I'm done talking} \\
		\hline
		Topic Commands & & \\
		\tabitem \texttt{SetTopic} 			
		& initiate a general chat activity
		& {\it let's chat about science} \\
		\tabitem \texttt{AskForFacts}
		& ask for facts
		& {\it tell me a fun fact about cats} \\
		\tabitem \texttt{AskForJokes} 		
		& ask for jokes
		& {\it do you know any joke about dogs} \\
		\tabitem \texttt{AskForTips}
		& ask for tips
		& {\it give me a tip about computers} \\
		\tabitem \texttt{AskForThoughts} 
		& ask for bot thoughts
		& {\it what are you thoughts about pizza} \\
		\tabitem \texttt{AskForNews}
		& ask for news
		& {\it recent technology news} \\
		\hline
		Other Commands & & \\
		\tabitem \texttt{PersonalityQuiz} 
		& request a personality quiz
		& {\it let's get to know each other} \\
		\tabitem \texttt{OutOfDomainCommand} 
		& out-of-domain requests
		& {\it play music} \\
		\hline
		Non-Command & & \\
		\tabitem \texttt{GeneralQuestion} 	
		& ask a non-command question 
		& {\it what's your name} \\
		\tabitem \texttt{GeneralStatement} 
		& make a general statement
		& {\it that's interesting} \\
		\hline\hline
	\end{tabular}
}%
	\caption{The list of intents used by Sounding Board.}
	\label{tab:nlu_intents}
\end{table}

{\bf Menu Commands}:
Both \texttt{ListActivities} and \texttt{ListTopics} are menu-like commands
which user may use to ask the bot to offer available conversation activities or
topics.
\texttt{Help} is a command for requesting state-dependent help messages.

{\bf Dialog Control Commands}:
This set of commands covers intents which the user can use to control the dialog
flow during a conversation activity.
\texttt{Continue} lets the bot continue its original dialog flow.
\texttt{Repeat} pauses the dialog flow by asking the bot to repeat itself.
\texttt{Cancel} terminates the activity but does not exits the conversation,
whereas \texttt{Exit} directly ends the conversation.

{\bf Topic Commands}: 
These intents are used to initiate a chat activity.
They are usually associated with a topic slot, whose value is extracted from the
user utterance by the slot filling model.
The topic slot is optional.
When there is no topic slot detected, the bot can choose a topic based on its
dialog policy.

{\bf Other Commands}:
\texttt{PersonalityQuiz} is a command that invokes the {\sl Ask Personality
Questions} miniskill.
\texttt{OutOfDomainCommand} includes commands to invoke specific activities that
are not currently implemented in Sounding Board.
They are usually Alexa functions such playing music or reading books.
The bot automatically detects these out-of-domain commands and responds with an
explanation of its limitations and instructions for exiting Sounding Board.

{\bf Non-Command Intents}:
Two additional general intents cover all user utterances that are not detected
as a command, distinguishing statements from questions.
Note not all interrogative utterances are treated as \texttt{GeneralQuestion}
since some commands can have a question form as well.
For example, the question ``{\it can you tell me some interesting facts}'' is
associated with the \texttt{AskForFacts} command;
``{\it what do you know about the University of Washington}'' is associated with
the \texttt{SetTopic} command.
Similarly, not all statements are treated as \texttt{GeneralStatement}.
For example, ``{\it I'm interested in science}'' is associated with the
\texttt{SetTopic} command.

To build the \texttt{IntentFrame}, we primarily use the Alexa Skill
Kits~\cite{Kumar2017NIPS}
to perform intent detection and slot filling.
This process is state-independent.
Note that intent detection may give an incorrect estimation, especially for
unseen utterances in the training data.
These utterances are usually indirect speech acts or involve ASR errors.
For example,
the utterances ``{\it I need to leave now}'' 
and ``{\it I'm gonna call it a day}'' are detected as the
\texttt{GeneralStatement} intent, but they are stating the \texttt{Exit} intent
in an indirect way.
The utterances ``{\it tell mean please}'' and ``{\it how bout technology}'' 
also confuse the intent detection model due to the word error ``mean'' and
``bout''. 
In principle, these corner cases can be added to re-train the models using the
Alexa Skill Kits.
However, since the Alexa Prize platform requires a manual certification process
to update the models that can take several days,
during the competition, we instead apply a set of regular-expression-based text
classifiers and rules to cover a set of corner cases. 

In the state-dependent stage, the intent analysis carries out coreference
resolution to replace certain intent or slot values by the actual value they
refer to.
For example, when the bot lists several topics, the topic slot 
value ``{\it the first topic}'' in the utterance 
``{\it I'm interested in the first topic}'' 
should be mapped to the actual topic.
Similarly, when the bot lists several activities, the user utterance 
``{\it let's do the first one}'' should be mapped to a corresponding command.
The coreference resolution is implemented using state-dependent rules.

Lastly, in addition to commands listed in Table\,\ref{tab:nlu_intents}, 
the bot can handle some fine-grained dialog control commands 
in certain dialog states, e.g., 
``{\it tell me the review of the movie}'' during the {\sl Discuss Movies}
miniskill.
These commands are also detected in the state-dependent intent analysis.

\subsection{Question analysis}
\label{chap3:ssec:question_analysis}
\begin{table}[t]
	\centering
	\begin{tabular}{l|p{7.5cm}|l}
		\hline\hline
		{\bf Question Type}  & {\bf Description} & {\bf Example} \\
		\hline
		Factual
		& general factual questions
		& \makecell[tl]{
			\tabitem {\it why is the sky blue} \\
		} \\
		\hline
		Backstory
		& questions about the bot persona, e.g., name, birthday, hobby 
		& \makecell[tl]{
			\tabitem {\it what's your name} \\
			\tabitem {\it what's your favorite color}
		} \\
		\hline
		Sensitive
		& questions about inappropriate topics or financial/legal advice
		& \makecell[tl]{
			\tabitem {\it can I have sex with you} \\
			\tabitem {\it which stock should I buy}
		} \\
		\hline
		Other
		& state-dependent questions such as greeting and information seeking 
		& \makecell[tl]{
			\tabitem {\it how are you} \\
			\tabitem {\it why did they do that}
		} \\
		\hline\hline
	\end{tabular}
	\caption{Coarse-grained question types distinguished by Sounding Board.}
	\label{tab:coarse_question_types}
\end{table}

We include the \texttt{QuestionFrame} in the meaning representation since
questions are prevalent in socialbot conversations.
The \texttt{QuestionFrame} consists of the question body and its type.
For non-command questions, we classify them into four coarse-grained types which
are handled with different question-answering methods or deflection strategies. 
These question types are described in Table\,\ref{tab:coarse_question_types}.
The question type classifier is implemented primarily based on regular
expressions and heuristics.

Sounding Board routes all {\bf factual questions} to Amazon Evi~\cite{Evi} to
retrieve an answer.
For {\bf backstory questions}, Sounding Board first examines its own backstory
repository and returns an answer if the requested attribute is found.
For each attribute, we create multiple answers and randomly choose one to
diversify responses.
If the attribute is not found in the backstory repository, the question is
routed to Amazon Evi.
{\bf Sensitive questions} are always deflected.
For {\bf other questions}, Sounding Board uses a combination of state-dependent
dialog control for frequent question and general responses from a universal
response repository.
For all question types, if no answer can be found, Sounding Board falls back to
use a random pre-defined deflection.

\subsection{User reaction analysis}
\label{chap3:ssec:user_reaction_analysis}
\begin{table}[!t]
	\centering
	\begin{tabular}{l|l|l}
		\hline\hline
		{\bf Attribute}  & {\bf Description} & {\bf Example} \\
		\hline
		\texttt{YesNoAnswer} & & \\
		\tabitem yes & user gives a ``yes'' answer & {\it yep} \\
		\tabitem no  & user gives a ``no'' answer & {\it nope} \\
		\hline
		\texttt{Acceptance} & & \\
		\tabitem accept 	& user accepts the confirmation & {\it sounds good} \\
		\tabitem reject 	& user rejects the confirmation & {\it not really} \\
		\tabitem uncertain & user is uncertain & {\it I don't know} \\
		\hline
		\texttt{Knowledge} & & \\
		\tabitem known & user knows about the information 
		& {\it I've heard about it} \\
		\tabitem not known & user does not know about the information 
		& {\it I didn't know it} \\
		\hline
		\texttt{Agreement} & & \\
		\tabitem agree & user agrees with the content/bot & {\it that's true} \\
		\tabitem disagree & user disagrees with the content/bot 
		& {\it I don't think so} \\
		\hline
		\texttt{Comment} & & \\
		\tabitem positive & user expresses positive sentiment 
		& {\it that's very cool} \\
		\tabitem negative interest & user expresses negative interests about the content 
		& {\it who cares} \\
		\tabitem duplicated content & user complains that the content has been presented 
		& {\it you've told me that} \\
		\tabitem doubt & user expresses doubts about the content 
		& {\it that's fake news} \\
		\tabitem confusion & user expresses confusion about the content/bot 
		& {\it I didn't get it} \\
		\tabitem other negative & user expresses other types of negative sentiment 
		& {\it that's sad} \\
		\hline\hline
	\end{tabular}
	\caption{User reaction attributes.}
	\label{tab:user_reaction_frame}
\end{table}

Sounding Board's chat activities involve frequent topic negotiation and content
presentation, both of which sometimes take the form of a yes-no question,
e.g., ``{\it do you want to \dots}'', ``{\it did you know that \dots}''.
While users frequently respond with a yes-no answer, the answers do not
necessarily contains ``{\it yes}'' or ``{\it no}''.
Analysis of user interactions further show that users also make comments by
expressing their emotions or taking positive or negative stance about the
content.
To characterize user reactions, the \texttt{ReactionFrame} uses a
multi-dimensional representation with five multi-valued attributes:
\texttt{YesNoAnswer},
\texttt{Acceptance}, \texttt{Knowledge}, \texttt{Agreement}, and
\texttt{Comment}.
These attributes and their values are described in
Table\,\ref{tab:user_reaction_frame}.
A null value is allowed for all attributes if the user utterance does not
reflect corresponding attribute.

For each attribute, we build a text classifier primarily using regular
expressions.
These classifiers are state-independent and executed in parallel.
Recall that a user utterance may specify multiple attributes in one turn.
For example, in the user utterance 
``{\it no I did not know that but that's interesting}'', \texttt{YesNoAnswer},
\texttt{Knowledge} and \texttt{Comment}
are all observed.
The state-dependent stage of the user reaction analysis primarily maps 
the value of \texttt{YesNoAnswer} to other attributes depending on the question
asked by the bot.

\subsection{Topic analysis}
\label{chap3:ssec:topic_analysis}
A primary activity in Sounding Board is chatting about interesting topics.
Here, a topic can be a noun or noun phrase that refers to an entity 
(e.g., Amazon, cat) or a generic topic (e.g., technology).
In addition to topic commands discussed in \S\ref{chap3:ssec:intent_analysis},
the NLU module further detects topics mentioned in the user utterance with the
following two motivations.
First, the intent analysis may sometimes fail detecting a \texttt{SetTopic}
command for unseen utterances or ASR errors. 
In this case, the DM module can use extracted candidate topics to make its
decision.
Second, sometimes we would like to introduce interesting asides to the
conversation by bringing up content about a topic mentioned in the user
utterance. 

The topic analysis uses constituency parsing results from the Stanford CoreNLP
server and extracts all nouns, their lemmas, and noun phrase segments.
We remove invalid topics 
(e.g., ``this year'', ``a lot'', ``something'')
and sensitive topics (e.g., ``sex'') from the list based on regular expressions.
The remaining topics are stored in the \texttt{TopicFrame} as a list.

\section{Hierarchical DM Framework}
\label{chap3:sec:dm}
The DM module is responsible for dialog context tracking and dialog control.
To handle the complex dialog control for socialbot conversations involving
multiple topics and miniskills, we propose a hierarchical DM framework that
decouples modules for efficient system development and maintenance.
The framework consists of a dialog context tracker, a master
dialog manager, and several miniskill dialog managers.
The dialog context tracker records attributes that are useful for dialog
control.
The master dialog manager handles the dialog control operating on topics and
miniskills, whereas the miniskill dialog manager controls the dialog flow within
a specific miniskill.

\subsection{Dialog context tracker}
\label{chap3:ssec:dialog_context_tracker}
\begin{table}[t]
	\centering
	\begin{tabular}{l|l}
		\hline\hline
		{\bf Attribute Set}  & {\bf Description} \\
		\hline
		Dialog State  			& the current miniskill and its state \\
		Dialog Topic 				& the current topic, its initiation context, and number
		of presented posts \\
		Topic History 			& the history of user-initiated and system-initiated topics \\
		Content History 		& the history of content posts that have been presented
		to the user \\
		Miniskill History 	& the history of miniskill usage \\
		User Personality 		& personality traits of the user estimated by the bot \\
		\hline\hline
	\end{tabular}
	\caption{Dialog context attribute sets tracked by Sounding Board.}
	\label{tab:user_profile}
\end{table}

The dialog context tracker maintains six sets of attributes as summarized in
Table\,\ref{tab:user_profile}.
The {\bf dialog state} is defined by the current miniskill and its state in the dialog
flow.
The set of possible dialog states is finite.
The {\bf dialog topic} records the current topic and its initiation context
information, e.g., whether it is suggested by the bot or the user.
It further counts the number of presented content posts about the current topic
after its initiation to estimate the user's interest level about the current
topic.
The {\bf topic history} tracks the status of observed topics, indicating whether
a topic has been discussed, suggested by the bot, or mentioned by the user.
Similarly, the {\bf content history} and the {\bf miniskill history} track the
usage of content posts and miniskills in the conversation, respectively.
They are primarily used by the DM module to avoid presenting duplicated content
and promote the diversity of conversation activities.
The {\bf user personality} records the personality traits of the user estimated
by the bot in the {\sl Ask Personality Questions} miniskill.
It also tracks user answers on individual personality-probing questions.  

Currently, the dialog context is reset at the beginning of the conversation.
We do not keep the dialog context across conversations since the device address
is not a reliable indicator of the user ID considering that each device may be
used by multiple users.
When speaker identification is available, maintaining long-term dialog context
for a user will be valuable for future work.

\subsection{Master dialog manager}
\label{chap3:ssec:master_dm}
The master dialog manager is responsible for handling user commands, executing
the miniskill dialog manager, and making decisions on the next topic and
miniskill.
At each turn, if a command is detected, the master dialog manager terminates the
previous miniskill and executes the command-specific dialog control logic.
Otherwise, it hands off the dialog control to the active miniskill dialog
manager.
The miniskill may be terminated if it has already finished its dialog flow or
certain problems are detected (e.g., system errors, consecutive understanding
failures, negative user reactions).
In this case, the master dialog manager needs to choose a new miniskill and/or a
new topic.

{\bf Miniskill Polling}: 
To check whether a miniskill can take the control of the dialog, the master
dialog manager calls the polling function defined in the corresponding miniskill
dialog manager.
The master dialog manager usually polls multiple miniskills at one time and
then keeps the set of candidate miniskills that are available for taking the
dialog control.
Recall that the dialog context tracker records the miniskill usage history,
specifically, the turn number when each miniskill is lastly used in the
conversation.
The master dialog manager ranks available miniskills based on the turn number of
last usage, i.e., a smaller turn number has a higher priority.
A manually crafted decision tree is implemented for the master dialog manager to
select the miniskill, which takes into account the ranking results, the target
miniskill(s) specified by the user in the command if any, and other dialog
context attributes (e.g., whether the miniskill has been recently proposed by
the bot but rejected by the user).

{\bf Topic and Miniskill Backoff}:
When the user makes a topic command listed in Table\,\ref{tab:nlu_intents}, the
master dialog manager may not be able to find a miniskill that satisfies the
requirements specified in the command. 
In this case, a backoff strategy is used.
First, the master dialog manager removes the miniskill constraint by allowing
all content-oriented miniskills to handle the requested topic.
If no miniskill is able to handle this topic, it relaxes the topic constraint to
allow all candidate topics in the \texttt{TopicFrame} described in
\S\ref{chap3:ssec:topic_analysis}.
If there is still no content-oriented miniskill available, the master dialog
manager uses the \textsl{List Topics} miniskill. 

\subsection{Miniskill dialog manager}
\label{chap3:ssec:miniskill_dm}
Each miniskill dialog manager controls the dialog flow for a specific activity.
Miniskill dialog managers are developed independently with each other so that
the developer can upgrade an existing miniskill and add new miniskills with
minimal impacts on other miniskills.

Recall that the dialog flow of a miniskill is formulated as a finite state
machine.
In a miniskill dialog manager, a state is implemented via a dialog agent which
specifies its response plan and state transition function.
At a turn, the miniskill dialog manager executes the current state transition
function depending on the NLU results and the dialog context.
When the miniskill transitions to a new state, the corresponding dialog agent
produces the response plan by selecting a sequence of bot dialog acts for the
current turn and any content to be communicated.
Content can be a topic, some information from a knowledge source, an answer
obtained from Evi, part of the previous user utterance to be echoed,
etc.
As required by the Alexa Prize platform, the bot response consists of a message
and a reprompt.
The device always reads the message, and the reprompt is optionally used if the
device does not detect a response from the user.  
The response plan specifies the dialog acts for both the message and the reprompt.

\section{State-Independent Language Generation}
\label{chap3:sec:nlg}
The NLG module realizes the response plan produced by the DM module.
It has three state-independent processes:
dialog act realization, prosody adjustment, and utterance purification. 

\begin{table}[t]
	\centering
	\begin{tabular}{l|l}
		\hline\hline
		{\bf Dialog Act} & {\bf Example} \\
		\hline
		Informing Acts & \\
		\tabitem inform a news headline
		& {\it I read this article recently. The title is: \dots} \\
		\tabitem inform a fun fact
		& {\it My friend in the cloud told me that: \dots} \\
		\tabitem list topics 
		& {\it I can talk about technology, science, and cats.} \\
		\hline
		Requesting Acts & \\
		\tabitem confirm to present a news headline
		& {\it Do you want to hear some technology news?} \\
		\tabitem confirm to continue 
		& {\it Do you want to continue chatting about cats?} \\
		\tabitem ask for topic
		& {\it What do you want to talk about?} \\
		\hline
		Grounding Acts & \\
		\tabitem back-channelling 
		& {\it Cool!} \\
		\tabitem echo recognized topic
		& {\it Looks like you want to talk about news.} \\
		\tabitem echo user question
		& {\it I heard you ask: what is the name of that game.} \\
		\tabitem express gratitude 
		& {\it I'm happy you like it.} \\
		\tabitem signal non-understanding
		& {\it Sorry, I couldn't understand what you said.} \\
		\hline
		Instruction Acts & \\
		\tabitem describe the \texttt{ListToipcs} command
		& {\it You can say ``topics'' to hear some popular topics.} \\
		\tabitem describe the \texttt{Continue} command
		& {\it To move on, please say ``continue''.} \\
		\hline\hline
	\end{tabular}
	\caption{Examples of bot response dialog acts in Sounding Board.}
	\label{tab:dialog_acts}
\end{table}

{\bf Dialog Act Realization}:
The NLG module generates word sequences for each response dialog act specified in the
response plan.
Each dialog act has several variants for communicating the act.
When the dialog act is realized, a template is randomly selected to render the
utterance.
In Sounding Board, response dialog acts are grouped into four broad categories:
informing acts, requesting acts, grounding acts, and instruction acts.
Some examples are shown in Table\,\ref{tab:dialog_acts}.
Informing acts are used to present content, e.g., a fact, a joke, or a list of
topics.
While in most cases the informing acts take a statement form, sometimes they can
have a question form when the primary goal is to convey the information to
the user, e.g., ``{\it did you know that \dots}''.
In contrast, requesting acts are mostly information-seeking questions,
e.g., making confirmations or asking for opinions.
Grounding acts are primarily transition phrases/sentences used for
acknowledging the user, e.g., back-channeling, thanking, echoing the recognized
topic or question, comforting the user, etc.
Finally, instruction acts are help messages to instruct the user with
available commands.

{\bf Prosody Adjustment}:
The templates for dialog acts make extensive use of SSML to adjust the prosody
and pronunciation.
It helps to improve the naturalness of concatenated dialog acts,
to emphasize suggested topics, 
to deliver jokes more effectively,
to apologize or backchannel in a more natural-sounding way,
and to more appropriately pronounce unusual words.

{\bf Utterance Purification}:
Since the bot response may contain part of the recognized user utterance which
may include profanity,
the realized bot response is processed by a purifier that replaces profanity
words with non-offensive words randomly chosen from a list of innocuous nouns.
We use the nodejs-profanity-util package~\cite{NodejsProfanity}
with a profanity word list merged from multiple sources.
Based on test user feedback, some of the word replacements have an amusing
result.

\section{Social Chat Knowledge Graph}
\label{chap3:sec:kg}
\begin{table}[t]
	\centering
	\begin{tabular}{l|p{12cm}}
		\hline\hline
		{\bf Content Type} & {\bf Content Source} \\
		\hline
		Trivia Questions 	& SQuAD dataset \cite{SQUAD2016} \\
		Fun Facts 				& subreddit \texttt{TodayILearned} \\
		Amusing Thoughts 	& subreddit \texttt{ShowerThoughts} \\
		Life Pro Tips 		& subreddit \texttt{LifeProTips} \\
		Jokes 						& Amazon Evi API \\
		News Headlines 		& multiple subreddits: \texttt{News},
		\texttt{WorldNews}, \texttt{Politics}, \texttt{Technology},
		\texttt{Science}, \texttt{Futurology}, \texttt{Sports} \\
		Movies 						& IMDb and Wikipedia \\
		\hline\hline
	\end{tabular}
	\caption{Types of content posts and their sources.}
	\label{tab:content_sources}
\end{table}

Sounding Board relies on a large collection of social chat content gathered from
multiple sources.
Types of content posts and their sources are listed in
Table\,\ref{tab:content_sources}.
These content posts are stored in the knowledge base at the back-end.
We use a graph structure to represent the knowledge base, following the widely
adopted concept of a knowledge graph.
There are two types of nodes in the graph.
A {\bf content post} node can represent a fun fact, an amusing thought, a trivia
question-answer pair, etc.
A {\bf topic} node represents an entity (e.g., Amazon), 
a general noun (e.g., technology), 
a general noun phrase (e.g., artificial intelligence),
a movie name (e.g., Frozen), 
or a movie genre (e.g., fantasy movie).
The edge between one content post and a topic is labeled by the corresponding
relation.
The {\bf topic mention} edge means that the topic is explicitly mentioned in the
textual body of a content post.
The {\bf category tag} edge indicates that the topic is a tag in the
meta-information associated with the content post.
Four types of edges are created for a movie node, i.e.,
the {\bf movie name},
the {\bf movie genre}, 
the {\bf movie director}, 
and the {\bf movie actor}. 
With these edges, a path can be established from one content post node to another
via an intermediate topic node.
For example, a news headline may mention a person, who may be mentioned in some
fun facts.
This allows the bot to drive the conversation forward by either traversing
content post nodes associated with the same topic or through relation edges to
content nodes on a relevant new topic.

The knowledge graph is stored in the AWS DyanmoDB tables and automatically
updated every day to include recent news.
The construction of the knowledge graph has three stages: 
content mining, content filtering, and topic indexing.

{\bf Content Mining}:
In this stage, content posts are gathered from multiple sources and
pre-processed. 
As shown in Table\,\ref{tab:content_sources}, Reddit is a primary source of
content posts, which allows us to use subreddit-specific karma thresholds to
filter out less highly endorsed posts that are likely to be either less
interesting or controversial.
In particular, fun facts, amusing thoughts, and life pro tips are acquired from
subreddits where almost all submissions have an informative and well-formatted
title following the community rules.
Thus, we can develop subreddit-specific processing pipelines to extract the
target content span from the title.

{\bf Content Filtering}:
We notice that many Reddit posts are not suitable for family-friendly discussions or
conversational interactions.
A filtering pipeline is developed to avoid posts with unwanted patterns such as
profanity, sex, violence, etc.
We use the same profanity word list as in the utterance purification described
in \S\ref{chap3:sec:nlg}
and expand it with sensitive and unpleasant words (e.g., ``kill'',
``suicide'').
We also included patterns that were not inherently inappropriate but were often
found in potentially offensive statements (e.g., ``your mother''). 
Content that is not well-suited in style to casual conversation (e.g., URLs and
lengthy content) is either removed or simplified.
We further remove content containing characters that are not well suited to
spoken interactions (e.g., hashtags).

{\bf Topic Indexing}:
In the third stage, kept posts are indexed by topics and corresponding
edges are created for the knowledge graph.
For fun facts, amusing thoughts, life pro tips, and news headlines, the system
extracts topics from post titles.
Topic mention edges are created for these topics.
Some subreddits (e.g., \texttt{LifeProTips}, \texttt{Science},
\texttt{Technology}, \texttt{Futurology}, \texttt{Sports}) maintain their own
sets of flair tags that can be used for categorizing posts, which are either
assigned by the post authors or the subreddit moderators.
The system converts flair tags to corresponding topics
using a pre-defined mapping and creates the corresponding category tag edges
between the post and the topic node.
For news headlines, the corresponding subreddit name is mapped to a high-level
topic, and the system additionally creates category tag edges between
them.
For trivia question-answer pairs, the text body is split into the question body
and the answer.
Only the question body is used for indexing topics and creating topic mention
edges.
Similarly, a joke is split into the joke setup and punchline, and only the joke
setup part is used for indexing topics and creating topic mention edges. 
Movies are indexed by their names, genres, directors, and actors, which are
connected to the movie node via corresponding movie-related edges.

To extract topics from text, the system relies on the constituency parser and
named entity recognizer (NER) in the Stanford CoreNLP toolkit \cite{CoreNLP}.
Specifically, the system extracts all nouns and noun phrases from the
constituency parse tree and all named entities from the NER as candidate topics.
The lemmas for nouns are also included as candidate topics.
Then, the system removes invalid topics including numbers, dates, 
noun phrases with pronouns (e.g., ``his house''), 
generic nouns (e.g, ``people'', ``place''), 
and phrases that are often incorrectly detected as noun phrases 
(e.g., ``a bit'', ``a lot'').
The system merges topics with same tokenized forms as a unique topic node, using
a tokenization process that involves lower-casing and
lemmatization.

\section{Implemented Miniskills}
\label{chap3:sec:miniskills}
Miniskills implemented in Sounding Board have been listed in
Table\,\ref{tab:miniskills}.
In this section, we provide details about these miniskills.

\subsection{General miniskills}
\label{chap3:ssec:general_miniskills}
The following four general miniskills are used to set up the context of the
conversation, i.e., letting the user have an impression about what the bot can
and cannot do.

{\bf Greeting}:
This miniskill is used at the opening and is only used once in a conversation.
It initiates the conversation with a how-are-you question 
and empathizes with the user's answer accordingly.

{\bf Exit}:
This miniskill instructs the user to explicitly say ``{\it stop}'' to exit
Sounding Board.
Depending on the invocation condition, Sounding Board adds different grounding
dialog acts before the instruction.
When the miniskill is invoked because the user issues a command for invoking an
activity that is not yet implemented by Sounding Board (e.g., play musics, read
books), the added dialog act explains the limitation of the bot.
When the miniskill is invoked because the user makes an implicit stop command
(e.g., ``{\it good night}'', ``{\it I need to leave now}''), the added dialog
act thanks the user for chatting.

{\bf List Activities}:
This miniskill introduces Sounding Board's available conversation activities.
It acts as a menu for Sounding Board.

{\bf List Topics}:
This miniskill negotiates with the user about the chat topic by listing several
top-ranking topics we curated based on our collected conversation data and
available content.

\subsection{Content-oriented miniskills}
\label{chap3:ssec:content_miniskills}

The major activity in Sounding Board is discussing topics.
Thus, content-oriented miniskills are critical to the success of Sounding Board.
Sounding Board acquires content from multiple sources as summarized in
Table\,\ref{tab:content_sources}.
The specific sources are chosen because they provide news or commentary of broad
interest and in a style that is reasonably well suited to spoken conversations.
In addition, since individual exchanges need to be relatively short, we choose
sources for which it is easy to extract snippets of information that are
informative and require little context to understand.

{\bf Ask Trivia Questions}:
During the conversation, the bot asks factual questions about the current
discussion topic and provides the answer after the user's response.
We extract trivia question-answer pairs from the recent popular SQuAD dataset
\cite{SQUAD2016}.
After filtering, we use around 23K question-answering pairs covering around 600
topics indexed from the question body.

{\bf Tell Fun Facts}:
We crawl a large collection of fun facts from the \texttt{TodayILearned}
subreddit.
Most posts in this subreddit have an informative and well-formatted title.
We index these posts by all possible topics appeared in the title.
Finally, around 30K posts are kept with around 186K indexed topics.

{\bf Tell Amusing Thoughts}:
Opinion-oriented content is another popular content type in Alexa Prize
conversations.
We initially tried to use the \texttt{ChangeMyView} subreddit, which turns out
to be not suitable for casual conversations since the opinion pieces there tend
to be serious, controversial, and political.
Therefore, we turn to the other subreddit \texttt{ShowerThoughts} where people
share amusing thoughts.
Similar to the \texttt{TodayILearned} subreddit, post titles in
\texttt{ShowerThoughts} usually do not require additional context to understand.
From our test user feedback, the use of \texttt{ShowerThoughts} provides a much
more engaging conversation than \texttt{ChangeMyView}.
In total, around 24K posts are kept contributing to around 117K indexed topics.

{\bf Tell Life Pro Tips}:
We add this miniskill for using posts from \texttt{LifeProTips} to increase the
diversity of the conversation.
The \texttt{LifeProTips} subreddit is another popular subreddit where people
share specific tips that improve people's daily life.
Similar to \texttt{TodayILearned} and \texttt{ShowerThoughts}, all information
is contained in the title and no additional context is necessary.
Although some tips can be interesting and useful, the percentage of boring posts
seems to be higher than \texttt{TodayILearned} and \texttt{ShowerThoughts} based
on manual inspection.
In total, we keep around 1K posts for around 9K topics.

{\bf Tell Jokes}:
Sounding Board uses jokes cached from Amazon Evi~\cite{Evi}.
In total, we cached 452 jokes with around 2K indexed topics.

{\bf Discuss News Headlines}:
To enable a conversation segment for news discussion, we crawl multiple
subreddits every day to dynamically update the bot with latest popular news.
These subreddits have posts with the original or user-modified headline for news
articles.
Currently, the bot only presents the headline and chooses a person entity or an
organization entity mentioned in the headline as a digression topic.  
The bot then uses other miniskills (e.g., {\sl Tell Fun Facts}) to push the
conversation forward on the digression topic.
Clearly, an advanced miniskill that is able to interactively tell and
discuss the news story is desirable.
This issue was addressed after the competition and is discussed in
Chapter~\ref{chap5}.

{\bf Discuss Movies}:
We implement this miniskill specifically for movie discussions for two
reasons:
1) movies are among the most popular entertainment topics,
and 2) it is relatively easy to obtain meta information about a movie from IMDb
and Wikipedia.
The miniskill manages a structured conversation segment about a specific movie.
During the conversation segment, the bot interactively discusses the movie's
name, year, genre, plot, reviews, and famous cast members.
The bot also tries to understand the user's knowledge and comments about the
movie, its director, and some chosen cast members.
After getting the user comments on a chosen cast member, the bot makes a
movie recommendation (without initiating a discussion on the recommended movie)
by choosing a different movie where the person is also a cast
member.
Additionally, this miniskill makes extensive use of facts, thoughts, jokes, and
news headlines crawled for other miniskills when a cast member is mentioned.

\subsection{Personality assessment}
\label{chap3:ssec:personality}
\begin{table}[!t]
	\centering
	\begin{tabular}{l|c|c}
		\hline\hline
		{\bf Question} & {\bf Yes} & {\bf No} \\
		\hline
		Extroversion & & \\
		\tabitem {\it Would you say you are often the life of the party?} 
		& $+$ & $-$ \\
		\tabitem {\it Do you talk to a lot of different people at parties?} 
		& $+$ & $-$ \\
		\tabitem {\it Do you usually talk a lot?}
		& $+$ & $-$ \\
		\tabitem {\it When you hang out with people, do you usually keep in the
		background?} 
		& $-$ & $+$ \\
		\hline
		Neuroticism  & & \\
		\tabitem {\it Do you have frequent moodswings?} 
		& $+$ & $-$ \\
		\tabitem {\it Are you relaxed most of the time?} 
		& $-$ & $+$ \\
		\tabitem {\it Do you often feel blue?} 
		& $+$ & $-$ \\
		\tabitem {\it Do you get upset easily?} 
		& $+$ & $-$ \\
		\hline
		Openness & & \\
		\tabitem {\it Do you have a vivid imagination?}
		& $+$ & $-$ \\
		\tabitem {\it Do you feel like you do not have a good imagination?}
		& $-$ & $+$ \\
		\tabitem {\it Do you usually have difficulty understanding abstract ideas?}
		& $-$ & $+$ \\
		\tabitem {\it Are you usually interested in abstract ideas?}
		& $+$ & $-$ \\
		\hline
		Conscientiousness & & \\
		\tabitem {\it Do you usually like order?}
		& $+$ & $-$ \\
		\tabitem {\it Do you often forget to put things back in their proper place?}
		& $-$ & $+$ \\
		\tabitem {\it Do you usually get chores done right away?}
		& $+$ & $-$ \\
		\tabitem {\it Do you often make a mess of things?}
		& $-$ & $+$ \\
		\hline
		Agreeableness & & \\
		\tabitem {\it Would you say you're not really interested in people's problems?}
		& $-$ & $+$ \\
		\tabitem {\it Do you sympathize with other people's feelings?}
		& $+$ & $-$ \\
		\tabitem {\it Are you usually not interested in other people?}
		& $-$ & $+$ \\
		\tabitem {\it Do you usually feel other people's emotions?}
		& $+$ & $-$ \\
		\hline\hline
	\end{tabular}
	\caption{Personality assessment questions.}
	\label{tab:personality_questions}
\end{table}

The {\sl Ask Personality Questions} miniskill calibrates user personality based
on the Five Factor model \cite{mccrae1992introduction} through exchanging
answers on a set of personality probing questions adapted from the mini-IPIP
scale \cite{donnellan2006mini}.
Questions used by Sounding Board for individual traits are listed in
Table\,\ref{tab:personality_questions}. 
The bot maintains a score for each trait, initialized as zero.
For each question, the bot increments or reduces the score for the corresponding
trait by one depending on the user answer.
The bot only asks a subset of these questions each time the miniskill is invoked. 
Users who answer the personality questions will subsequently get topics tailored
to their personality based on findings from \cite{Schwartz2013PLOS}.

To make the miniskill more engaging, Sounding Board uses three strategies. 
First, for each question, the bot provides its own answer after receiving the
user's response.
We find through test user feedback that having its own answers greatly enhances
the user experience.
Second, we interleave personality questions with hand-made fun questions
designed to keep the conversation engaging.
Before terminating the miniskill, the bot assigns the user to a Disney
character using a pre-defined mapping based on the scores on individual
personality traits, e.g., ``{\it you remind me a lot of Elsa from the movie
Frozen.}''

\section{A Sample Conversation}
\label{chap3:sec:sample_conversation}
\begin{table}
	\centering
{\small
	\begin{tabular}{c|l|p{13cm}}
		\hline\hline
		{\bf Turn} & {\bf Speaker} & {\bf Utterance} \\ 
		\hline
		1 & User 	& {\it let's chat} \\
		2 & Bot 	& {\it Hi! This is an Alexa Prize socialbot. How's your day?} \\
		3 & User 	& {\it I'm \underline{five}} \\
		4 & Bot 	& {\it Glad to hear it! We could talk about robots, science, or
		sports. 
		} \\
		\hline
		5 & User 	& {\it Superman} \\
		6 & Bot 	& {\it It looks like you wanna chat about Superman. I
			was high up in the cloud when I realized: If
			people don't recognize Clark Kent as Superman
			because of his glasses, does his eye doctor know
		his true identity?} \\
		7 & User  & {\it I guess so} \\
		8 & Bot 	& {\it Did you know that Henry Cavill almost missed the
			call for the role of Superman cause he was
		playing ``World of Warcraft''?} \\
		9 & User 	& {\it really I didn't know that} \\
		10 & Bot 	& {\it Weird, right? Speaking of this, 
			did you watch the movie: Superman?
		It was released in 1997. It's a comedy.} \\
		11 & User & {\it yes it was hilarious} \\
		12 & Bot 	& {\it I'm glad you feel this is hilarious. Which part do you like
		best about this movie?} \\
		13 & User & {\it the part when he met \underline{Lewis Leah}} \\
		14 & Bot 	& {\it Interesting. Meccartin and Raffi co-directed this film. The
		movie has a 6.3 out of 10 on IMDb, which seems pretty good! Do you like the
		movie's director?} \\
		\hline
		15 & User & {\it yes but let's chat about something else i want to talk
		about science} \\
		16 & Bot 	& {\it I love it! Did you see this article from Last Friday?
			The title was: Deep learning neural networks power self-driving cars,
			predictive policing, and medical diagnoses. 
			} \\
		17 & User & {\it no i didn't} \\
		18 & Bot 	& {\it Okay. This news mentioned car. Would you like to
			hear some advice 
			about that?} \\
		19 & User & {\it yes} \\
		20 & Bot 	& {\it Great! My friend gave me this advice. If you get into an
			accident, try to take note of the plate of at least one other car which
		might have seen it occur.} \\
		21 & User & {\it that's true} \\
		22 & Bot 	& {\it Looks like you think that it makes sense. I heard this from
			my friend: where does the dog leave his car when he shops?  
			The barking lot.} \\
		23 & User & {\it alexa how many stars in the universe?} \\
		24 & Bot 	& {\it The universe has 70,000,000,000,000,000,000,000 stars. Do
		you still wanna to chat about science?} \\
		\hline
		25 & User & {\it no thank you let's talk about celebrities} \\
		26 & Bot 	& {\it No problem! How about Beyonc\'{e}? My friend asked me this
		question. What's the first year that Beyonc\'{e} appear on the Time 100 list?} \\
		27 & User & {\it two thousand and two} \\
		28 & Bot 	& {\it The answer is 2013. Do you still want to chat about
			Beyonc\'{e}?} \\
		29 & User & {\it no thanks stop} \\
		\hline\hline
	\end{tabular}
}%
	\caption{A sample conversation between the user and Sounding Board.
		The conversation starts with a small talk subdialog (turns 1--4),
		followed by subdialogs on three user-requested topics: Superman (turns
		5--14), science (turns 15--20), and celebrities (turns 21--25).
		Suspected speech recognition errors in the user utterances are underlined.
	}
	\label{tab:sample_conversation}
\end{table}

In this section, 
a sample conversation is shown in Table\,\ref{tab:sample_conversation}, which is
a combination of conversation segments from real users.
The conversation starts with the user's ``{\it let's chat}'' invocation and ends
with the ``{\it stop}'' command.
Early in the conversation, turn 2 is produced using the {\sl Greeting}
miniskill. 
The {\sl List Topics} miniskill is used at turn 4.
For brevity, the {\sl Exit}, {\sl List Activities} and 
{\sl Ask Personality Questions} miniskills are not included in this example.

In Table\,\ref{tab:sample_conversation}, the user-centric strategy is reflected
in several aspects.
First, the user has the freedom to take the initiative in the conversation.
For example, at turn 15 and turn 21, the user interrupts the conversation flow
and switch the topic.
Similarly, at turn 19, the user ignores the bot confirmation and asks a
question.
Second, the bot responses usually start with acknowledging user requests or
reactions, as can be seen at 
turn 4 (``{\it Glad to hear it!}''), 
turn 6 (``{\it It looks like you wanna chat about Superman}''),
turn 12 (``{\it I'm glad you feel this is hilarious.}''),
turn 22 (``{\it Looks like you think that it makes sense}''),
etc. 
Lastly, the dialog control is influenced by the user's topic interests.
For example, during turns 5--14, the user seems to be engaged in the conversation
about Superman, and thus the bot keeps pushing the conversation forward on this
topic.
At turn 20, the bot decides to confirm whether to continue on the previous topic
``science'' since the user interrupts the conversation flow which could be a
sign of losing interest in the topic.

The content-driven strategy is obvious from the example, i.e., the bot
actively pushes the conversation forward by suggesting topics and interesting
content.
The {\sl Tell Amusing Thoughts} miniskill is used at turns 6.
Turn 8 uses the {\sl Tell Fun Facts} miniskill. 
Turns 10--14 use the {\sl Discuss Movies} miniskill.
At turn 14, the bot offers to briefly discuss the movie's director, but is
turned down by the user at turn 15.
The {\sl Discuss News Headline} miniskill is used at turn 16, followed by a
briefly discussion on the topic ``car'' using the {\sl Tell Life Pro Tips}
miniskill at turn 20 and the {\sl Tell Jokes} miniskill at turn 22.
At turn 24, the bot provides the answer obtained from Amazon Evi for the user
question in turn 23.
At turn 26 and turn 28, the {\sl Ask Trivia Questions} miniskill is used.

Lastly, it can be seen that there are speech recognition errors in the
conversation since the socialbot relies on the Alexa Prize's speech recognizer
to automatically transcribe user utterances.
At turn 3, the hypothesis with the highest confidence score is ``{\it I'm
five}''.
Since the speech recognizer produces multiple alternative hypotheses with
slightly lower confidence scores, including the actual user utterance ``{\it I'm
fine}'', the bot can correctly acknowledge the user at turn 4.
Another speech recognition error occurs at turn 13, where the name 
``{\it Lewis Leah}'' is likely to be ``{\it Lois Lane}''.
In this case, although none of the hypotheses is correct, the bot pushes the
conversation forward as if it understood the user utterance.  


\chapter{Multi-Level Evaluation for Socialbot Conversations}
\label{chap4}
The Alexa Prize has provided an opportunity for getting real user interactions
and ratings at a very large scale, but challenges remain for evaluating
socialbot conversations.
First, recall that the Alexa Prize frames the socialbot as an open-domain
conversational system that can talk about a wide range of topics in a coherent
and engaging way as long as possible.
This framing of a conversational system calls for understanding the mechanisms
of socialbot conversations and aspects that influence user ratings.
Second, users only give a single subjective rating at the end of the
conversation, and many conversations are unrated.
Prior studies have tried to find interpretable and automatic metrics that
measure the quality of socialbot conversations, but the results are mixed.
The number of turns seems to be a strong baseline as no other model-free metric
has shown to have higher correlation with conversation-level user ratings.
However, judging a conversation simply by the number of turns or conversation
duration alone has the potential problem of encouraging redundant responses
which may turn out to be annoying.
Third, a conversation can contain both good and bad segments, and a
conversation-level rating does not uniformly apply to individual segments.
Therefore, an automatic segment-level scoring method would be useful to identify
bad dialog segments and potential system issues such as dialog policies and
content sources.

In this chapter, we first review related work on dialog system evaluation
approaches and models of dialog structure in \S\ref{chap4:sec:related_work}.
The data used in this work is described in \S\ref{chap4:sec:data}.
Then, in \S\ref{chap4:sec:factors}, we propose two sets of conversation acts for
tagging user and bot turns, respectively.
These conversation acts are examined in terms of the correlation of the number
and the percentage of corresponding turns with conversation-level user ratings
to identify the most useful features for evaluation.
Linear regression models are also learned to combine these
conversation-act-based metrics to predict conversation-level user ratings.
In \S\ref{chap4:sec:multi_level}, we describe a novel dialog segmentation model
to segment socialbot conversations.
Then, two multi-level scoring models are proposed to predict scores of
different types of dialog segments.
Experiments are carried out to examine the performance of the learned scoring
models in \S\ref{chap4:sec:experiments}, where we also demonstrate ways to
use the predicted segment scores for system diagnosis in three case studies.
Finally, we conclude this chapter with a summary and discussion of implications
in \S\ref{chap4:sec:discussion}.

\section{Related Work}
\label{chap4:sec:related_work}
Research on socialbot evaluation has been discussed in
\S\ref{chap2:sec:evaluation}.
In this section, we review prior work on evaluation approaches for other types
of dialog systems in \S\ref{chap4:ssec:dialog_evaluation}.
Then, prior work on models of dialog structure is discussed in
\S\ref{chap4:ssec:dialog_model}.
Finally, we discuss the uniqueness of our work in
\S\ref{chap4:ssec:discussion_related_work}.

\subsection{Dialog system evaluation approaches}
\label{chap4:ssec:dialog_evaluation}
With increasing interests in open-domain chatbots (also known as
non-task-oriented or chat-oriented systems), evaluation methods for these
systems have become an important research topic.
One way to evaluate bot responses relies on human ratings from either experts
and crowdsourced workers.
In this framework, bot responses are usually annotated based on coherence or
appropriateness.
For example, a three-class (valid, acceptable, invalid) annotation scheme is
designed for annotating bot utterances in the WOCHAT datasets
\cite{Banchs2016WOCHAT}.
In \cite{Yu2016WOCHAT}, individual bot turns are annotated with three levels
of appropriateness (inappropriate, interpretable, appropriate).
User self-reported ratings are also used for evaluation in \cite{Yu2016WOCHAT},
including both turn-level ratings of bot responses and conversation-level
user satisfaction or engagement, but the authors do not study the correlation
between turn-level appropriateness and conversation-level ratings.
In \cite{Devault2011Sigdial,Gandhe2014IWSDS}, human votes on a set of bot
responses generated by different policies are collected by asking a person to
play the role of the bot while another person plays the role of the user.
The statistics of human votes are used to evaluate different dialog policies.

Reference-based evaluation is widely used in recent neural response generation
models
\cite{Sordoni2015NAACL,Shang2015ACL,Vinyals2015ICML,Serban2016AAAI,Li2016NAACL,Li2016ACL,Luan2017IJCNLP,Ghazvininejad2018AAAI}.
In \cite{Vinyals2015ICML}, the perplexity is used to compare neural response
generation models.
Popular unsupervised reference-based metrics include BLEU~\cite{BLEU2002},
deltaBLEU~\cite{Galley2015ACL}, METEOR~\cite{Meteor2014}, and ROUGE
\cite{Rouge2008}, which all measure the similarity between individual bot
responses and their corresponding reference responses.
However, these metrics cannot account for the fact that responses with
completely different meanings can be equally acceptable, and it has been shown
that they have very weak correlation with turn-level human ratings
\cite{Liu2016EMNLP}.
Reference-based methods are particularly problematic in longer conversations,
since an acceptable bot response can take the conversation in different
directions, for which subsequent references are not relevant.

Model-based evaluation has also been proposed for open-domain chatbots when
human annotations are available.
A neural network model is proposed in \cite{Lowe2017ACL} to estimate human
ratings of candidate bot responses using distributed representation of dialog
context, the candidate bot response, and the reference response.
This model can be viewed as a reference-based metric since it uses 
reference responses as one of the input, but it is a supervised model rather
than an unsupervised metric such as BLEU. 
Without using references, in \cite{Higashinaka2014Interspeech},
turn-level coherence is annotated for individual bot utterances and a binary
classification model is built using the coherence label and a set of
hand-crafted features including dialog acts, question types, predicate-argument
structure, named entities, and dependency parsing structure.
A model is developed in \cite{Xiang2014Problematic} by training on human
annotations of problematic turns to automatically recognize such turns using
intent and sentiment features.
Using the annotations in the WOCHAT datasets \cite{Banchs2016WOCHAT}, several
binary classification models are compared in \cite{Yuwono2018IWSDS} for
estimating whether a bot utterance is valid.
Yu et el.\ \cite{Yu2016Strategy} train models using hand-crafted features and
pre-trained word and document embeddings to make binary appropriateness
predictions (inappropriate vs.\ appropriate/interpretable) for individual bot
turns and binary conversation depth estimations (deep vs.\ shallow/intermediate)
for complete conversations.
The model-based methods discussed here all focus on evaluating individual bot
turns given the corresponding previous user utterance.
They do not evaluate the conversation as a whole, and it remains unclear how
these turn-level metrics are correlated with with conversational-level human
ratings.

Several model-free and reference-free metrics are used in the literature as
well.
In \cite{Yu2016Strategy}, the information gain metric is proposed to estimate
the amount of information in the conversation by counting the number of unique
words in both bot and user utterances.
To demonstrate the advantage of a proposed diversity-promoting objective
function based on maximum mutual information for training neural response
generation models, Li et al.\ \cite{Li2016NAACL} use the number of distinct
unigrams and bigrams in generated responses as a metric that measures the degree
of diversity.
This diversity proxy has also been used in \cite{Zhang2018NIPS} to compare
neural responses models with the proposed adversarial information maximization
model.
However, these metrics are usually used for comparing different dialog policies,
and the authors do not study whether they are correlated with human
ratings.

Finally, to combine reference-based and model-based metrics, a hybrid
metric named RUBER is developed in \cite{Tao2018AAAI} that combines a metric
calculating the similarity between the bot utterance and the reference
based on pre-trained word embeddings, with an appropriateness score of
the bot response with respect to the user utterance predicated by a neural
network trained on pairs of user utterances and their reference bot responses.
In \cite{Tao2018AAAI}, correlations with human ratings are studied at the turn
level, not the conversation level, since this method only takes into account the
bot utterance and its preceding user utterance rather than the complete dialog
context.

A line of research has applied reinforcement learning to training or improving
dialog systems, mostly focused on task-oriented systems (e.g.,
\cite{Singh2002JAIR,Young2013IEEE}).
The reward function in the reinforcement learning framework can be treated as an
evaluation metric.
For open-domain chatbots, Li et al.\ \cite{Li2016EMNLP} use a hand-crafted
reward function that combines scores measuring the ease of answering,
information flow, and semantic coherence, which are all estimated based on
neural response generation models trained on a conversation corpus without human
annotations.
Yu et al.\ \cite{Yu2016Strategy} also use a hand-crafted reward function, but
using scores predicted from models learned on human annotations of turn-level
appropriateness and conversation depth.
Neither has studied whether the ratings estimated by the reward function
are correlated with human ratings.
For socialbots, MILA \cite{Serban2017arXiv} learns a linear regression model as
the reward function, training on user ratings using 23 features
including dialog length, features that characterize the bot utterance
(appropriateness, genericness, length), and features that characterize the user
utterance (sentiment, genericness, length, confusion).
However, they do not analyze how individual features are correlated with
conversation-level user ratings, since their focus is to use the reward function
for training the socialbot under the reinforcement learning framework.

Unlike open-domain chatbots which usually do not have a well-defined task,
task-oriented systems are mostly evaluated by task completion, but there are
also a verity of other evaluation approaches.
Notably, PARADISE is a widely adopted model-based evaluation framework which
focuses on
using task cost and success metrics to estimate overall user ratings via a
linear regression model \cite{Walker1997ACL}.
It has been used for conversation-level user rating prediction
\cite{Walker2000NLE}, as well as optimizing a dialog policy under a reinforcement
learning setting \cite{Rieser2011CL}.
A questionnaire-based subjective evaluation approach is suggested
in~\cite{Hone2000NLE,Hone2001Eurospeech}.
The MeMo framework integrates user simulations into a workbench that predicts
usability ratings via a linear regression model~\cite{Moller2006INTERSPEECH}.



\subsection{Models of dialog structure}
\label{chap4:ssec:dialog_model}
In \cite{UCSCSlugbot2018}, traditional models of dialog structure in
conversational AI systems are categorized into three types.
The search model treats a conversation as a sequence of user-initiated search
queries and bot-retrieved search results;
the task model suggests a conversation is composed of task sequences;
and the script model assumes a conversation follows a finite-state script which can
be written manually to support all kinds of conversations a user might want to
have with the system.
As pointed in \cite{UCSCSlugbot2018}, since these models are developed for other
types of dialog systems, it is not obvious how they can be used for socialbot
conversations.
To address this issue, a dialog model is proposed in \cite{UCSCSlugbot2018}
that uses four high-level discourse relations (expansion, comparison,
contingency, temporal) from the Penn Discourse TreeBank \cite{PDTB2008} to model
the coherence of open-domain socialbot conversations.

Several studies have attempted to model the structure of human-human casual
conversations.
An early study \cite{Ventola1979Pragmatics} has suggested that a conversation
develops a sequence of structural elements including 
greeting (e.g., ``{\it how are you}''),
address that defines the addressee and indicates the relationship between the
speakers (e.g., ``{\it excuse me, sir}''),
approach that is basically smalltalk (e.g., ``{\it how's the family}''),
centering which may cover a wide range of topics,
leave-talking that expresses the desire or need to end conversation (e.g.,
``{\it I must be off now}''), 
and goodbye that concludes the conversation (e.g., ``{\it see you}''). 
Another view \cite{Eggins2005} is that a conversation consists of 
chat segments that are highly interactive and chunk segments where one speaker
takes the floor and dominates the conversation for an extended period.
This binary distinction of segments has been recently used in
\cite{Gilmartin2019IWSDS} to annotate multi-party casual conversations as well.
In \cite{Collins2016LREC}, conversations from the Switchboard corpus
\cite{SwitchboardCorpus} have been annotated based on a proposed taxonomy of
stories for analyzing the narrative structure of casual conversations.
Along the line of speech act theory \cite{Austin1962,Searle1969}, the
Switchboard corpus has also been annotated by a dialog act tag set proposed in
\cite{Jurafsky1997SWDA}, which has facilitated understanding the discourse
structure of conversations.
Several statistical dialog models have been developed on this annotated dataset
since
then~\cite{Stolcke2000CL,Kal2013RCNN,Tran2017ACL,Tran2017EACL,Tran2017EMNLP,Ji2016NAACL,Liu2017EMNLP,Lee2016NAACL}.

\subsection{Discussion}
\label{chap4:ssec:discussion_related_work}

For socialbot evaluation, some metrics have been studied as automatic measures
of the conversation quality as discussed in \S\ref{chap2:sec:evaluation}.
However, none of them achieve higher correlation with conversation-level user
ratings than the number of turns.
Furthermore, while a variety of metrics have been used for open-domain chatbot
evaluation in the literature, there is limited study of how these metrics are
correlated with user ratings, probably due to the lack of large scale
conversation data.
Model-based evaluation approaches have shown some promising results of using
hand-crafted features and/or distributed representations to predict human
ratings.
Nevertheless, there is little in-depth analysis of how different features
contribute to the human ratings.
Moreover, these models mostly evaluate individual bot turns (e.g.,
\cite{Lowe2017ACL}).
The relation between conversation-level ratings and turn-based metrics
calculated for bot utterances remains unclear.
Additionally, none of these studies have investigated how user reactions at
individual turns impact conversation-level ratings.
In this work, we carry out in-depth analysis of several metrics that
characterize both bot and user turns by a set of proposed conversation acts.
Through correlation analysis on these metrics and conversation-level ratings,
we provide some insights on how different bot behaviors and user reactions
contribute to conversation-level user ratings.
We also show that some of these metrics have higher correlation with
conversation-level user ratings than the number of turns in the
conversation.

Furthermore, most prior work has focused on evaluating individual turns or
entire conversations.
Although several models of dialog structure have observed that human-human
casual conversations behave like a sequence of structural segments, dialog
segment evaluation is an understudied problem.
In this work, using a model of socialbot conversations that divides a
conversation into topic-based segments, we investigate methods that can be used
to score conversations at both conversation and segment levels.

\section{Data}
\label{chap4:sec:data}
The analyses and experiments in this chapter make use of human-socialbot
conversations collected from a stable version of Sounding Board described in
Chapter~\ref{chap3}, running for one month with a consistent dialog policy, but
with up-to-date content dynamically added to the back-end.
From manual inspection, most short conversations are accidentally started
because of incorrect invocation detection.
Thus, we only consider conversations with at least three user turns, of which
68,453 (43\%) are rated.

\begin{table}[t]
	\centering
	\begin{tabular}{c|c|c|c}
		\hline\hline
		user rating & duration (min) & \# user turns & \# topics \\
		\hline
		3.65 $\pm$ 1.40
		& 4:30 $\pm$ 5:18
		& 22.3 $\pm$ 25.4
		& 8.4 $\pm$ 8.3 \\
		\hline\hline
	\end{tabular}
	\caption{Statistics of the rated conversations in our dataset.}
	\label{tab:data_stats}
\end{table}

Conversation statistics are given in Table\,\ref{tab:data_stats}.
The user ratings are on a scale of 1 to 5.
It can be seen that there is great variability in user interactions,
which agrees with observations in other studies on
socialbot conversations~\cite{Venkastech2017NIPS}.
The data is randomly split into training, development and test sets with a ratio
of 3:1:1.

\section{Conversation Acts in Socialbot Conversations}
\label{chap4:sec:factors}
We posit that a socialbot conversation can have good and bad turns, which can
contribute to the conversation-level user rating in opposite directions. 
With this motivation in mind, we first propose two sets of conversation acts 
in \S\ref{chap4:ssec:user_acts}--\ref{chap4:ssec:bot_acts} to tag user turns
and bot turns, respectively.
Through correlation analysis using conversation-level user ratings, the
number and percentage of turns for individual conversation acts are compared in
\S\ref{chap4:ssec:correlation_analysis} as candidate metrics that provide
indicators of positive and negative interactions.
In \S\ref{chap4:ssec:regression_analysis}, we investigate the
performance of a model-based evaluation method that uses linear regression
models with metrics in \S\ref{chap4:ssec:correlation_analysis} as input
features.
Finally, implications and limitations of our analysis are discussed in
\S\ref{chap4:ssec:implications}.

Note the term ``conversation act'' is coined in \cite{Traum1992ConversationActs}
for task-oriented spoken dialogs, which encompasses not only traditional speech
acts but also turn-taking, grounding, and higher-level argumentation acts.
In the literature, several similar terms are used with slight differences in
their definitions, including speech acts, dialog acts, communicative acts,
conversational moves, and dialog moves (see \cite{Traum2000Semantics} for
discussions).
In this chapter, with a slight abuse, we use conversation acts to encode user
and bot actions in the conversation.

\subsection{Conversation acts for user turns}
\label{chap4:ssec:user_acts}

\begin{table}[t]
	\centering
	\begin{tabular}{l|l}
		\hline\hline
		{\bf Conversation Act} & {\bf Description} \\
		\hline
		\texttt{AskQuestion} 							& ask an information/opinion-seeking question \\
		\texttt{RequestHelpOrRepeat} 			& ask for help messages or the last bot utterance \\
		\texttt{ProposeTopic} 						& propose a(n) topic/activity \\
		\texttt{AcceptTopic} 							& accept the topic/activity proposed by the bot \\
		\texttt{RejectTopic} 							& reject the topic/activity proposed by the bot \\
		\texttt{FollowAndNonNegative} 		& agrees with the dialog flow with non-negative reaction \\
		\hline
		\texttt{InterestedInContent} 			& express interest in discussing the content \\
		\texttt{NotInterestedInContent} 	& express no interest in discussing the content \\
		\texttt{PositiveToContent} 				& express positive opinion about the content \\
		\texttt{NegativeToContent} 				& express negative opinion about the content \\
		\texttt{PositiveToBot} 						& express positive opinion about the bot or its maker \\
		\texttt{NegativeToBot} 						& express negative opinion about the bot or its maker \\
		\hline\hline
	\end{tabular}
	\caption{Conversation acts for user turns. A turn can have multiple
	conversation acts.}
	\label{chap4:tab:user_dialog_acts}
\end{table}

Through manual inspection of our data, we propose several user conversation acts in
Table\,\ref{chap4:tab:user_dialog_acts} that may be relevant to user engagement
and satisfaction.
Recall that current speech recognition hypotheses provided to the bot do not
contain punctuation or segmentation information.
Further, a user utterance may express multiple conversation acts in one turn
explicitly or implicitly.
Therefore, we do not require the mutually exclusivity for these conversation acts.
For example, ``{\it this is boring how about we talk about cats}'' can be tagged
by \texttt{NotInterestedInContent}, \texttt{NegativeToContent} and
\texttt{ProposeTopic}.

{\bf System-Log-Based Conversation Act Tagging}:
For the first six conversation acts in Table\,\ref{chap4:tab:user_dialog_acts}, it is
straightforward to automatically tag user turns using the language understanding
output of the system.
In particular, \texttt{AskQuestion}, \texttt{RequestHelpOrRepeat}, and
\texttt{ProposeTopic} are mapped from a subset of intents used by the system
(see \S\ref{chap3:ssec:intent_analysis}),
whereas \texttt{AcceptTopic} and \texttt{RejectTopic} rely on the user reaction
analysis results (see \S\ref{chap3:ssec:user_reaction_analysis}).
The \texttt{FollowAndNonNegative} act is specially designed to capture user
engagement and satisfaction, which means the user agrees with the current dialog
flow without expressing negative reaction.
In other words, the user does not initiate a topic switch either explicitly or
implicitly.
It covers cases where the user accepts a bot proposal, 
answers an information/opinion-seeking question,
instructs the bot to continue, 
asks an information/opinion-seeking question about the content,
and makes a neutral or positive comment about the content.
A hand-crafted decision tree is developed for \texttt{FollowAndNonNegative}
based on these criteria using the language understanding output.

\begin{table}[t]
	\centering
{\small
	\begin{tabular}{l|ccc|c|c}
		\hline\hline
		{\bf Question} 
		& pos & neg & unk 
		& {\bf $\kappa$} 
		& {\bf Acc.} \\
		\hline
		Is the user interested in discussing the information presented by the bot? 
		& .51 & .29 & .22
		& .51 
		& .58 \\
		What is the user's opinion about the information presented by the bot? 
		& .37 & .15 & .48
		& .59 
		& .75 \\
		What is the user's opinion about the bot or its maker? 
		& .12 & .03 & .85 
		& .56 
		& .85 \\
		\hline\hline
	\end{tabular}
}%
	\caption{Annotation questions for user conversation acts, associated with
	their label distributions, inter-annotator agreement, and model accuracy.}
	\label{chap4:tab:user_dialog_act_annotation}
\end{table}

{\bf Model-Based Conversation Act Tagging}:
The last six conversation acts in Table\,\ref{chap4:tab:user_dialog_acts} are beyond
the granularity of the language understanding module in the system.
Therefore, we collect human annotations and build tagging models for them.
Since all these conversation acts are reactions to content, we only annotate user
responses to bot utterances that present a content post. 
During the annotation, workers are asked to read an bot utterance and the
subsequent user utterance, and then answers the three multiple choice questions
in Table~\ref{chap4:tab:user_dialog_act_annotation}.
Three choices are provided, i.e., positive, negative, and unknown.
The unknown option is provided to indicate that the user utterance does not
express such information or it is hard to tell.
In total, 330 unique user utterances are annotated through the FigureEight
platform.\footnote{\url{https://www.figure-eight.com}}
Each utterance gets 5 judgments per question, and we use the majority voted
answer as the label.
Using the collected annotations, we train a three-class support vector machine
(SVM) for each question.
All models use the number of tokens, bag-of-words, and sentiment polarity as
features, where the sentiment polarity is predicated by the Stanford CoreNLP
toolkit~\cite{CoreNLP}.
These SVMs are trained with 3-fold cross-validation on 75\% annotated data and
evaluated on the remaining 25\%.
Label distributions, inter-annotator agreement (Fleiss's $\kappa$), and
classifier accuracy are reported in Table\,\ref{chap4:tab:user_dialog_act_annotation}.

\subsection{Conversation acts for bot turns}
\label{chap4:ssec:bot_acts}

Our view of socialbot conversations is in line with the study in
\cite{Ventola1979Pragmatics} for human-human casual conversations, i.e.,
treating the conversation as a sequence of structural segments.
Specifically, since the system tracks the current topic of the conversation, we
are able to identify individual topic segments in the conversation.
We further say a multi-turn segment that discusses a topic has three stages,
i.e., negotiation, discussion, and termination.
During the negotiation stage, the bot proposes a topic and checks if the user
wants to talk about it.
Upon user agreement, the segment enters its discussion stage and the bot starts
to present relevant content on this topic.
The segment may also directly enter the discussion stage without the negation
stage when the user initiates a topic which the bot is able to talk about.
During the discussion stage, the bot may signal understanding failure, provide
help messages, or repeat its last utterance according to its the dialog control
logic.
Before terminating the discussion on the topic and/or switching to another
topic, the bot may make a confirmation turn to check if the user is still
interested in the current topic.

\begin{table}[t]
	\centering
	\begin{tabular}{l|l}
		\hline\hline
		{\bf Conversation Act} & {\bf Description} \\
		\hline
		\texttt{NegotiateTopic} 				& negotiate topics/activities \\
		\texttt{ConfirmToContinue} 			& confirm to continue the current topic/activity \\
		\texttt{InstructOrRepeat} 			& provide help messages or repeat the last
		bot utterance \\
		\texttt{SignalNonUnderstanding} & apologize for understanding failure \\
		\texttt{DiscussTopic} 					& discuss a specific topic or carry out a
		specific activity \\
		\hline\hline
	\end{tabular}
	\caption{Mutually exclusive conversation acts for bot turns.}
	\label{chap4:tab:bot_dialog_acts}
\end{table}

Based on our view of a socialbot conversation, we propose five bot conversation acts as
described in Table\,\ref{chap4:tab:bot_dialog_acts}.
Unlike user conversation acts, these bot conversation acts are mutually exclusive due to the
system design.
The bot turns can be automatically tagged by these conversation acts based on the
dialog state maintained by the dialog management module of the system (see
\S\ref{chap3:sec:dm}).

In the PARADISE framework, which is developed for task-oriented systems, the
dialog efficiency (e.g., the number of turns or elapsed time to complete the
task) and measures of qualitative phenomena (e.g., inappropriate or repair
utterances) are suggested to be relevant contributors to dialog
costs~\cite{Walker1997ACL}.
However, socialbot conversations have very different measures for dialog costs.
In particular, it is not clear that the concept of dialog efficiency is
relevant, since extended conversations can be good for socialbots.
In this work, we hypothesize that some types of turns may be less desirable,
i.e., \texttt{NegotiateTopic}, \texttt{ConfirmToContinue},
\texttt{SignalNonUnderstanding}, and \texttt{InstructOrRepeat}.

\subsection{Correlation analysis}
\label{chap4:ssec:correlation_analysis}

To examine how different conversation acts contribute to the quality of a
conversation, we calculate two metrics for each conversation act, i.e., the number and
the percentage of turns tagged by the conversation act.
Looking at the turn numbers for individual conversation acts has potential issues since
these numbers are highly correlated with each other as well as the total number
of turns.
Thus, a positive correlation coefficient does not necessarily suggest such turns
are positive factors of socialbot conversations.
They may be all positively correlated with user ratings since a larger number
usually suggest a longer conversation.
On the other hand, the percentage of turns is normalized by the total number
turns, and thus, is unlikely to have this issue.  

\begin{table}[t]
	\centering
	\begin{tabular}{l|r|r}
		\hline\hline
		{\bf Conversation Act} 
		& $\bm{r}$\textsubscript{\bf num} 
		& $\bm{r}$\textsubscript{\bf pct} \\
		\hline
		\texttt{AskQuestion} 							& \underline{.01} & --.09 \\
		\texttt{RequestHelpOrRepeat} 			& .06 & \underline{.04} \\
		\texttt{ProposeTopic} 						& .04 & --.05 \\
		\texttt{AcceptTopic} 							& .14 & .12 \\
		\texttt{RejectTopic} 							& .03 & --.10 \\
		\texttt{FollowAndNonNegative} 		& {\bf .17} & {\bf .24} \\
		\texttt{InterestedInContent} 			& .11 & --\underline{.00} \\
		\texttt{NotInterestedInContent} 	& .03 & {\bf --.17}\\
		\texttt{PositiveToContent} 				& .12 & .03 \\
		\texttt{NegativeToContent} 				& .07 & --.03 \\
		\texttt{PositiveToBot} 						& \underline{.02} & -\underline{.02} \\
		\texttt{NegativeToBot} 						& -\underline{.02} & -\underline{.02} \\
		\hline
		\texttt{NegotiateTopic} 					& .09 & --.10 \\ 
		\texttt{ConfirmToContinue} 				& .12 & .03 \\ 
		\texttt{InstructOrRepeat} 				& .06 & \underline{.00} \\ 
		\texttt{SignalNonUnderstanding} 	& --.\underline{00} & --.06 \\ 
		\texttt{DiscussTopic} 						& {\bf .17} & .09 \\ 
		\hline\hline
	\end{tabular}
	\caption{Correlation coefficients of the number ($r$\textsubscript{num}) and percentage
	($r$\textsubscript{pct}) of turns for individual conversation acts with respect to
	conversation-level user ratings.
	Numbers with underline indicate statistical insignificance 
	($p > 10^{-6}$).
	Numbers in bold indicate stronger correlation than the total number of turns
	($r=0.15$).
	}
	\label{chap4:tab:correlation_analysis}
\end{table}

In our analysis, we compute the Pearson $r$ for individual metrics with respect
to conversation-level user ratings.
Since we focus on correlation analysis here, only the training data (41076
conversations) are used.
The results are summarized in
Table\,\ref{chap4:tab:correlation_analysis}.
For clarity, the following discussion ignores results that are not statistically
significant.

{\bf Percentage vs.\ Number}:
We can observe that only metrics based on the percentage of turns reflect
negative correlation with user ratings; all metrics based on the number of turns
are positively correlated with user ratings.
This observation verifies our concerns about using the number of turns alone to
interpret positive and negative factors of socialbot conversations.
For all conversation acts with negative $r$\textsubscript{pct}, the
corresponding corresponding $r$\textsubscript{num} coefficients are positive but
have a small magnitude ($0<r$\textsubscript{num}$\le 0.10$).
For all conversation acts with a large positive $r$\textsubscript{num} coefficient
($r$\textsubscript{num}$>$0.10), the corresponding $r$\textsubscript{pct} is
always positive.

{\bf Negative Factors}:
Based on our results, \texttt{RejectTopic}, \texttt{NotInterestedInContent}, and
\texttt{NegotiateTopic} are negative conversation acts which have a strongly
negative $r$\textsubscript{pct} ($r$\textsubscript{pct}$\le-0.10$).
The first two user acts are strong indicators of negative engagement, since the user
rejects the bot proposed topic or expresses no interest in discussing the bot
presented content.
The bot act \texttt{NegotiateTopic} is less desirable since frequent confirmation
can be annoying.

{\bf Mildly Negative Factors}:
Some conversation acts have mildly negative $r$\textsubscript{pct} 
($-0.10<r$\textsubscript{pct}$<0$), including
\texttt{AskQuestion}, \texttt{ProposeTopic}, \texttt{NegativeToContent}, and
\texttt{SignalNonUnderstanding}.
The former two user acts indicate that the user is likely bored with the current
conversation flow and decides to take the control, which sometimes is a negative
sign of user satisfaction.
Additionally, the bot is not always able to answer the question or find content
about the user proposed topic, which could also explain the negative correlation.
Unlike \texttt{NotInterestedInContent}, the \texttt{NegativeToContent} does not
necessarily means the user is not engaged since the user only makes a negative
comment about the content (e.g., ``{\it that's sad}''). 
The bot usually performs error recovery using explicitly instructions when
making the \texttt{SignalNonUnderstanding} act, which may not be very annoying
but is probably less favorable for a conversation.
Note that negative correlation does not means that the bot should not use these
acts at all.
Instead, it suggests that we should improve the bot's topic preference
prediction and language understanding capabilities.

{\bf Positive Factors}:
Positive conversation acts ($r$\textsubscript{num}$\ge$0.10) include
\texttt{AcceptTopic}, \texttt{FollowA-}\texttt{ndNonNegative},
\texttt{InterestedInContent}, \texttt{PositiveToContent},
\texttt{ConfirmToContinue}, and \texttt{DiscussTopic}.
In particular, the \texttt{FollowAndNonNegative} act which is specially designed
for capturing user engagement has the highest $r$\textsubscript{num} and
$r$\textsubscript{pct} among all studied conversation acts.
This observation suggests that having the socialbot lead the conversation flow
is a reasonable strategy as long as it does not trigger user's negative
interest.
The strongest correlated bot conversation act is \texttt{DiscussTopic}, since
these turns are much more informative comparing to other bot acts.
A higher number of \texttt{DiscussTopic} turns indicates more information
is introduced in the conversation, which agrees with the hypothesis in
\cite{Yu2016Strategy} that the quality of open-domain chatbot conversations is
better if they bring in more information.

{\bf Other Factors}:
Both \texttt{RequestHelpOrRepeat} and \texttt{InstructOrRepeat} are mildly
positive.
They are almost equivalent since the bot always uses \texttt{InstructOrRepeat}
when the user makes a \texttt{RequestHelpOrRepeat} act, but the bot can also use
it when the it detects user confusion.
Thus, although these turns are not as informative as \texttt{DiscussTopic}, they
still play a positive role in the socialbot conversation.
The two user conversation acts \texttt{PositiveToBot} and \texttt{NegativeToBot}
that express opinions about the bot or its maker do not show significant
correlation with user ratings.
This is primarily due to their rare occurrences.
As shown in Table\,\ref{chap4:tab:user_dialog_act_annotation}, 85\% of annotated
user utterances do not reflect the user's opinion about the bot or its maker.
Among the 41076 conversations used for correlation analysis, only 5162 have
user turns tagged as \texttt{PositiveToBot} or \texttt{NegativeToBot}, and the
average number of such turns are 1.06 and 0.05, respectively, correspond to an
average number of user turns of 31.6 for this subset of conversations.
Nevertheless, we can still observe a trend that the number of
\texttt{PositiveToBot} turns is positively correlated with user ratings, and
that of \texttt{NegativeToBot} turns is negatively correlated.
Note that $r$\textsubscript{pct} is not very informative in this case since the
percentage values are extremely small.

{\bf Number-of-Total-Turns vs.\ Conversation-Act-Based Metrics}:
As discussed in \S\ref{chap4:ssec:discussion_related_work}, 
previous studies have not reported a metric that achieves higher correlation with
conversation-level user ratings than the number of total turns.
From our experiments, we show that the following four metrics have stronger
correlation than the baseline using the number of total turns, i.e., 
\#\texttt{FollowAndNonNegative}, \%\texttt{FollowAndNonNegative},
\%\texttt{NotInterestedInCont-}\texttt{ent}, and \#\texttt{DiscussTopic}.

\subsection{Regression analysis}
\label{chap4:ssec:regression_analysis}

The conversation-act-based metrics discussed in
\S\ref{chap4:ssec:correlation_analysis} can be viewed as turn-based metrics that
capture the conversation quality in different ways.
In the following set of experiments, we investigate weighted combinations of
these turn-based metrics to obtain a conversation-level metric that
characterizes the
overall conversation quality as a proxy of user rating.
Specifically, we use the conversation-act-based metrics as input features for a linear
regression model that estimates user ratings, i.e., 
$\gamma = {\bf u}^\top [{\bf x}; 1]$,
where ${\bf x} \in \mathbb{R}^n$ is the feature vector 
and ${\bf u} \in \mathbb{R}^{n + 1}$ is the weight vector including the bias.
Three feature sets are compared:
${\bf x}$\textsubscript{user} with 24 features for the 12 user conversation acts,
${\bf x}$\textsubscript{bot} with 10 features for the 5 bot conversation acts,
and ${\bf x}$\textsubscript{user+bot} with 34 features for both user and bot
conversation acts. 
These features are z-normalized using the mean and variance computed on the
training data.
The regression models $\gamma$\textsubscript{user}, $\gamma$\textsubscript{bot},
and $\gamma$\textsubscript{user+bot} are trained using corresponding features
with an $\ell_1$-regularized mean squared error (MSE) objective.
The regularization penalty weight is tuned based on the MSE on the validation
set.

\begin{table}[t]
	\centering
	\begin{tabular}{l|ccc|ccc}
		\hline\hline
		\multirow{2}{*}{\bf Model} 
		& \multicolumn{3}{c|}{Train}
		& \multicolumn{3}{c}{Validation} \\
		\cline{2-7}
		& {\bf $r$} 
		& {\bf $\rho$} 
		& {\bf $R^2$}
		& {\bf $r$} 
		& {\bf $\rho$} 
		& {\bf $R^2$} \\
		\hline
		NumTurns
		& .15 & .16 & .02 
		& .13 & .15 & .02 \\
		$\gamma$\textsubscript{user}
		& .28 & .26 & .08 
		& .26 & .24 & .07 \\
		$\gamma$\textsubscript{bot}
		& .22 & .21 & .05
		& .19 & .19 & .04 \\
		$\gamma$\textsubscript{user+bot}
		& .28 & .26 & .08
		& .26 & .24 & .07 \\
		\hline\hline
	\end{tabular}
	\caption{Regression performance using different feature sets.}
	\label{chap4:tab:conversation_level_regression}
\end{table}

The regression performance results are summarized in
Table\,\ref{chap4:tab:conversation_level_regression}, using the NumTurns as the
baseline metric which is the number of total turns in the conversation.
We report the Pearson $r$, Spearman $\rho$, and $R^2$ on both the train and
validation sets.
Note that a low $R^2$ is expected due to the high variance of user ratings.
It can be seen that all three combinations $\gamma$\textsubscript{user},
$\gamma$\textsubscript{bot}, and $\gamma$\textsubscript{user+bot} outperform the
NumTurns baseline.
The $\gamma$\textsubscript{user} is significantly better than
$\gamma$\textsubscript{bot}, and the $\gamma$\textsubscript{user+bot} does not
bring extra improvement.
This observation indicates that the user conversation acts contain the
most useful information for predicting the user ratings.

\subsection{Implications and limitations}
\label{chap4:ssec:implications}

From our analysis, we have shown that it is a positive sign when the user
accepts the conversation flow when the bot is the primary speaker who leads
the conversation, as evidenced by the positive correlation of
\texttt{FollowAndNonNegative}, \texttt{AcceptTopic}, and
\texttt{InterestedInContent}, as well as the negative correlation of
\texttt{RejectTopic} and \texttt{NegativeToContent}.
It suggests the importance of designing and maintaining good conversation flows
for socialbots.
It should be noted that being a primary speaker of the conversation is not
equivalent to solely using system initiatives and ignoring user initiatives.
Properly handling user initiatives is a challenging problem considering the
open-domain nature of the socialbot conversation.
The observation that \texttt{AskQuestion} and \texttt{ProposeTopic} slightly
impact the user rating in the negative direction suggests that we should improve
the bot's capability of handling user questions and topic/activity requests.
For example, one type of activity requests which is not handled well by our
system is the user request of taking the primary speaker role and telling a
story to the bot.

Recently, there is a surge of research interest in using data-driven models for
response generation, which seems to be promising as it can take advantage of
large-scale conversation data and can produce grammatical and coherent
responses.
However, as mentioned in \cite{CMUTartan2018} and observed by several other
Alexa Prize teams, these models currently do not perform well enough compared to
the hand-crafted conversation flows.
A potential reason we hypothesize is that these models are not trained to
switch between the roles of primary speaker and active listener.
Thus, the socialbot may make better use of data-driven models for learning and
managing conversation flows if the model can distinguish the role of
participants in the conversation it is trained on so that it can learn how to
switch between them properly.
We leave this as an open question for future studies.

The analysis in this section has provided high-level insights
on positive and negative factors of socialbot conversations, which helps us find
potential issues of the system.
However, to further pin down the system issues for improving the bot,
correlation analysis using turn-level conversation acts is not enough, and
approaches for diagnosing the socialbot performance at other levels are needed.
For example, for the \texttt{NegotiateTopic} act, it is useful to compare
different topic initiation strategies that the bot uses.
It is also useful to study the bot's overall performance in discussing a specific
topic, which may be affected by available content in the knowledge base and
language understanding capabilities. 
Another important aspect that influences the socialbot performance is the
content source, since the usefulness of individual content sources can vary
significantly.
While conversation-level user ratings can be used to identify problematic
conversations, it is not straightforward to find which part of the conversation
went wrong.
On the other hand, conversation acts at the turn level do not measure multiple
turns as a coherent whole that influence a specific system decision.
For example, although the user may accept a bot-proposed topic, the bot's
performance in discussing this topic should be measured on the conversation
segment of this topic and/or its relevant topics.
The same is true for topic initiation strategies and content sources.
Thus, it is desirable to have a method to score dialog segments for locating
problems and diagnosing system performance.
Dialog segment scoring is also useful to avoid training response generation
models on problematic data in the corpus.
With these motivations, the remaining of this chapter focuses on developing
methods for automatically scoring conversation segments.

\section{Multi-Level Segmentation and Scoring}
\label{chap4:sec:multi_level}
As discussed in \S\ref{chap4:sec:related_work}, it has been observed that casual
conversations usually behave like a sequence of structural segments.
Naturally, a socialbot conversation can have both good and bad segments in terms
of user satisfaction.
However, users only provide conversation-level ratings, and it is difficult to
obtain segment-level ratings without interrupting the conversation flow, which
makes it challenging to evaluate individual conversation segments.
Further, dialog segment evaluation also calls for a model to segment a socialbot
conversation, since previous dialog structure models are either designed for
task-oriented systems or human-human casual conversations, which are not
straightforward to apply to socialbot conversations.

In this section, we first present a hierarchical dialog segmentation model for
socialbot conversations in \S\ref{chap4:ssec:segmentation}, which allows us to
obtain topical conversation segments at multiple levels.
Using this segmentation method, we explore two multi-level scoring methods,
i.e., a linear scoring model in \S\ref{chap4:ssec:linear_scoring} and a
sequential scoring model in \S\ref{chap4:ssec:bilstm_scoring}.
These two models are designed based on two different hypotheses.
In the linear scoring model, we hypothesize that segment scores are predicted
just like conversation scores, and thus, we can treat a segment the same as a
complete conversation.
The sequential scoring model instead assumes a conversation score is some
non-linear aggregation of the sequence of segment scores.

Throughout the remainder of this chapter, {\it ratings} are used exclusively to
refer to the rating of a conversation provided by the user.
{\it Scores} are the output of scoring models, which are similar to the term {\it
metric} in previous discussions since they are used for evaluation as a proxy
for user ratings.
{\it Features} are the input to scoring models, which can be either
conversation-act-based metrics as used in the regression analysis in
\S\ref{chap4:ssec:regression_analysis} or hand-crafted representations of
conversation segments.

\subsection{A dialog segmentation model for socialbot conversations}
\label{chap4:ssec:segmentation}
An open-domain socialbot conversation can involve a wide range of topics, and
users can switch between quite unrelated topics, e.g., ``sports,'' ``science,''
and ``Christmas.'' 
In this work, we model the dialog structure as a sequence of topical 
{\bf subdialogs}.
The subdialog and its topic can be initiated by either the bot or the user.
Further, an issue that comes up in a discussion on one topic can lead to a brief
diversion or spawn a discussion of other related topics. 
Therefore, we model a subdialog as a sequence of {\bf microsegments} on related
topics that either diverge or are spawned from the previous discussion by either
the user or the bot.
Note that we treat functional segments (e.g., greeting, goodbye, listing topics)
as special subdialogs with only one microsegment.
Within each topical subdialog/microsegment, the bot presents multiple posts on
the topic, each of which may span multiple consecutive turns.
While a conversation can be modeled as a sequence of subdialog/microsegment
segments, it cannot be treated as a sequence of posts since not all turns
are related to a post.
Nevertheless, we can still construct a {\bf post segment} using the consecutive turns
spanned by the post, i.e., a post segment is a subsequence of turns in a
microsegment.

\begin{table}[t]
	\centering
	\begin{tabular}{l|l}
		\hline\hline
		{\bf Policy} & {\bf Description} \\
		\hline
		T1-accept		& accept the user topic \\
		T2-backoff 	& reject the user topic \& propose a related topic \\
		T3-reject		& reject the user topic \& propose a random topic \\
		T4-popular 	& propose one of the generally popular topics \\
		T5-implicit	& extract a topic from the user utterance \\
		T6-spawn 		& extract a topic from the last discussed post \\
		T7-inherit	& inherit a topic from the previous discussion \\
		\hline\hline
	\end{tabular}
	\caption{Topic initiation policies for microsegments.}
	\label{chap4:tab:topic_context_types}
\end{table}

For the sample conversation illustrated in
\S\ref{chap3:sec:sample_conversation}, turns 1--3 form a greeting subdialog and
turn 4 forms a subdialog for listing topics.
Turns 5--14 is a subdialog about superman consisting of two microsegments.
The first one is from turn 5 to 13 since the topic has not changed.
At turn 14, the bot wants to talk about a spawned topic which is the movie's
director.
This initiates a new microsegment, but is not extended since the user initiates
a new topic at turn 15.
Turns 15--24 constitute a subdialog about science, consisting of 3 microsegments:
a microsegment from turn 15 to 17 talking about a specific item of science news,
a microsegment from turn 18 to 23 talking about the topic ``car'' spawned from
the news headline, and a microsegment with a single turn 24 using the topic
``science'' inherited from the subdialog's first microsegment.
Turns 25--29 form a subdialog about celebrities, specifically, with a single
microsegment about``Beyon\'{e}''.
Within individual microsegments, there are several post segments:
turns 6--7 form a segment of a fun fact,
turns 8--9 form a segment of an amusing thought,
turns 10--14 form a segment of a movie,
turns 16--17 form a segment of a news headline,
turns 18--21 form a segment of a life pro tip,
turns 22--23 form a segment of a joke,
and turns 26--28 form a segment of a trivia question.

The dialog context tracking module in the system logs how the topic of a
microsegment is initiated based on different policies described in
Table\,\ref{chap4:tab:topic_context_types}.
For T1--T3, the topic is originally initiated by the user.
The bot may reject the user-initiated topics depending on the content
availability and topic appropriateness, e.g., the bot may refuse to talk about
controversial topics or give legal advice.
Since the bot is encouraged to push the conversation forward, it always
proposes a topic when rejecting the user-initiated topic.
For T2-backoff, the topic proposed by the bot is a noun (or noun phrase)
extracted from the user-initiated topic.
For example, if the user initiates a topic ``Google Scholar,'' the bot may
propose to talk about either ``Google'' or ``Scholar.''
For T3-reject, the bot cannot talk about the user-initiated topic or any noun
phrases in the topic, and thus, proposes a topic based on user personality
and/or topic popularity.
For T4-popular, no explicit topic change intent is detected, but the bot
proposes a new topic chosen from a set of popular topics.
Similarly, the bot can choose a topic from the noun phrases in the user
utterance (T5-implicit) or topics mentioned in the last discussed post
(T6-spawned), based on the assumption that they are likely to be relevant to the
current discussion.
For T7-inherit, the bot uses the topic that initiates the current
subdialog.
In our dialog structure model, T1--T4 would result in a new subdialog and the
corresponding microsegment is the leading microsegment of the new subdialog,
whereas T5--T7 only initiate a new microsegment for the current
subdialog.

\subsection{Linear scoring model}
\label{chap4:ssec:linear_scoring}
\begin{table}[t]
	\centering
	\begin{tabular}{c|c||c|c||c}
		\hline\hline
		& \# Turns & $x_1$ & $x_2$ & $\gamma$ \\ 
		\hline
		conversation 	& 100 & 30 & .30 & 3.06 \\
		\hline
		subdialog 1   & 60 & 15 & .25 & 1.55 \\
		subdialog 2 	& 30 & 6 	& .20 & 0.64 \\
		subdialog 3 	& 10 & 9 	& .90 & 1.08 \\
		\hline\hline
	\end{tabular}
	\caption{A toy example of computing subdialog scores using a learned linear model 
		($\gamma = 0.1 x_1 + 0.2 x_2$).
		$x_1$ and $x_2$ are the number and percentage of
		\texttt{FollowAndNonNegative} turns in the conversation/subdialog,
		respectively.
	}
	\label{chap4:tab:example_linear_scoring}
\end{table}

Recall that the data only have conversational-level user ratings,
and our goal is to estimate scores for individual subdialogs and microsegments.
In our first approach, a linear scoring model is trained on conversation-level
user ratings using a similar setting as in the regression analysis in
\S\ref{chap4:ssec:regression_analysis}.
Then, we simply treat individual segments (subdialogs/microsegments/posts) in
the same way as a conversation and apply the learned linear scoring models to these
segments with features constructed at the segment level.
We shall see in \S\ref{chap4:sec:experiments} that this simple assumption is
quite effective, probably due to the robustness and generalizability of the
learned linear scoring model.

A toy example for computing subdialog scores is illustrated in
Table\,\ref{chap4:tab:example_linear_scoring}.
Note that in this example, feature normalization and bias in the linear model
are omitted for brevity.
In practice, dense and numerical features are z-normalized based on their mean
and variance in the training data, whereas sparse features are not normalized.

\subsection{BiLSTM scoring model}
\label{chap4:ssec:bilstm_scoring}
To better exploit the sequential nature of conversations, we also investigate
obtaining segment scores with a bi-directional LSTM (BiLSTM)
\cite{Hochreiter1997} recurrent neural network (RNN). 
In particular, we model the scoring process by learning the sequence of segment
scores as latent variables that contribute to the final conversation-level
score.

In a conversation with $T$ segments (subdialogs/microsegments), each segment is represented by its
segment-level features ${\bf x}_t \in \mathbb{R}^d$.
Unlike the linear scoring model, we do not normalize any features for neural
network models.
Instead, we project the raw feature vector ${\bf x}_t$ to context-independent
embeddings ${\bf y}_t \in \mathbb{R}^n$, i.e.,
${\bf y}_t = \sigma({\bf P} {\bf x}_t + {\bf b}_y) $, where
${\bf P} \in \mathbb{R}^{n \times d}$ is the weight matrix, 
${\bf b}_y \in \mathbb{R}^{n}$ is the bias vector, and $\sigma(\cdot)$ is the
sigmoid activation function.
We also explore using multiple BiLSTM layers to obtain contextualized segment
embeddings:
\begin{align*}
	[{\bf h}_{1}^{(l)}, ... , {\bf h}_{T}^{(l)}] 
	= {\rm BiLSTM}^{(l)} ({\bf h}_1^{(l - 1)}, ... , {\bf h}_{T}^{(l - 1)}),
\end{align*}
where ${\bf h}_{t}^{(l)} \in \mathbb{R}^{n}$ is the concatenated LSTM hidden
outputs of the forward and backward RNN networks for the $t$-th segment at the
$l$-th BiLSTM layer, and ${\bf h}_t^{(0)} = {\bf y}_t$.
All BiLSTM layers use the hyperbolic tangent activation function.
The estimated score for the $t$-th segment is
$\gamma_t = {\bf w}^T {\bf h}^{(L)}_t + b_\gamma$,
where ${\bf w} \in \mathbb{R}^n$ is the weight vector,
$b_\gamma \in \mathbb{R}^1$ is the bias,
and ${\bf h}^{(L)}_t$ is the hidden outputs of the final BiLSTM layer.
The conversation-level score is computed from the sequence of segment scores
using three (mean, max, min) pooling-over-time functions $ z_i = f_i ([\gamma_1,
\dots, \gamma_T]) $ 
as input to a linear projection output layer, i.e.,
$o = \sum_{i=1}^3 v_i z_i + v_0$,
where $v_i$ and $v_0$ are weights to be learned.
The BiLSTM model described here is different from the hierarchical RNN proposed
in \cite{Serban2016AAAI} which generates responses using a bi-directional RNN
with gated recurrent hidden units \cite{Cho2014EMNLP} to encode the utterance
word by word and a uni-directional RNN to encode the dialog context turn by
turn.

The BiLSTM models are trained separately for microsegments and subdialogs for
comparison.
Not all turns involve posts, so the BiLSTM is not applied at this level.
Each model is trained end-to-end using the MSE loss and ADAM
\cite{Kingma2015ICLR} for stochastic gradient descent.
The batch size is set to 16.
The learning rate is halved at each iteration once the objective on the
validation set decreases.
The whole training procedure terminates when the objective decreases for the
second time. 
All learning parameters except bias terms are initialized randomly according
to the Gaussian distribution $\mathcal{N}(0, 0.01)$. 
In our experiments, we tune the initial learning rate, and choose the best one
based on the objective on the validation set at the first epoch.
The training data are randomly shuffled at every epoch.
To reduce the number of tuning parameters, we use the same hidden layer
dimension for all BiLSTM layers and the segment embedding layer.
We tune the hidden layer size $n$ from 16 to 128 with a step size of 16, and the
number of BiLSTM layers $L$ from 1 to 3.

\section{Multi-Level Scoring Experiments}
\label{chap4:sec:experiments}
In this section, we first describe features used for the multi-level scoring
models in \S\ref{chap4:ssec:features}.
Automatic prediction of conversation-level scores is evaluated in
\S\ref{chap4:ssec:exp_conversation_level} via experiments with different
combinations of features by examining the correlation between the predicted
scores with user ratings.
Automatic subdialog score predictions are evaluated in
\S\ref{chap4:ssec:exp_subdialog_level} using crowdsourced human judgments for
subdialog pairs.
Lastly, we evaluate predicated segment scores in
\S\ref{chap4:ssec:exp_diagnosis} in terms of their utility for identifying
system issues.

\subsection{Features}
\label{chap4:ssec:features}
Other work on automatic prediction of conversation scores uses a variety of
features \cite{Venkastech2017NIPS,Serban2017arXiv}.
Motivated by these studies and our own conversation act analyses in
\S\ref{chap4:sec:factors}, we use several features that characterize the
conversation in terms of the user, bot, and content aspects.

{\bf User-ConversationAct}:
The number and percentage of turns in the conversation or segment for individual
user conversation acts listed in
Table\,\ref{chap4:tab:user_dialog_acts}.

{\bf User-Verbosity}:
Users vary in their talking styles (succinct vs. verbose).
To capture this, we compute the total number of user tokens and the average
number of user tokens per turn in the conversation or its corresponding segment.

{\bf User-BagOfWords}:
To capture the user's language use, the bag-of-word-count feature is used where
the count is normalized by the total number of user utterance tokens in the
conversation or its corresponding segment.
The size of the vocabulary is 4658 after using an out-of-vocabulary token to
replace rare words that occur less than 10 times in the training data.

{\bf Bot-ConversationAct}:
The number and percentage of turns in the conversation or segment for individual
bot conversation acts listed in
Table\,\ref{chap4:tab:bot_dialog_acts}.

{\bf Bot-DialogState}:
As described in Chapter~\ref{chap3}, the bot uses a finite-state-based dialog
management.
In total, there are 47 dialog states in the system. 
We compute the number and percentage of turns for individual states
in the conversation or its corresponding segment, resulting in 94 features.

{\bf Bot-BagOfWords}:
To capture the bot's language use, we also use the bag-of-word-count feature
where the count is normalized by the total number of bot utterance tokens in the
conversation or its corresponding segment.
The size of the vocabulary is 6137 after using an out-of-vocabulary token to
replace rare words that occur less than 100 times in the training data.
Note that the bot vocabulary is much bigger than the user vocabulary because of
the content posts it presents (e.g., news, facts).

{\bf Content-Coherence}:
As a proxy measure for coherence of content mentioned in the bot response, we
compute cosine similarity scores between the bot utterance and its context using
the term-frequency inverse-document-frequency (TF-IDF) vector representation.
Three context scopes are considered: 
1) the last user and bot turns, 
2) the current microsegment,
and 3) the current subdialog.
To construct coherence features, we compute the sum, mean, and maximum of the
coherence scores for individual bot turns with respect to these three context
scopes, resulting in 9 numerical features in total.

{\bf Content-TopicPolicy}:
For corresponding segment spans (a conversation or a subdialog), the number and
percentage of microsegments for each topic initiation policy listed in
Table\,\ref{chap4:tab:topic_context_types} are computed, resulting in 14
features.

{\bf Content-BagOfTopics}:
The bag-of-topics-count feature is used where the count is normalized by the
total number of topics in the conversation or segment.
The size of the topic vocabulary is 7079 after removing topics that occur only
once in the training data.
The topic can be a noun or noun phrase that refers to an entity (e.g., Amazon,
cat) or a generic topic (e.g., technology, politics).

Note that for the linear scoring model, non-sparse numerical features are
standardized based on their mean and variance on the training set,
whereas sparse features (User-BagOfWords, Bot-BagOfWords, Content-BagOfTopics)
are used in their raw form.

\subsection{Evaluation of conversation scores}
\label{chap4:ssec:exp_conversation_level}
\begin{table}[t]
	\centering
	\begin{tabular}{l|c|c|c|c}
		\hline\hline
		{\bf Feature} 
		& $|\mathcal{F}|$
		& {\bf $r$} 
		& {\bf $\rho$} 
		& {\bf $R^2$} \\
		\hline
		User-ConversationAct & 24
		& .26 & .24 & .07 \\
		User-Verbosity & 2
		& .17 & .17 & .03 \\
		User-BagOfWords & 4k+ 
		& .30 & .28 & .09 \\
		\hline
		Bot-ConversationAct & 10
		& .19 & .19 & .04 \\
		Bot-DialogState & 94 
		& .22 & .20 & .05 \\
		Bot-BagOfWords & 6k+ 
		& .26 & .23 & .07 \\
		\hline
		Content-Coherence & 6
		& .12 & .14 & .01 \\
		Content-TopicPolicy & 14
		& .13 & .11 & .02 \\
		Content-BagOfTopics & 7k+ 
		& .14 & .13 & .02 \\
		\hline
		User-All & 4k+
		& {\bf .32} & {\bf .30} & {\bf .10} \\
		Bot-All & 6k+
		& .27 & .26 & .08 \\
		Content-All & 7k+
		& .22 & .20 & .05 \\
		User + Bot + Content & 17k+
		& .32 & .30 & .10 \\
		\hline\hline
	\end{tabular}
	\caption{Validation set results for conversation score prediction of linear
		scoring models using different features.
	}
	\label{chap4:tab:exp_conversation_level_validation}
\end{table}

We explore different combinations of features by examining their
conversation-level score prediction performance using linear scoring models
trained with $\ell_1$ regularization where the regularization penalty is tuned
based on the MSE of the validation set.
The conversation-level user ratings are used as the targets for predicting
conversation scores.
We report the size of the feature set $|\mathcal{F}|$, Pearson $r$, Spearman
$\rho$, and $R^2$ on the validation set in
Table\,\ref{chap4:tab:exp_conversation_level_validation}.
It can be seen that user-related features (User-All) significantly outperform
both bot-related features (Bot-All) and content-related features (Content-All).
As shown by the last row, merging all three feature groups does not give
additional gain over User-All. 
This indicates that user reactions, intents, and language use contain the most
useful information for predicting the final conversation-level user ratings for
this version of the system\footnote{Findings as to the importance of
	different feature groups may change as features are improved.
	For example, it is possible that the simple TF-IDF similarity measure is not
	sufficient for characterizing topic coherence. Findings are also likely to
change as the system improves.}, which mostly agrees with our observations in
\S\ref{chap4:ssec:regression_analysis}.	

\begin{table}[t]
	\centering
	\begin{tabular}{l|c|c|c}
		\hline\hline
		{\bf Scoring Model} 
		& {\bf $r$} 
		& {\bf $\rho$} 
		& {\bf $R^2$} \\
		\hline
		NumTurns
		& .15 & .16 & .02 \\
		GBDT: User-All
		& .30 & .28 & .09 \\
		GBDT: User + Bot + Content
		& .31 & .29 & .10 \\
		\hline
		Linear
		& .32 & .30 & .10 \\
		Subdialog BiLSTM
		& .31 & .29 & .09 \\
		Microsegment BiLSTM
		& .31 & .29 & .10 \\
		\hline\hline
	\end{tabular}
	\caption{Test set results for conversation score prediction using different
	models.}
	\label{tab:exp_regression_test}
\end{table}

Since User-All performs the best, we only use these features for our models in
subsequent experiments and analysis throughout this chapter.
Table\,\ref{tab:exp_regression_test} reports the performance of conversation
score prediction on the test set using different models, i.e.,
the {\bf Linear} scoring model, the {\bf Subdialog BiLSTM} scoring model that
segments the conversation by subdialogs, and the {\bf Microsegment BiLSTM}
scoring model that segments the conversation by microsegments.
These models are compared with the baseline method that uses the number of
total turns.
In addition, for comparison to prior study in \cite{Venkastech2017NIPS},
we also train gradient boosted decision tree (GBDT) models
\cite{Friedman2001Stats} using User-All features and User+Bot+Content
features.
All other models perform similarly well in predicting conversation scores.
These results are in the range of other reported correlation results on Alexa
Prize data, though they are not directly comparable due to system and data
differences.
Using another socialbot, Serban et al.\ \cite{Serban2017arXiv} report
$\rho=0.19$ for a linear regression model with 23 features 
including dialog length, features that characterize the bot utterance
(appropriateness, genericness, length), and features that characterize the user
utterance (sentiment, genericness, length, confusion).
On an aggregate dataset from multiple Alexa Prize socialbots, 
Venkastech et al.\ \cite{Venkastech2017NIPS} train GBDT and hierarchical LSTM
models \cite{Serban2016AAAI} using features including n-grams of user-bot turns,
token overlap between user utterance and bot response, conversation duration,
number of turns, and mean response time.
They obtain $r=0.24$ and $\rho=0.23$ for the hierarchical LSTM model, and
$r=0.35$ and $\rho = 0.35$ for the GBDT model.

\subsection{Evaluation of subdialog scores}
\label{chap4:ssec:exp_subdialog_level}
\begin{table}[t]
	\centering
	\begin{tabular}{l|c|c|c}
		\hline\hline
		{\bf Scoring Model} & Within Conv. & Cross Conv. & All  \\
		\hline
		NumTurns 							& .02 			& .13 			& .08 \\
		UserRating  					& - 				& .05 			& - \\
		Linear 								& .36$^*$ 	& .14 			& .26$^*$ \\
		Subdialog BiLSTM 			& .34$^*$ 	& .21$^*$ 	& .29$^*$ \\
		\hline\hline
	\end{tabular}
	\caption{Spearman rank correlations between scoring models and human
		judgements.
		$^*$ indicates statistical significance ($p \ll$ .001).
	}
	\label{tab:exp_human_eval}
\end{table}

We assess the quality of subdialog scores estimated from learned multi-level
scoring models via human judgments crowdsourced from the FigureEight platform.
Each worker is presented with two subdialogs and asked to predict in which case
the user might give a higher rating to the bot, 
i.e. ``{\it the user found the exchange amusing, interesting or useful}.''
To avoid bias introduced by the conversation length, subdialogs are paired to
have a length difference within 5 user turns.
The worker can choose either subdialog or ``cannot tell.''
Given multiple human judgments, a difference value for a subdialog pair (A, B)
is calculated by the number of votes on A minus that on B.
Similarly, a difference value can also be calculated for the pair using scores
estimated from a learned scoring model.
To evaluate the predicted subdialog scores, we compute the Spearman rank
correlation between the pair-wise difference values calculated on predicted
subdialog scores and that calculated on the human votes.
This pair-wise evaluation protocol avoids the issue of numerical scale
difference between human votes and predicted scores. 
Five judgments are collected for each subdialog pair.
Empirical results show that the correlation coefficient converges after there
are at least 3 judgments per pair.
In total, 500 random subdialog pairs are compared: 250 within-conversation pairs
compare subdialogs from the same conversation to control for user differences,
and 250 cross-conversation pairs compare subdialogs from different conversations
but on the same topic to control for topic differences.

Table\,\ref{tab:exp_human_eval} shows the results for four scoring models:
1)~NumTurns, which uses the number of user turns in the subdialog, 
2)~UserRating, which uses the conversation-level user rating applied as is to
all subdialogs in the conversation,
3)~Linear, which uses the learned linear scoring model,
and 4)~Subdialog BiLSTM, which uses the learned BiLSTM model that segments the
conversation by subdialogs.
Note that the correlation for within-conversation pairs is not informative for
the UserRating baseline, since the pair-wise score is always 0.
It can be seen that both Linear and Subdialog BiLSTM models significantly
outperform the NumTurns and UserRating baselines. 
Recall that the within-conversation pairs control for user, and the
cross-conversation pairs control for topic.
The cross-conversation pairs tend to have lower correlation coefficients,
suggesting that user differences contribute more to score variation than topic
differences.
The BiLSTM model performs much better than the Linear model for the
cross-conversation pairs, perhaps because the BiLSTM model is learning something
about the user by taking into account the surrounding subdialog context.

\subsection{System diagnosis using segment scores}
\label{chap4:ssec:exp_diagnosis}
In this series of experiments, we demonstrate ways that scores at different
segment levels can be used to identify
system weaknesses.
Results are reported using the learned linear scoring function since it applies
to all segment types, i.e., subdialog, microsegment, and post.

\begin{table}[t]
	\centering
	\begin{tabular}{c|p{14cm}}
		\hline\hline
		{\bf Rank} & {\bf Topics} \\
		\hline
		high
		& santa claus, video games, time travel, aliens, disney, food, music,
		artificial intelligence, pets \\ 
		\hline
		low
		& the article, that article, a story, snow, words, quantum physics,
		bad, home alone, point \\
		\hline\hline
	\end{tabular}
	\caption{Topics ranked highest and lowest based on predicated subdialog
	scores.}
	\label{tab:topic_analysis}
\end{table}

{\bf Topic diagnosis}:
Hypothesizing that the segment scores might be useful for identifying topics
that were handled successfully vs.\ problematically by the socialbot, we
developed a topic ranking protocol that ranks topics based on the predicted
scores for subdialogs associated with specific topics.
To control for user interests on the topic, only subdialogs initiated by the
user are considered, i.e., excluding subdialogs started with a bot-proposed
topic, which may or may not be interesting to the user.
We use subdialog scores rather than microsegment scores since the subdialog
score also captures the quality of the extended discussion on topics that are
potentially relevant to the subdialog topic, and we want the ranking metric to
take into account the long-term conversation flow for the topic.
For example, when discussing a movie, available content about its director and
actors may influence the overall quality of the subdialog.
In our study here, the ranking metric for topic $t$ is the difference between
the relative frequency of high subdialog scores $\Pr(\gamma_t > 3.6)$ and that
of low subdialog scores $\Pr(\gamma_t < 2.9)$, where the thresholds are manually
chosen based on the score distribution over all topics.
Positive vs.\ negative ranking metric values indicate topics that are
well-handled vs.\ problematic, respectively.
Table\,\ref{tab:topic_analysis} shows the highest and lowest ranked topics with
at least 5 subdialogs in our data.
Top-ranked topics usually have plenty of interesting content in the socialbot's
knowledge base.
It appears that many low-ranking topics are generic noun phrases associated with
the artifacts of parsing errors and imperfect topic extraction.
The topic ``quantum physics'' is ranked low probably because there are only 3
content posts (except news headlines) about this topic in the knowledge base.
The fact that it is a frequent user-initiated topic means we should probably
find some good content sources for this topic.

\begin{figure}[t]
	\centering
	\includegraphics[width=0.9\textwidth]{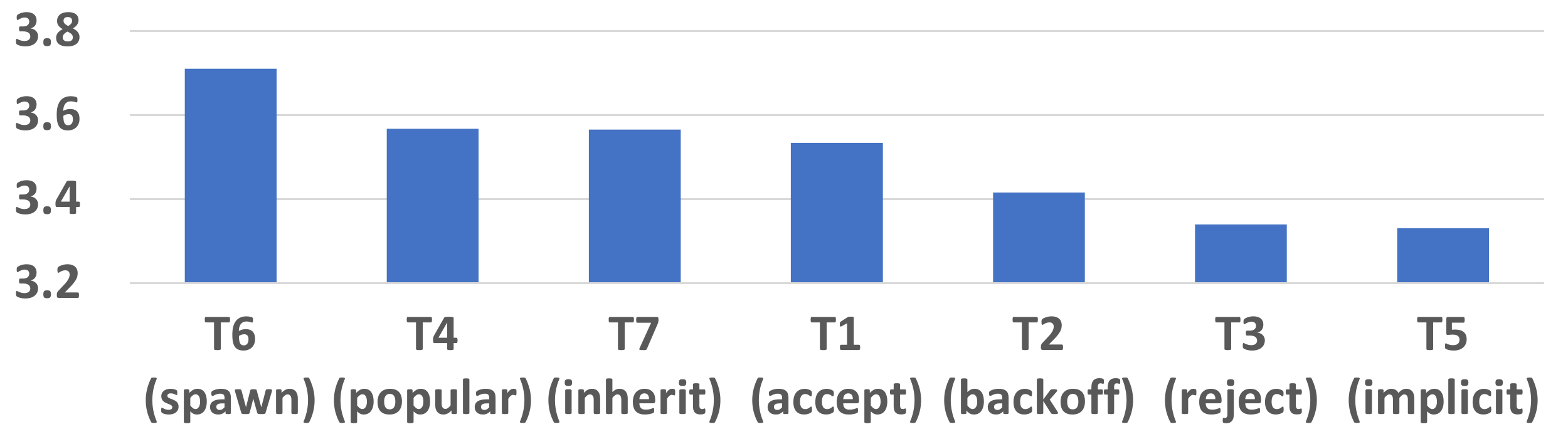}
	\caption{Average microsegment scores for different topic initiation policies described in
		Table\,\ref{chap4:tab:topic_context_types}.}
	\label{fig:topic_init_rewards}
\end{figure}

{\bf Topic initiation policy diagnosis}:
As given in Table\,\ref{chap4:tab:topic_context_types}, there are 7 different
topic initiation strategies for a microsegment.
Here we compare them by averaging all predicted microsegment scores across
conversations for individual strategies.
Results are shown in Fig.\,\ref{fig:topic_init_rewards}.
The highest average scores are obtained when the bot initiates a topic that is a
mentioned entity in the most recent post being discussed (T6-spawned).
Three strategies obtained good but somewhat lower average scores: the bot
returns to the topic of the first microsegment in the current subdialog
(T7-inherit), the bot suggests a generally popular topic (T4-popular), and the
bot accepts the user-initiated topic (T1-accept).
Not surprisingly, the average score is lower when the bot rejects the user topic
and proposes a backoff topic (T2-backoff) or a random topic (T3-reject). 
This suggests that increasing the topic coverage of the knowledge base would be
useful.
Further, it is potentially useful to develop better ways to handle topics that
are not covered by the knowledge base, e.g., using a better algorithm to find
topics that are relevant to the user requested topic.
For the topic initiation policy with the lowest average score (T5-implicit), no
explicit topic change intent is detected, and a candidate topic is chosen from
the noun phrases in the user utterance.
Often these topics are associated with general chit-chat phrases that are not
intended as a discussion topic or an artifact of imperfect language
understanding, as in:
\begin{tabbing}
	User: I watched that movie the \underline{music} is good \\
	Bot: 	You mentioned \underline{music}, do you want to hear something
	interesting about that?
\end{tabbing}
By observing such issues, we can improve the dialog policy using a list of
general chit-chat phrases to avoid the bot unnecessarily changing the topic.

\begin{figure}[t]
	\centering
	\includegraphics[width=0.9\textwidth]{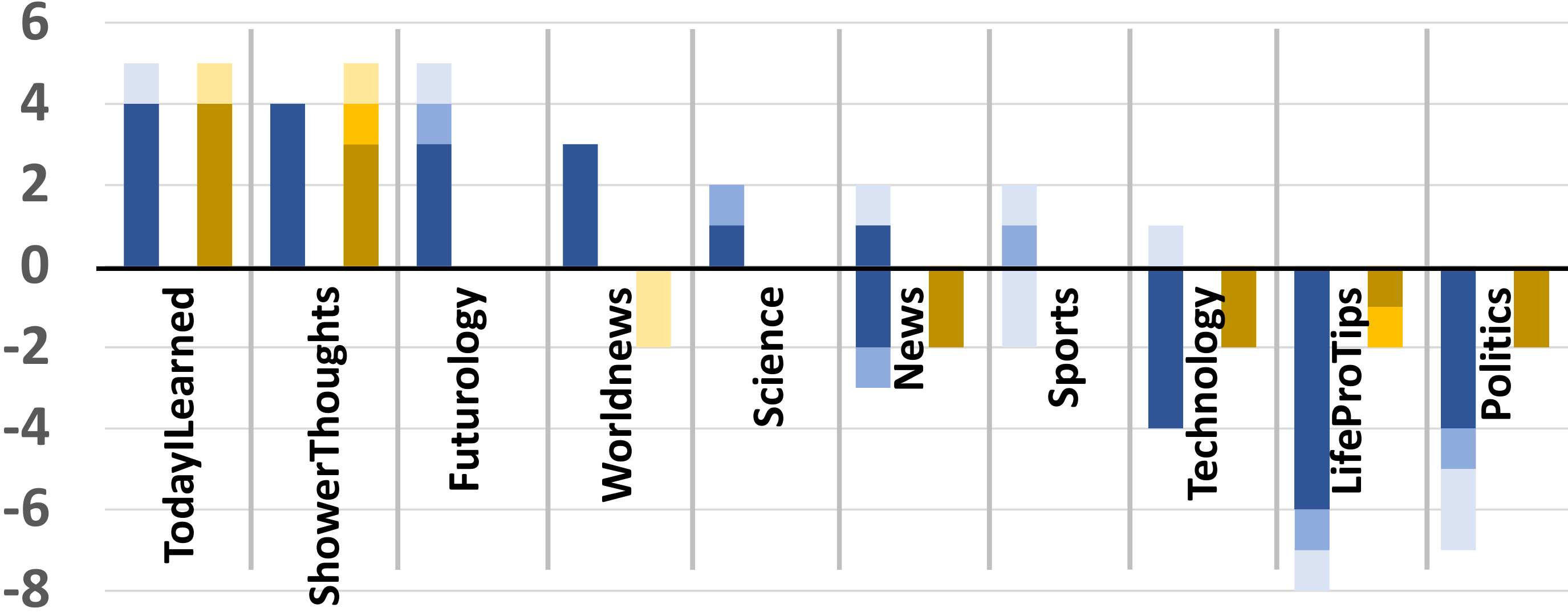}
	\caption{The win-loss records for subreddits.
		The positive and negative scales indicate wins and losses, respectively.
		Each bar is color-scale-coded by the p-value threshold $p_0$ (dark: 1e-5, medium:
		1e-4, light: 1e-3).
		Blue bars are computed using the linear scoring function, whereas yellow
		bars are computed using conversation-level user ratings.
	}
	\label{fig:subreddit_tournament}
\end{figure}

{\bf Content source diagnosis}:
We also explored using predicted post scores to assess quality of our content
sources irrespective of topic.
Here we compare 10 content sources (specifically, subreddits in Reddit) via
tournaments using estimated scores for posts.
For each pair of subreddits, we carry out a Mann-Whitney U test for the null
hypothesis that the post score distributions of the two subreddits are the same.
Subreddit pairs are kept if the null hypothesis is rejected with significance
($p < p_0$).
Among the kept subreddit pairs, we say a subreddit wins if the mean post score
is higher and loses if lower.
In this way, we can compute the wins and losses for each subreddit.
The win-loss records are shown in Fig.\,\ref{fig:subreddit_tournament}.
We compare the linear scoring model with the baseline that uses the
conversation-level user rating as the score for the post in the conversation.
Although we do not have a gold standard for subreddit rankings, we can
reasonably expect these subreddits (and therefore their post score
distributions) to be different considering the community style and topic
interests.
In this sense, the linear scoring model is clearly better than the baseline as
the magnitude of blue bars is consistently larger than the corresponding yellow
bars, i.e., we can get much more signifiant results using the linear scoring
model, except \texttt{TodayILearned} where both methods perform the same and
\texttt{ShowerThoughts} where the linear scoring model has less significant
results at the threshold of $10^{-3}$.
According to the win-loss records, the best sources are {\tt TodayILearned}
(interesting facts) and {\tt ShowerThoughts} (often funny).
The worst are {\tt Politics} (often controversial), {\tt Technology} (often
political, e.g., net neutrality), and {\tt LifeProTips} (often off topic,
boring, and/or annoying).
This study indicates that we should increase the use of the top three subreddits
and explore mechanisms for better identifying boring and controversial content
especially for the those relatively less interesting subreddits.

\section{Discussion}
\label{chap4:sec:discussion}
This work has carried out in-depth analysis on user ratings and aspects of
socialbot conversations using conversation acts proposed for characterizing both
user and bot turns.
The findings have provided valuable insights into how different bot behaviors
and user reactions contribute to conversation-level user ratings.
We have also found several metrics which achieve higher correlation with the
user rating than the strong baseline that uses the number of total turns.
To our knowledge, this is the first work that have examined socialbot
conversations with carefully designed conversation acts.

Using a novel dialog segmentation model for socialbot conversations, we have
introduced two approaches to estimate scores for different types of segments.
Experiment results have shown that the subdialog scores achieve much higher
correlations with human judges compared with the baselines that use the number
of turns and the conversation-level user ratings.
Additionally, we have provided examples showing how segment scores can be used
to diagnose system performance in terms of identifying topics that are not well
handled by the bot, examining topic initiation policies, and comparing content
sources.

In other work, reinforcement learning has shown promising results for learning
and improving open-domain chatbots in recent work, e.g.,
\cite{Li2016EMNLP,Yu2016Strategy,Serban2017arXiv}.
A future direction is to use the learned scoring functions in reinforcement
learning to facilitate the learning process.
In the reinforcement learning setting, reward signals at the end of the
conversation are usually weak and delayed, leading to the issue of slow
learning, also known as the temporal credit assignment problem
\cite{SuttonPhDThesis}.
Reward shaping is a remedy to address this issue by introducing intermediate
reward signals to complement the original sparse reward signal
\cite{Mataric1994ICML,Randlov1998ICML,Ng1999ICML}.
Both the turn-level dialog acts and the linear scoring function proposed in our
work can be immediately used for reward shaping.
It is also straightforward to replace the BiLSTM with a single-direction LSTM 
to make it suitable for reward shaping.


\chapter{A Graph-Based Document Representation \\
	for Socialbot Dialog Control}
\label{chap5}
Carrying out multi-turn conversations grounded on a document is an important
capability for dialog systems.
Two key challenges for achieving this goal are machine reading and dialog
control.
Machine reading can be viewed as a process that automatically extracts
interesting information from unstructured text such that the extracted knowledge
can be used for specific end tasks
\cite{Etzioni2007MachineReading,Poon2010MachineReadingAtUW}.
Existing machine reading systems are applied to tasks such as 
document summarization \cite{Hickl2007DUC}, question answering
\cite{Clark2010MachineReading}, and information retrieval
\cite{Nishida2018CIKM}.
In practice, these tasks need to work as a coherent whole for a dialog system to
control the dialog flow in mixed-initiative conversations.
However, to the best of our knowledge, there is little study on bringing
together machine reading technologies for dialog control in open-domain dialog
systems.
Bridging machine reading and dialog control is especially crucial for
open-domain socialbots which intend to discuss the latest events with users.
Unlike task-oriented dialog systems which are usually backed by a
well-structured knowledge base that can be populated by domain experts based on
specific ontologies, socialbots need to continuously comprehend unstructured
documents (e.g., news articles) at scale.
In this chapter, to bridge the gap between machine reading and dialog control,
we propose a universal document representation using a graph
structure.
Natural language processing (NLP) methods can be used to analyze the document
and construct the graph representation.
Given this representation, mixed-initiative dialog strategies can be developed based on
searches and moves on the graph.
While the proposed document representation, its construction methods, and dialog
strategies are universal to many document types, we focus on news articles in
this chapter since chatting about news is demanding for socialbots.


In \S\ref{chap5:sec:related_work}, we first review methods used by prior
socialbots for chatting about news articles and discuss the connection of this
work to other research problems.
We then describe the proposed document representation for news articles in
\S\ref{chap5:sec:representation} and methods for constructing the
document representation in \S\ref{chap5:sec:construction}.
The mixed-initiative dialog strategies using the proposed document
representation are illustrated in \S\ref{chap5:sec:dialog_design}. 
Experiment results are presented in \S\ref{chap5:sec:experiments}.

\section{Related Work}
\label{chap5:sec:related_work}

\subsection{News article discussion in socialbots}
\label{chap5:ssec:news_discussion_socialbots}
To discuss news articles, existing socialbots have tried to read a summary (usually 4-5 sentences) of the
article that is either accompanied with the document
\cite{EmUEmersonbot2017,EmUIrisBot2018,CMUTartan2018} or automatically generated
from the article
\cite{UWSoundingBoard2017,UTrentoRovingMind2017,HWUAlana2017,HWUAlana2018}.
While there is a lot of work on automatic summarization (see a comprehensive
review in \cite{Nenkova2011Summarization}), Sounding Board
\cite{UWSoundingBoard2017} has used the TextRank algorithm
\cite{Mihalcea2004EMNLP,Barrios2016arXiv} 
and Alana \cite{HWUAlana2017,HWUAlana2018} has used the LexRank algorithm
\cite{Erkan2004JAIR}, 
both of which are based on unsupervised extractive summarization.
In terms of dialog control, most socialbots read the entire summary in one turn.
The problem is that there is no actual interaction involved in this type of
conversation segment, and the user may be interested in different information
about the news.
To address this issue, Alana \cite{HWUAlana2018} has used an
information-retrieval method that retrieves a summary sentence in each turn
based on the named entities, noun phrases, n-grams in the current user utterance.
Nevertheless, using the summary alone for news discussions still has potential
issues.
As the conversation evolves, details in the news article may also be useful to
present.
Similarly, the information requested by the user may not be included in the
summary, but could be provided somewhere else in the article.
In this work, the proposed dialog strategy grounds the discussion in a
graph-structured document representation of the article, not just a few summary
sentences.

Some socialbots have built question-answering components for news articles using
either heuristic rules \cite{CMUMagnus2017} or information extraction methods
\cite{CTUAlquist2017}.
A problem with this strategy is that it solely relies on user initiative.
In actual socialbot conversations, many users do not initially ask questions and
instead engage in the conversation as responsive recipients who simply use
tokens such as ``{\it yeah}'' and ``{\it okay}'' to signal they are listening,
comprehending, or agreeing, until the conversation reaches a point when the user
elicits additional information or makes comments on some aspects of the news.
Therefore, the bot needs to pave the path for in-depth discussions and maintain
user engagement during the conversation, while being able to handle user
initiatives properly.
In our approach, the proposed document representation enables a mixed-initiative
dialog strategy, which takes into account both user questions and bot moves that
push the conversation forward.

Extending the conversation using relevant Twitter comments on a news article is
used in \cite{RPIWiseMacaw2017,UCDavisGunrock2018}.
However, online comments on social media such as Twitter and Reddit may target a
specific part of the article, and it is not straightforward to link them to the
corresponding article sentences.
Thus, these comments may not work well when the news article is presented
progressively, since the corresponding information may have not been presented to the
user in the conversation.
In our approach, we explore a different source, i.e., comments from other
conversations.
While they can be collected from interactions with live users, in this work,
comments are crowdsourced with a socialbot conversation setup which makes it
possible to align the collected comments with article sentences.
Although it has been studied in other applications of conversational artificial
intelligence (AI)
\cite{Lasecki2013UIST,Huang2015HCC,Huang2016CHI,Swaminathan2017UIST,Huang2017CIC,Huang2018CHI},
the use of crowd-powered method for conversational comment collection is new.

\subsection{Connection to other research problems}
\label{chap5:ssec:connection}
A number of connections can be drawn between our work and other research
problems in the literature of AI and NLP.  

{\bf Storytelling}:
Presenting a news article in a conversation is similar to storytelling (or
narrative) to some extent.
The study of storytelling in human-human conversations has a long history in the
field of conversation analysis
\cite{Jefferson1978,Labov1999,Norrick2000,Mandelbaum2013,Ruhlemann2013}.
Given the prevalence of narrative in human life, storytelling has also been
brought into the field of AI and NLP, triggering a substantial amount of
research interest \cite{Mateas1999AAAI}.
One line of research has focused on understanding stories, notably, the early work
by Schank and Abelson using scripts \cite{Schank1977Script} and the recent work
by Chambers and Jurafsky using narrative chains \cite{Chambers2008ACL}.
Another type of work is closely related to the field of natural language
generation, making advances in story generation using narrative planning (e.g.,
\cite{Riedl2010JAIR}), based on semantic representation of narrative (e.g.,
\cite{Rishes2013ICIDS}), grounded on images (e.g., \cite{Huang2016NAACL}), etc.
A series of work has also taking into account the involvement of a user, with
application to interactive storytelling and computer games, e.g.,
\cite{Cavazza2002IEEE,Thue2007AIIDEC}.
There are some attempts at adding the storytelling capability to conversational
AI systems.
For example, Bickmore and Cassell \cite{Bickmore1999AAAI} have developed a
conversational system for the real estate sales domain, using conversational
storytelling to help build rapport and trust with the prospective buyer.
For socialbots, Slugbot \cite{UCSCSlugbot2018} manually annotated stories in
chunks such that each chunk ends with a natural pausing point.  
They additionally append tag questions (e.g. ``{\it didn't it?}'') to the end of
each chunk to implicitly discourage the user from asking the bot hard questions
they are not capable of answering.

{\bf Document-Grounded Conversations}:
Recently, there is a surge of research on neural conversation models
\cite{Vinyals2015ICML,Shang2015ACL,Sordoni2015NAACL,Li2016EMNLP,Serban2017AAAI},
which learn dialog strategy based on a corpus of conversations.
However, researchers often find that these models generate
responses without grounding on structural knowledge or textual documents.
A knowledge-grounded model is proposed in \cite{Ghazvininejad2018AAAI} that can
use relevant documents (e.g., Foursquare tips, Amazon Reviews) to steer the
response generation.
Zhou et al.\ \cite{Zhou2018EMNLP} curate a dataset of document-grounded
conversations, asking two humans to chat about a Wikipedia article about a
specific movie.
They show that a neural conversation model that is grounded on the document
generates better responses than a model that is not.
In both \cite{Ghazvininejad2018AAAI} and \cite{Zhou2018EMNLP}, the unstructured
textual document is represented as continuous embeddings via the neural
conversation model.
Since there is no corpus available for socialbot conversations on news
discussions, neural conversation models are not a viable approach for the
problem studied in our work.
Instead, we construct a graph-structured document representation and propose a
mixed-initiative dialog strategy that grounds the news discussion on the
document representation.

{\bf Question Generation}:
There is a plenty of work in question generation associated with different
applications, particularly education settings for strategies in tutorial dialogs
and virtual agent settings for asking and answering questions (see discussions
in \cite{HeilmanPhDThesis,Piwek2012QGen} and references therein).
In our work, the bot uses automatically generated questions in two ways: 
to facilitate question answering and to guide the discussion.
In the second case, the bot selects a generated question at each turn to signal
turn taking and maintain user engagement.
The question is thus presented in the form of an introductory clause,
which allows the user to know what the bot wants to tell next and to decide
whether they want to follow the dialog flow or take the control of the dialog.
This new use of question generation is different from all previous question
generation work.

\section{Graph-Based Document Representation}
\label{chap5:sec:representation}
The socialbot's capability of discussing a news article is largely constrained by
its representation of the article.
If a socialbot represents an article as a sequence of independent sentences,
it can do little more than present these sentences turn by turn and use a
generic question (e.g., ``{\it do you want to hear more?}'') at each turn to
signal turn taking.
Clearly, this type of interaction is not satisfying.
First, not all sentences in the article will be equally important to the
listeners.
Second, the socialbot will fail when the user wants to take the initiative for
requesting specific details (e.g., ``{\it tell me more about the robot}'') or
asking questions (e.g., ``{\it how many people are affected}'').
Third, the repetitive nature of using a generic confirmation is tedious and
likely to discourage a user from engaging.
To enable a coherent and engaging conversation on a news article,
additional knowledge needs to be accessible in the document representation.

\begin{figure}[!t]
	\centering
	\includegraphics[width=0.9\textwidth,trim=0cm 0cm 4cm 3cm,clip]{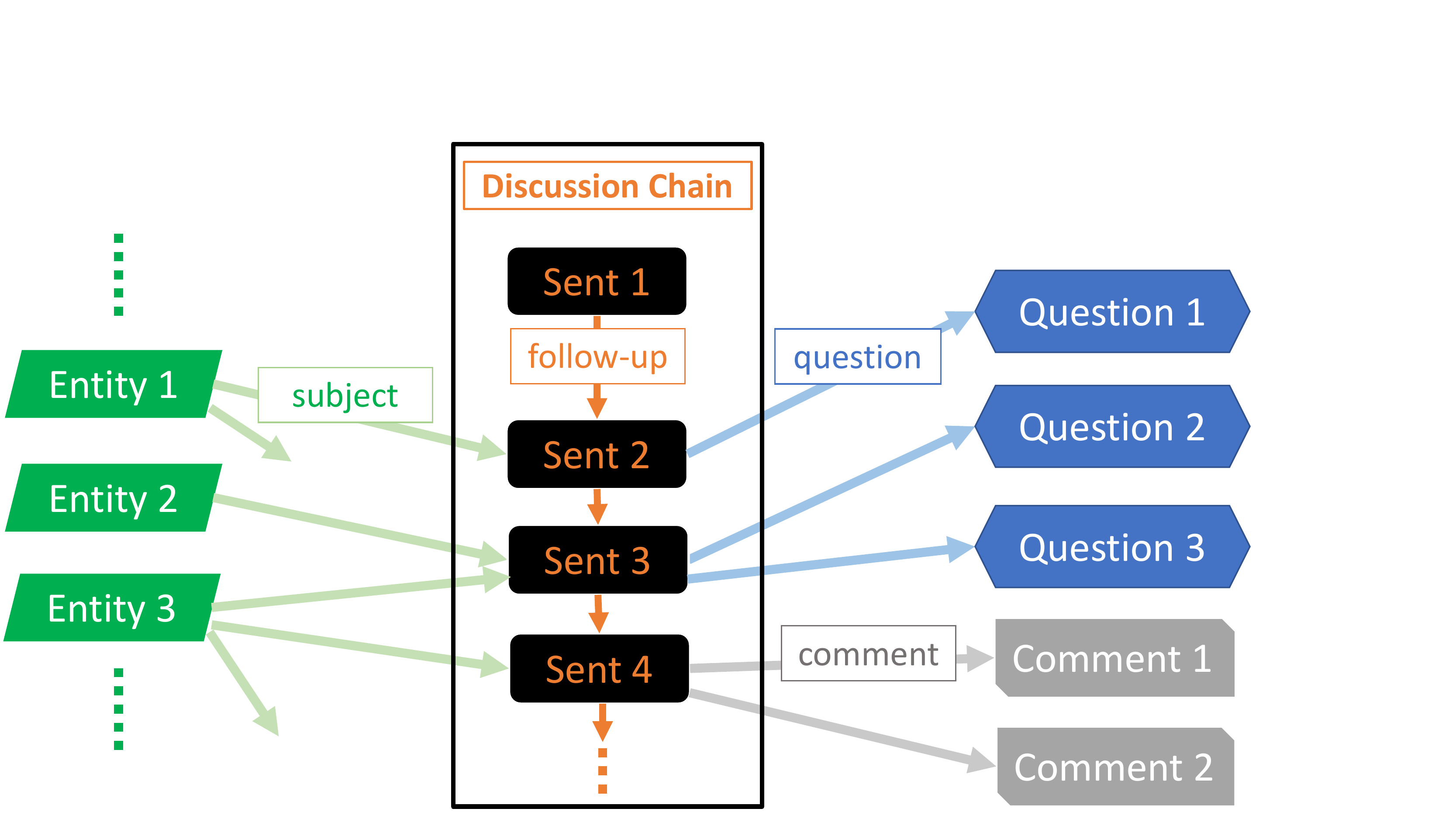}
	\caption{The graph-based document representation for news articles.}
	\label{chap5:fig:graph_repr}
\end{figure}

With these considerations in mind, we propose a document representation 
for news articles as illustrated in Fig.\,\ref{chap5:fig:graph_repr}.
The representation is a graph consisting of sentence and entity nodes extracted
from the article, augmented with a collection of question nodes and comment
nodes about the article.
Below we discuss how these nodes are connected in the graph.

{\bf Discussion Chain}:
A discussion chain of sentence nodes is created for the socialbot.
Consecutive sentence nodes in the chain are connected via a directed edge to
indicate the \texttt{follow-up} relation.
The discussion chain always starts from the title of the article.
It allows the socialbot to present article sentences in a coherent order during
the conversation and skip redundant or less important sentences.
As an example, below we show the title sentence (S1) and four body sentences
(S2)--(S5) of a news article.\footnote{%
	\url{https://tcrn.ch/2R2NBtL}
}
\begin{enumerate}[label={(S\arabic*)},leftmargin=1.2cm]
	\item {\it Google has launched its first clean energy project in Taiwain, its
		first in Aisa.}
	\item {\it Google has launched its first clean energy project in Asia.}
	\item {\it The company announced today that it struck a long-term
			agreement to buy the output of a 10-megawatt solar array in Tainan
		City, Taiwan.}
	\item {\it Google already has solar and wind projects across North and
		South America, as well as Europe.}
	\item {\it The agreement means Google will get a long-term, fixed electricity
		price for its operations in Taiwan.}
\end{enumerate}
A good discussion chain for these five sentences could be
(S1)-(S3)-(S4)-(S5).
(S2) should be skipped since it duplicates information in the title (S1). 
The chain (S1)-(S5)-(S3)-(S4) is not coherent since it is unclear what ``the
agreement'' refers to in (S5) without knowing (S3).
There can be multiple reasonable discussion chains for an
article, e.g., the chain (S1)-(S3)-(S5)-(S4) would also work.

{\bf Subject Edges}:
A sentence node is connected to an entity node via a \texttt{subject} edge,
indicating that the entity is a subject of the sentence.
For the aforementioned example article, (S1), (S3) and (S4) are all connected to
the entity node ``Google'';
(S2) is also connected to ``Google'' after resolving the coreference between
``the company'' and ``Google'';
and (S5) is connected to ``the agreement'' node.
This \texttt{subject} edge allows the socialbot to retrieve sentences when the
user interrupts the default discussion chain and requests specific details
about an entity during the conversation.
For example, given the discussion chain (S1)-(S3)-(S4)-(S5), if the user
makes a request ``{\it tell me more about the agreement}'' after the socialbot
presents (S3), the socialbot can skip (S4) and present (S5) directly.

{\bf Question Edges}:
A question node is connected to a sentence node via a \texttt{question} edge
indicating that the question can be answered by the sentence.
For instance, the question node ``{\it what does the agreement mean}'' is
connected to (S5) in the document representation of the aforementioned example
article.
Besides answering user questions about the article, the bot can select a
question connecting the target sentence as a transition strategy for introducing
the target sentence during the conversation.
For the example question node and (S5), the socialbot can ask the user ``{\it do
you want to know what the agreement means?}''
In this way, the socialbot avoids repetitive use of generic turn-taking
questions such as ``{\it do you want to hear more?}''


{\bf Comment Edges}: 
Comment nodes are connected to corresponding sentence nodes via \texttt{comment}
edges if the comment is a response to the sentence.
In this way, the socialbot can bring in opinion statements by presenting a
comment during the conversation.
Comments need to be aligned with a sentence so that the socialbot can decide the
timing to present them.
For example, the comment ``{\it that's a good deal for Google}'' is connected to
(S5), but not (S4).

\section{Document Representation Construction}
\label{chap5:sec:construction}
In this section, we present methods used for constructing the proposed document
representation.
First, in \S\ref{chap5:ssec:preprocessing}, we describe the pipeline for news
article collection and pre-precessing steps that create sentence nodes, entity
nodes, and \texttt{subject} edges.
Then, the annotations and models for creating the discussion chain are
described in \S\ref{chap5:ssec:model_discussion_chain}. 
The proposed method for creating question nodes and corresponding
\texttt{question} edges is illustrated in
\S\ref{chap5:ssec:algo_iclause_generation}.
Lastly, the method for collecting comments and creating corresponding
\texttt{comment} edges is discussed in
\S\ref{chap5:ssec:data_comment_retrieval}.

\subsection{News article collection and pre-processing}
\label{chap5:ssec:preprocessing}
We crawl news headline submissions in three subreddits 
(\textsl{technology}, \textsl{politics}, \textsl{news}) using
PRAW\footnote{\url{https://praw.readthedocs.io}}, which is a python wrapper of
the Reddit API.
Reddit submissions come with karma scores that reflect the community
endorsement, which we use to filter less popular news articles.
Since the Reddit submissions only contain the URL to the news article, we use
the Newspaper3k package\footnote{\url{https://github.com/codelucas/newspaper/}} to fetch
the article and extract the text in it.

{\bf Sentence Node Creation}:
The extracted text body is split into sentences and annotated using the Stanford
CoreNLP toolkit (v3.9.1) \cite{CoreNLP}.
Note that the extracted text may inevitably include unwanted text (e.g.,
advertisements, image captions, stock information), ill-formed paragraph (e.g.,
no space after the punctuation), duplicated sentences, etc.
We develop an in-house post-processing pipeline to filter unwanted sentences
and skip articles with too many extraction errors. 
In addition, we remove sentences that are too short (less than 8 tokens) or too
long (more than 80 tokens) for a turn.
We also exclude sentences containing URLs or characters that are not friendly to
the text-to-speech synthesizer, such as backslash, angle brackets, hashtag, etc.
We further exclude sentences that are part of a multi-sentence quote since they
usually need to be presented together in a turn to be meaningful.
Quotes are also detected by the Stanford CoreNLP toolkit.



{\bf Entity Node Creation}:
To create entity nodes, we extract entity instances in the article using
the named entity recognizer in the Stanford CoreNLP toolkit and the entity
linking tool provided by the DBpedia Spotlight~\cite{Daiber2013ISS}.
We further extract noun phrases in article sentences as entity instances based
on the constituency parsing tree generated from the Stanford CoreNLP toolkit.
To merge entity instances that refer to the same entity node, we use the
coreference resolution model in the Stanford CoreNLP toolkit and a coreference
resolution pipeline developed using heuristics about person and organization in
order to handle mentions by first/last name or acronyms.
After coreference resolution, we obtain a set of unique entity nodes.

\begin{figure}[t]
	\centering
	\includegraphics[width=\textwidth]{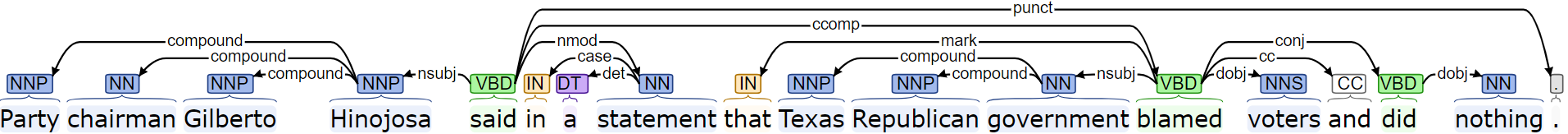}
	\caption{A sample sentence with its dependency parsing and POS tagging output.
		The predicate is the word ``said''.
	}
	\label{chap5:fig:sample_sentence_dependency}
\end{figure}

{\bf Subject Edge Creation}:
To create \texttt{subject} edges that connect sentence nodes with corresponding
entity nodes, we use the Stanford CoreNLP toolkit to parse individual sentences
to obtain their dependency structure and part-of-speech (POS) tags.
The toolkit uses a neural transition-based dependency parsing
model \cite{Chen2014EMNLP} and follows the universal dependency
relations\footnote{\url{http://universaldependencies.org/docsv1}} defined
in \cite{UniversalDependenciesV1}.
An example sentence with its dependency parsing and POS tagging output is shown
in Fig.\,\ref{chap5:fig:sample_sentence_dependency}. 
The nominal subject (\textsf{nsubj}) and passive nominal subject
(\textsf{nsubjpass}) of the predicate are extracted with their modifiers.
Clausal subjects (\textsf{csubj} and \textsf{csubjpass}) are ignored currently
since they are usually not associated with an entity in the graph.
To associate the subject with the entity nodes, we examine whether the head word
of the subject is in the span of the entity.
In the example in Fig.\,\ref{chap5:fig:sample_sentence_dependency}, the head
word of the nominal subject is ``Hinojosa'', which is associated with the entity
node ``Gilberto Hinojosa''.

\subsection{Discussion chain creation}
\label{chap5:ssec:model_discussion_chain}

Based on our observation that it is useful to skip some sentences in an article,
we developed a crowdsourced annotation collection method for creating the
discussion chain.
Then, we train models using the collected data to automatically generate
the discussion chain for unseen articles.

\begin{figure}[!t]
	\centering
	\includegraphics[width=\textwidth,trim=1cm 0cm 1cm 0cm,clip]{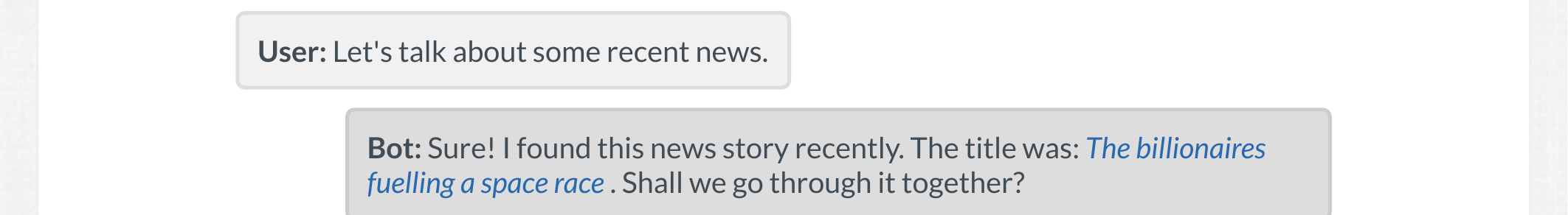}
	\\
	\vspace{-0.2ex}
	\includegraphics[width=\textwidth,trim=1cm 1cm 1cm 0cm,clip]{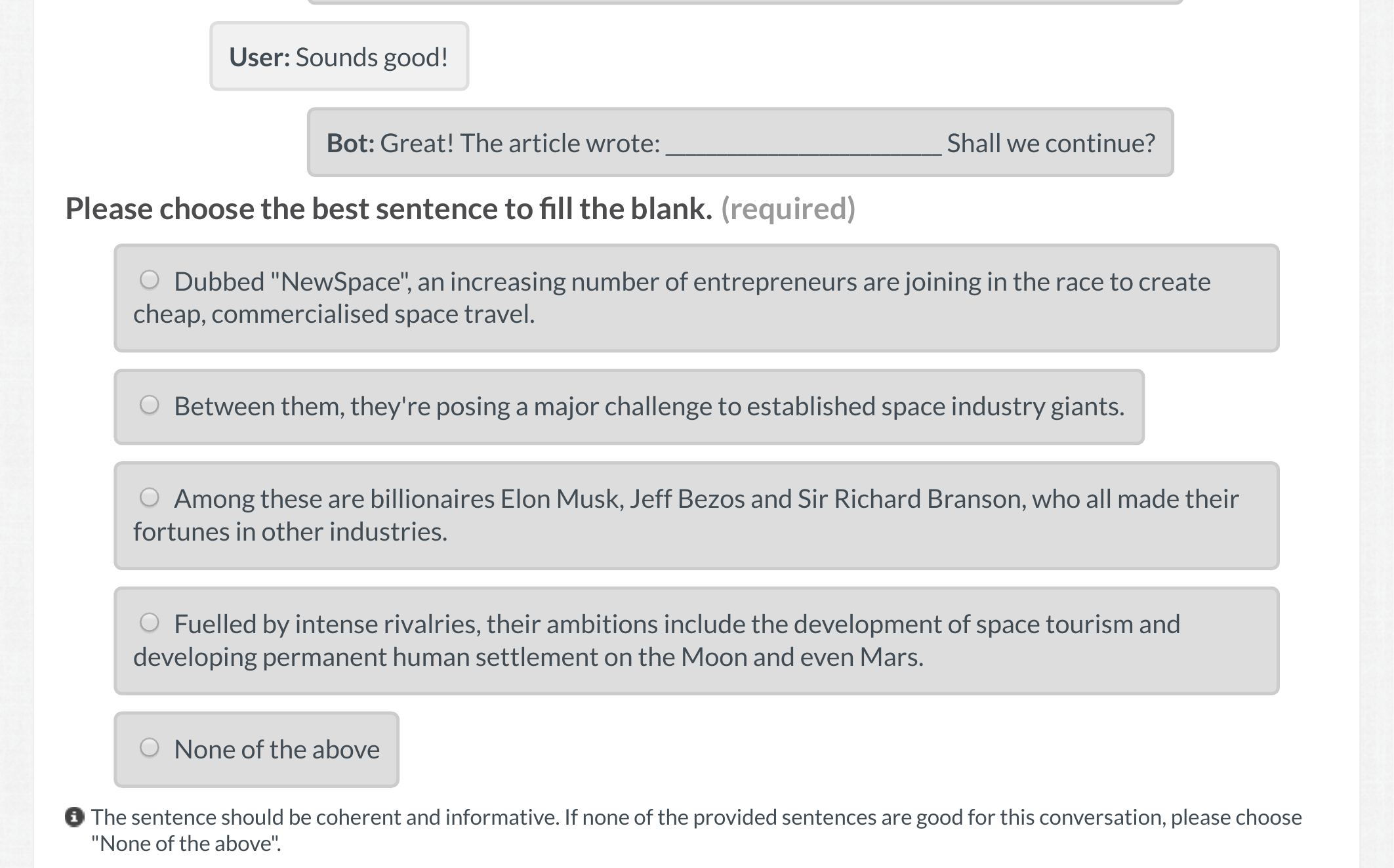}
	\caption{A sample annotation unit for the lead sentence selection.}
	\label{chap5:fig:sample_annotation_unit_lead_sentence}
\end{figure}

\begin{figure}[!t]
	\centering
	\includegraphics[width=\textwidth,trim=0.05cm 0.2cm 0cm 0.5cm,clip]{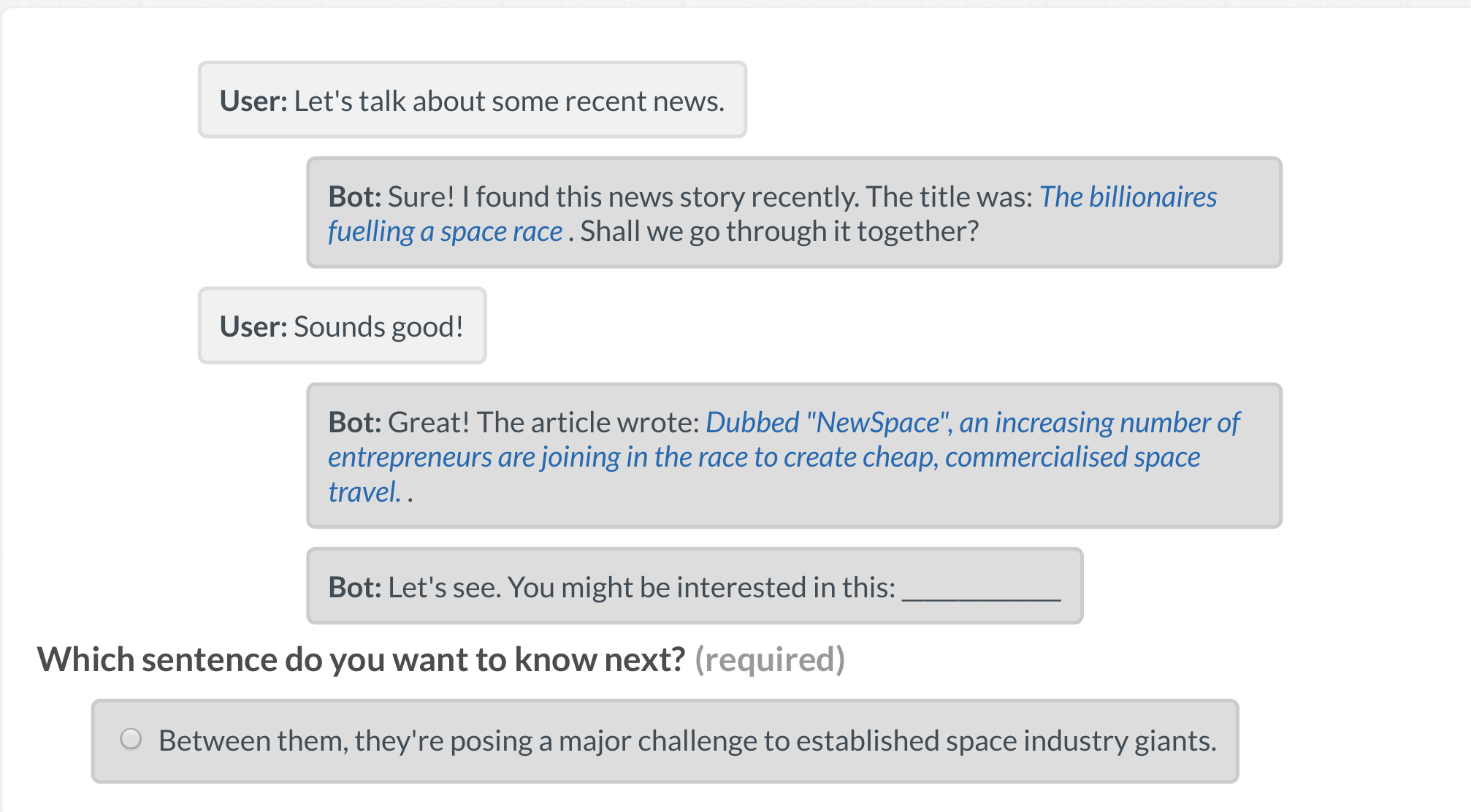}
	\\
	\includegraphics[width=\textwidth,trim=0cm 1cm 0cm 0cm,clip]{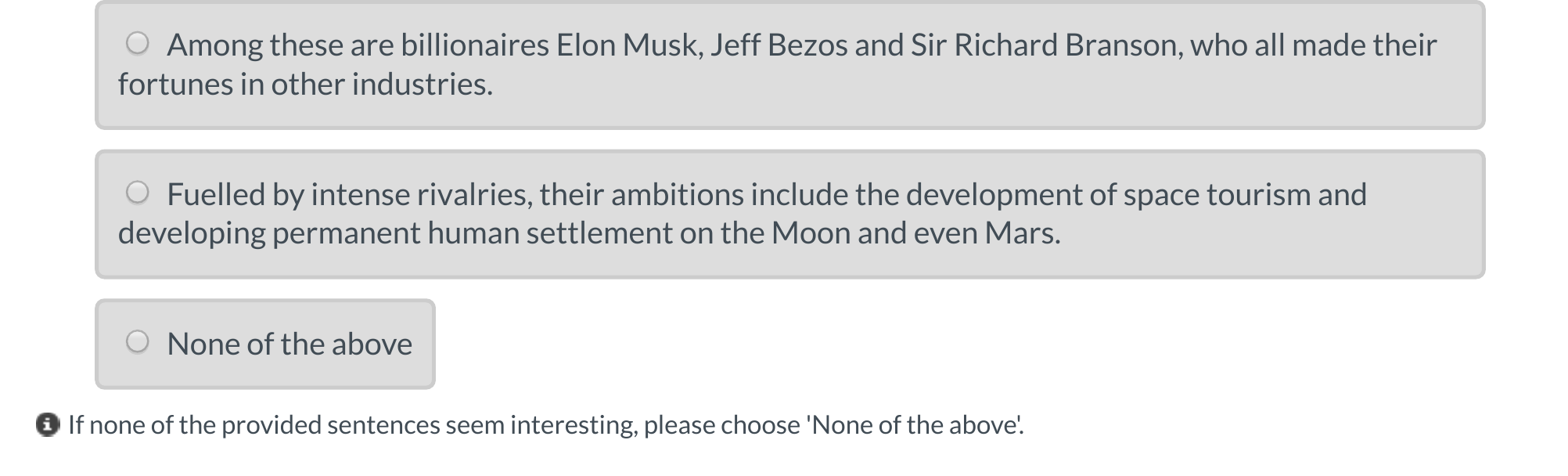}
	\caption{A sample annotation unit for the follow-up sentence selection.}
	\label{chap5:fig:sample_annotation_unit_followup_sentence}
\end{figure}

{\bf Annotation Interface}:
Rather than asking a worker to read a complete article and create a discussion
chain, we break down the annotation task into incremental steps.
First, we ask the worker to fill the blank in a socialbot conversation snippet
by choosing the best sentence from four candidates.
A fifth option ``none of the above'' is provided if there is no good candidate.
A sample annotation unit is shown in
Fig.\,\ref{chap5:fig:sample_annotation_unit_lead_sentence}.
The four candidates are the first four sentences after the pre-processing steps
described in \S\ref{chap5:ssec:preprocessing}.
This round of data collection provides partial discussion chains consisting of 2
sentence nodes.
Then, we create a new task to ask a worker to choose the best follow-up sentence
given a partial discussion chain of length 2, as shown in
Fig.\,\ref{chap5:fig:sample_annotation_unit_followup_sentence}.
In this task, the choices are three sentences right after the sentence at the
end of the partial discussion chain in the pre-processed article.
A last option ``none of the above'' is also provided if the worker thinks none
of the candidates are good.
In principle, we can keep repeating the steps until the discussion chain is
complete, i.e., we reach the end of the article.
This incremental annotation method has the advantage of providing a simple task
that is well suited to crowdsourcing, unlike
asking the worker to create a discussion chain from scratch.
In this setting, a discussion chain is created in a streamlined fashion by a
sequence of workers.
Moreover, the annotation interface encourages the worker to consider their choice
in the context of our envisioned socialbot interactions.
Currently, our study only obtains partial discussion chains of lengths 2 and 3,
but it is straightforward to repeat the same process to extend the partial
discussion chain, which we leave for future work.
Note the model we developed using these annotations is able to build discussion
chains of arbitrary length, which will be discussed at the end of this
subsection.

\begin{figure}[t]
	\centering
	\includegraphics[width=0.9\textwidth,trim=0cm 16cm 9cm 0cm,clip]{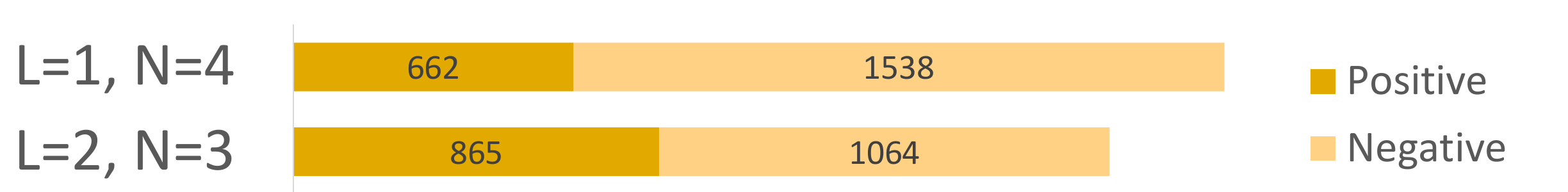}
	\caption{The numbers of positive and negative sentence chains in the collected
	data.}
	\label{chap5:fig:sent_chain_distribution}
\end{figure}

{\bf Data}:
Annotations on crawled news articles are collected using the FigureEight
platform.\footnote{\url{https://www.figure-eight.com}}
We collect 5 judgments for each annotation unit and pay the worker at the rate
of 3 cents per judgment.
The worker's geographical location is restricted to United States, Canada, and
United Kingdom, considering that most Reddit users and news article submissions
are from these countries.
In an annotation unit, each candidate sentence appended to the presented partial
discussion chain of length $L$ forms a candidate sentence chain.
If the candidate sentence receives at least 2 votes from workers, the
corresponding sentence chain is positive, i.e., it forms a new partial
discussion chain of length $L+1$. 
Otherwise, the corresponding sentence chain is negative.
In total, we use 550 articles for data collection.
We randomly split these articles into training, validation, and test sets with a
ratio of 3:3:1.
As described above in the annotation interface, we carry out two rounds of data annotation.
In the first round with $L=1$, there are $N=4$ candidate sentences.
In the second round with $L=2$, there are $N=3$ candidate sentences.
The distributions of positive and negative sentence chains are shown in
Fig.\,\ref{chap5:fig:sent_chain_distribution}.

{\bf Model}:
To classify a candidate sentence chain as positive or negative, we train a
binary logistic regression model which uses features that characterize the
sentence chain. 
Several features are compared in our experiments.
The logistic regression model is trained with $\ell_2$ regularization,
and the final model is the one with the highest area under the receiver
operating characteristic curve (ROC-AUC) for the validation set.
The model can be trained independently for $L=1$ and $L=2$.
Alternatively, we can use the same model regardless of $L$ by training it on the
joint data.
In this setting, the model can be used to extend partial discussion chains of
arbitrary lengths.

\subsection{Question generation} 
\label{chap5:ssec:algo_iclause_generation}

We propose a method using the dependency structure to automatically generate
question-answer pairs for individual sentences in the article.
Consider the example sentence in
Fig.\,\ref{chap5:fig:sample_sentence_dependency}, 
we can generate a question ``{\it what did Party chairman Gilberto Hinojosa say
in a statement?}''
The question can be viewed as a sequence of sentence constituents following the
interrogative word, i.e., 
\textlangle what, Party chairman Gilberto Hinojosa, said, in a statement\textrangle,
where ``said'' is the predicate (\textsf{root}) of the sentence, ``Party
chairman Gilberto Hinojosa'' is the subject (\textsf{nsubj}), ``in a statement''
is the nominal modifier (\textsf{nmod}), and ``what'' is the question type
corresponding to the clausal complement (\textsf{ccomp}), which is the answer of
the question.

\begin{table}[!t]
	\centering
	\begin{tabular}{@{}l@{\hspace{0.1cm}}l|p{3cm}|p{7.3cm}}
		\hline\hline
		& {\bf Dependent} & {\bf Question Types}  & {\bf Description} \\
		\hline
		\dag 	& \textsf{root} 								& what 
		& the {\it noun} predicate \\
		\hline
		* 		& \textsf{root}/\textsf{neg}  	& whether 
		& the negation modifier of the {\it verb} predicate \\
		\hline
		* 		& \textsf{root}/\textsf{dep} 		& what
		& the unspecified dependent of the {\it verb} predicate \\
		\hline
		& \textsf{root}/\textsf{ccomp} 				& what 
		& the clausal complement of the predicate (functioning as an object) \\
		\hline
		& \textsf{root}/\textsf{dobj} 				& what, who
		& the direct object of the predicate \\
		\hline
		& \textsf{root}/\textsf{nmod-tmod} 		& when
		& the temporal modifier of the predicate \\
		\hline
		& \textsf{root}/\textsf{nmod} 				& how, what, when, where, why
		& the nominal modifier of the predicate (functioning as an adverbial) \\
		\hline
		& \textsf{root}/\textsf{advcl} 				& how, what, when, why 
		& the adverbial clausal modifier of the predicate \\
		\hline
		& \textsf{root}/\textsf{xcomp}  			& what, how
		& the open clausal complement of the predicate \\
		\hline
		& \textsf{root}/\textsf{xcomp}/\textsf{ccomp} 		& what
		& the clausal complement of \textsf{root}/\textsf{xcomp} \\
		\hline
		& \textsf{root}/\textsf{xcomp}/\textsf{dobj} 			& what, who
		& the direct object of \textsf{root}/\textsf{xcomp} \\
		\hline
		& \textsf{root}/\textsf{xcomp}/\textsf{nmod}  		& how, what, when, where, why 
		& the nominal modifier of \textsf{root}/\textsf{xcomp} \\
		\hline
		& \textsf{root}/\textsf{nsubj(pass)}/\textsf{nummod} 			& how many
		& the numeric modifier of a (passive) nominal subject \\
		\hline
	\end{tabular}
	\caption{The subset of dependents and corresponding question types considered
		for questions generation.
		The dependent is expressed using the path of dependency relations.
		\dag: only used for sentences with a noun predicate.
		*: only used for sentences with a verb predicate.
	}
	\label{chap5:tab:question_edges}
\end{table}

Given a sentence, the proposed algorithm analyzes its dependency structure rooted
from the sentence predicate and generates question-answer pairs for each
dependent which has pre-defined question templates.
Currently, there is no template developed for following cases:
1) sentences with a clausal subject (\textsf{csubj} and \textsf{csubjpass}),  
2) sentences with a non-clausal subject that has a negation modifier
(\textsf{neg}) or a preconjunct (\textsf{cc:preconj}) modifier,
and 3) sentences with a predicate that is not a verb, noun, or
adjective.
These cases constitutes around 17.7\% of all sentences in our data after the
pre-processing steps described in \S\ref{chap5:ssec:preprocessing}.

As pointed out in \cite{Vanderwende2008Importance}, a question generation system
should take into account the importance of the generation questions given the
context, e.g., their utility or informativeness.
In this work, we limit the set of generated questions given the sentence
by constraining the generation to only consider a subset of dependents within
three steps from the notional sentence root, i.e., the predicate, the dependents
of the predicate, and the grand-dependents of the
predicate.\footnote{Anecdotally, we found that this constraint results in
questions that are easier to follow in spoken form.}
The dependent, together with its modifiers, is treated as the answer for the
generated question.
Table\,\ref{chap5:tab:question_edges} summarizes the subset of dependent types
used in our work.
The dependency paths originating from the notional sentence root are described
using the property path syntax of the SPARQL query
language.\footnote{\url{https://www.w3.org/TR/sparql11-query}}
For some dependents, questions are only generated for sentences with a predicate
with specific POS tags.
Some dependents have multiple question types, which are determined based on
further analysis of the dependency structure and entity recognition results.
For example, the algorithm generates a who-question or what-question for
\textsf{root}/\textsf{dobj} by checking whether the object is a person entity or
not.
The question type for a nominal modifier (\textsf{nmod}) is determined by its
case-marking dependent (\textsf{case}).
The question type of an adverbial clausal modifier (\textsf{advcl}) is
determined by its marker (\textsf{mark}), which is usually a subordinating
conjunction such as ``because'', ``after''.
In our implementation, the RDFLib\footnote{\url{https://rdflib.readthedocs.io/}}
is used to convert the dependency parsing results to a semantic graph according
to the resource description framework (RDF), where the edge from the governor
node to the dependent node is labeled by the corresponding dependency relation.

After the target dependent and the corresponding question type is determined,
the algorithm retrieves a sentence plan which specifies the sequence of
constituents for the question.
The constituent is represented by the dependency path from the sentence root to
the its head word.
For example, a sentence plan can be specified as
``what \textsf{root/nsubj} \textsf{root} \textsf{root}/\textsf{dobj}
\textsf{root}/\textsf{nmod}''.
To convert the sentence plan to an interrogative form, the algorithm checks
whether an axillary verb should be inserted and the verb conjugation should be
applied.
The verb conjugation is implemented using the pattern.en
library.\footnote{\url{https://www.clips.uantwerpen.be/pages/pattern-en}}
Finally, to realize the sentence plan, constituents are replaced by their
textual form generated based on a subset of their modifiers in the original
sentence with pre-defined ordering.

The question node stores the sentence plan and the textual form of individual
constituents, as well as the dependency path for the answer.
For each generated question node and the corresponding sentence node, a
\texttt{question} edge is created between them in the document
representation.
Note multiple questions may be generaetd for a sentence according to the
proposed algorithm. 
On average there are 1.4 questions generated for each sentence in our
data.

\subsection{Comment collection}
\label{chap5:ssec:data_comment_retrieval}

\begin{figure}[!t]
	\centering
	\includegraphics[width=\textwidth,trim=0cm 0cm 0cm 0.5cm,clip]{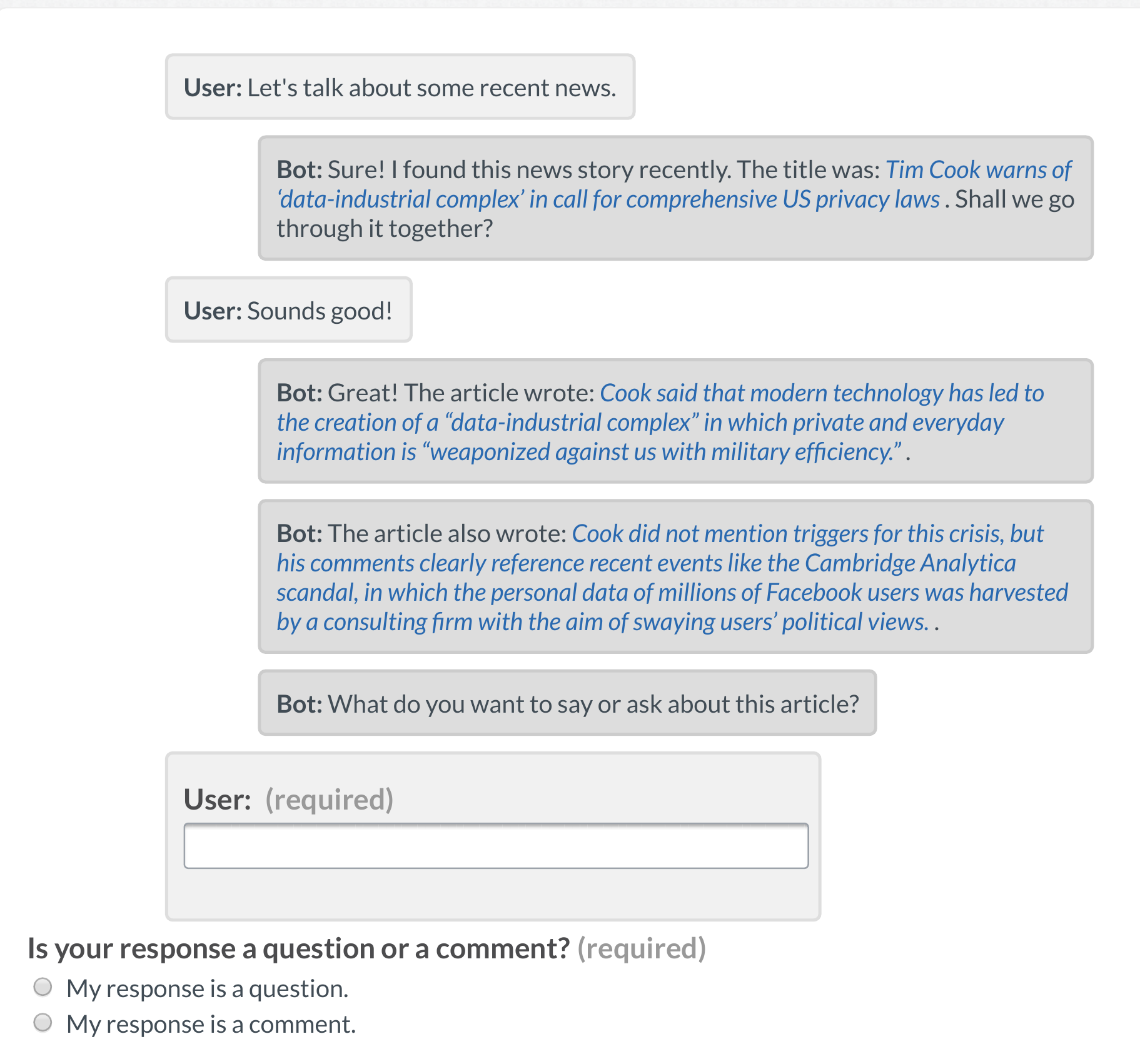}
	\caption{A sample unit for comment collection.}
	\label{chap5:fig:comment_collection}
\end{figure}

We propose a method to collect comments on news articles that can
later be used in socialbot conversations.
Fig.\,\ref{chap5:fig:comment_collection} shows an example annotation unit for
data collection on the FigureEight platform.
We currently pay 4 cents per response and collect 5 responses per
annotation unit.
In each annotation unit, we present the first few sentences of the discussion
chain in the context of a socialbot conversation snippet, skipping the
intermediate turn between article sentence nodes to reduce the reading effort
for the workers.
The worker is asked to record whether they make a comment or ask a question,
so that we can extract statement comments from the collected responses.
A filtering pipeline is developed to avoid generic responses (e.g., 
``{\it that's interesting}'') or single-word responses (e.g., ``{\it cool}'').
In this work, collected questions are used for analysis on conversational
question answering.
There are 54\% worker responses are questions.
Among these questions, the most frequent leading words are ``what'' (27\%)
and ``how'' (16\%).
Remaining leading words include ``why'' (7\%), ``who'' (6\%), ``will'' (5\%), ``is'' (5\%), 
``where'' (3\%), ``which'' (3\%), ``was'' (3\%), ``can'' (3\%), ``are'' (2\%),
``were'' (2\%), ``when'' (2\%), ``did'' (2\%),
``does'' (1\%), ``have'' (1\%), ``has'' (1\%), ``do'' (1\%),
and a few other rare words.

The collected responses are connected to the last sentence node presented in the
conversation snippet via the \texttt{comment} edge.
At a socialbot turn which informs a sentence node, the socialbot can check if
there is a \texttt{comment} edge and determine whether to start an
opinion-exchange segment.

{\bf Crowd-powered Conversational AI}:
In response to the challenge of fully-automated open-domain conversational AI
systems, a line of work proposes to power the conversational AI using a crowd of
human
actors~\cite{Lasecki2013UIST,Huang2015HCC,Huang2016CHI,Swaminathan2017UIST,Huang2017CIC,Huang2018CHI}.
A socialbot using the method described here for comment collection can be
thought as a crowd-powered conversational AI.
For a live socialbot, we can imagine collecting user comments during the
conversation for enriching the document representation.
Since the socialbot in this work is currently hosted in a laboratory
environment, we leave this for future work.

{\bf Issues of Social Media Comments}:
Since we crawl news articles from Reddit, they come with Reddit user comments.
We explored the use of Reddit comments, but chose not to use them for the
following reasons.
First, Reddit comments tend to be long and provide more information than is
suitable to be communicated in a single socialbot turn.
Second, there is frequent use of external links and text markup in Reddit
comments, which are not easy to communicate in the voice-based socialbot
conversation.
Third, there is more offensive language in Reddit comments than responses
collected from paid crowd workers.
Last but not least, some Reddit comments may target a specific part of the
article,
and it is not straightforward to link the comment to the corresponding
article sentences.
It would be confusing if the socialbot brings up a comment about a specific
article sentence that has not been informed to the user.
This is not an issue in our crowdsourcing setup, because they are not presented
with the whole article.
As discussed in \S\ref{chap5:ssec:news_discussion_socialbots}, some socialbots
have used Twitter comments for extending discussions on a news article.
While the first issue is less severe for Twitter comments since the length of
the comment is limited, they still suffer from other issues discussed
above.
In spite of these issues, it may be possible to better exploit social media comments
after careful filtering and in-depth language analysis, which calls for future
efforts along this direction.

\section{Mixed-Initiative Dialog Strategy}
\label{chap5:sec:dialog_design}
An advantage of the proposed document representation using a graph structure is
that the dialog control can be carried out through retrieving nodes or moving
along edges in the graph.
In this section, a mixed-initiative dialog strategy is proposed for discussing a
news article.
During a socialbot conversation, the subdialog about a news article starts with
the first node in the discussion chain, which is the article
title.
In general, the socialbot follows the chain during the conversation
until the user takes the initiative by asking a question or making an
information request about a specific entity.
The socialbot can also digress to present a comment and return to the
discussion chain afterwards.
In this section, we describe 
the dialog strategies for user initiative and system initiative
in
\S\ref{chap5:ssec:user_init_strategies}--\ref{chap5:ssec:system_init_strategies},
respectively.
A sample conversation is illustrated in \S\ref{chap5:ssec:sample_conversation}.

\subsection{Dialog strategy for user initiative}
\label{chap5:ssec:user_init_strategies}
User initiative impacts the response plan for the current bot
turn.
These response strategies are designed based on different types of user
requests.
When a response fails, the socialbot apologies and resorts to a
system-initiative strategy to push the conversation forward.

{\bf Entity-Based Sentence Retrieval}:
Given the user intent of requesting information about a specific entity, (e.g.,
``{\it tell me more about Amazon}''), the socialbot searches for the entity node
that matches the requested entity.
If an entity node is found, the socialbot retrieves corresponding sentence nodes
that are connected with the entity node via \texttt{subject} edges.
Sentence nodes that have already been presented in the conversation are
excluded.
The socialbot then selects a sentence node to present according to its dialog
policy.
This strategy fails if no entity node is found or sentence node is retrieved.

{\bf Question-Based Sentence Retrieval}:
For information-seeking questions about the article (e.g., ``{\it what did Jeff
Bezos say in the letter?''}), 
the socialbot searches for question nodes that match the user question and
retrieves the sentence node connected to the question node that gets the highest
match score.
A match score is computed by counting the number of matched non-stop words.
If no question node matches the user question, the socialbot tries to query an
external question-answering engine in case the question can be answered using
general world knowledge alone, e.g., 
``{\it what is net neutrality}'', ``{\it where is Yemen}''.
This strategy fails if both question answering attempts fail.

{\bf Comment Retrieval}:
For the opinion-seeking intent (e.g., ``{\it what do you think about this?}''),
the socialbot checks if there is any comment node connected to the last
presented sentence node via a \texttt{comment} edge. 
Comment nodes that have already been presented in the conversation are excluded.
The socialbot then selects a comment node to present according to its dialog
policy.
This strategy fails if no comment node is retrieved.

\subsection{Dialog strategy for system initiative}
\label{chap5:ssec:system_init_strategies}
The system-initiative strategies are used when the user does not take the
initiative or the user-initiative strategies fail in handling user requests.
These system-initiative strategies should not fail before the socialbot finishes
the discussion on the article.
For dialog control, the bot uses a hand-crafted decision tree to choose among
these strategies.

{\bf Discussion-Chain-Based Move}:
The socialbot proceeds to the next sentence node via the \texttt{follow-up} edge
along a discussion chain.
The sentence is presented at the current turn.

{\bf Question-Edge-Based Move}:
Rather than immediately moving to the next sentence in the discussion chain,
the socialbot can select a question generated for the sentence and present it in
a confirmation question, e.g., ``do you want to know \dots?''
In this case, the generated question is realized in a clause form rather than
the interrogative form.  
To select the best question for introducing the sentence, a hand-crafted
decision tree is developed primarily based on the dependency path of the
answer.
Priority is roughly organized into three tiers from high to low:
1) 
\textsf{root}/\textsf{neg},
\textsf{root}/\textsf{ccomp},
\textsf{root}/\textsf{dep},
\textsf{root}/\textsf{xcomp}/\textsf{ccomp},
\textsf{root}/\textsf{xcomp}/\textsf{dobj};
2) 
\textsf{root}/\textsf{dobj},
\textsf{root}/\textsf{advcl},
\textsf{root}/\textsf{nsubj}/\textsf{nummod},
and 3) others.
Note we do not use this move if the sentence subject contains certain pronouns
(we, I, this, it) since they are often confusing, e.g., the question ``do you
want to know what it's clear'' for the sentence ``it's clear we can't rely on
the old ways of doing things for this next generation.''

{\bf Comment-Edge-Based Move}:
When there are comment nodes connected to the current sentence node in the
discussion chain via \texttt{comment} edges, the socialbot can initiate a
multi-turn opinion exchange segment.
In this segment, the socialbot can elicit user opinions on the article sentence,
and then present the comment node in its second turn.

\subsection{A sample conversation}
\label{chap5:ssec:sample_conversation}

\begin{table}[!t]
	\centering
	\begin{tabular}{c|l|p{13cm}}
		\hline\hline
		{\bf Turn} & {\bf Speaker} & {\bf Utterance} \\ 
		\hline
		1 & User 	& let's chat about recent news \\
		2 & Bot 	& Sure! I found this news story recently. The title was: 
		{\it The billionaires fuelling a space race}.
		Shall we go through it together? \\
		3 & User 	& Sounds good! \\
		4 & Bot 	& Great! The article wrote: 
		{\it Dubbed ``NewSpace'', an increasing number of entrepreneurs are joining in
		the race to create cheap, commercialised space travel.}
		Let's see. 
		Do you know \uline{who are among these entrepreneurs}? \\
		5 & User 	& No, I did not. Please tell me. \\
		6 & Bot 	& {\it Among these entrepreneurs are billionaires Elon Musk, Jeff
			Bezos and Sir Richard Branson, who all made their fortunes in other
		industries.}
		Okay, what do you think about this? \\
		7 & User 	& I think a new space race is beginning. How about you?\\
		8 & Bot 	& I'm not sure, but someone said: \uwave{this probably won't happen
		in their lifetime.} Do you agree? \\
		9 & User 	& That's probably true. \\
		10 & Bot 	& Okay. Anyway, do you want to know \uline{how South
		African-born businessman Elon Musk started SpaceX}? \\
		11 & User & No. Tell me more about Jeff Bezos. \\
		12 & Bot 	& Sure. {\it Amazon founder Jeff Bezos, the world's richest man,
			was one of the first billionaires to jump into the commercial ``space
		race'', starting Blue Origin in 2000.} \\
		13 & User & What does Blue Origin do? \\
		14 & Bot & Let's see if the article tells us. Here is what I found. 
		{\it Gaining momentum, Blue Origin is now securing lucrative government
			contracts and recently gained national security certifications from the US
		government.} 
		Do you want to know \uline{what it was chosen by the US Air Force to
		develop?} \\
		15 & User & Sure! \\
		\hline\hline
	\end{tabular}
	\caption{A sample multi-turn socialbot conversation for discussing a news
	article.
	The article title and sentences are shown in {\it the italic font}. 
	Generated questions presented as in the introductory clause form are \uline{underlined}.
	The retrieved comment is indicated using \uwave{a wavy underline}.
}
	\label{chap5:tab:sample_conversation}
\end{table}

To illustrate the mixed-initiative dialog strategy,
Table\,\ref{chap5:tab:sample_conversation} shows a sample conversation about a
news article.\footnote{\url{https://www.bbc.com/news/business-45919650}}
The socialbot starts with the article title at turn 2.
It then carries out a discussion-chain move at turn 4, where
sentences in the discussion chain are presented.
The discussion-chain move is also used at turn 6.

The question-edge move is carried out at turn 4, turn 10, and turn 14.
These turn-taking questions using the generated questions presented in the
introductory clause form to provide information to the user for deciding whether
they would like to hear the sentence or not.
The use of the introductory clause can be viewed as following the maxim of
quantity from Grice's Maxims for conversational communication~\cite{Grice1975},
i.e., the socialbot tries to be as informative as possible at a turn but avoids
being more informative than is required.

The comment-edge move is illustrated from turn 6 to turn 9.
The socialbot first asks an opinion-seeking question at turn 6.
Then at turn 8, it reads a comment and asks whether the user agrees.
After the opinion-exchanging segment finishes, the socialbot attempts to return
to the discussion chain at turn 10 using a question-edge move.
The use of comments increases the diversity of the interaction and brings new
social chat content that is not included in the article itself.
Note that at turn 7, the user actually takes the initiative and asks an
opinion-seeking question as well.
In this case, the comment retrieval strategy turns out to agree with the
comment-edge move.

In the example, the user also takes the initiative at turn 11 and turn 13.
The entity-based retrieval is used at turn 12 for the entity ``Jeff Bezos''.
The question-based retrieval is used at turn 14.
In this example, the socialbot searches among question nodes that have a
question type ``what''.
The question node \textlangle what, Blue Origin, is securing\textrangle\
turns out to have the highest matching score for the user question.
Since the match is not always exact, the socialbot uses the sentence
``{\it let's see if the article tells us}'' to reflect its low confidence
level (e.g., the percentage of matched content words) before presenting the
retrieved sentence node.

Note that the socialbot can use multiple strategies in one turn.
At turn 4, the socialbot carries out a discussion-chain move and a
question-edge move.
At turn 6, the comment-edge move follows the discussion-chain
move.
At turn 14, both the question-based sentence retrieval and
question-edge move are used.

\section{Experiments and Analysis}
\label{chap5:sec:experiments}
In this section, experiments for discussion chain creation are carried out in
\S\ref{chap5:ssec:exp_sentence_selection}, where we describe features used in
the logistic regression model and evaluate their performances.
In \S\ref{chap5:ssec:exp_introductory_clause}, the quality of automatically
generated questions is evaluated using crowdsourced human judgments.
Lastly, we assess the proposed mixed-initiative dialog strategies through human
studies in \S\ref{chap5:ssec:exp_dialog_strategies}.

\subsection{Discussion chain experiments}
\label{chap5:ssec:exp_sentence_selection}

We treat the problem of discussion chain creation as a sequential sentence
selection problem: given the partial discussion chain $x_1, \dots, x_L$, choose
the next sentence for extending the chain from $N$ candidates \{$y_1, \dots,
y_N$\}.
Below, we describe features used in the logistic regression model for
characterizing candidate sentence chain $(x_1, \dots, x_L, y_n)$.
Two features build on the BERT model \cite{Devlin2018Bert}, which is a recent
new method for pre-training language representations, and it has achieved
state-of-the-art results on many tasks such as question answering and language
inferences.
\begin{itemize}
	\item {\bf SentenceDistance}: This feature is a single value that represents
		the distance of the candidate sentence from the ending context sentence
		$x_L$.
		The distance is calculated as $1 / (d + 1)$ where $d$ is the number of
		sentences in the article between the candidate sentence and the ending
		context sentence $x_L$.
	\item {\bf TextRank}: We use the TextRank
		algorithm \cite{Mihalcea2004EMNLP,Barrios2016arXiv} implemented in
		Gensim \cite{Gensim} to score individual sentences in the article.
		The algorithm is originally proposed for unsupervised extractive
		summarization by scoring sentences based on their centrality in the article.
		The score of the candidate sentence is used as the feature.
	\item {\bf BERTSingleSentence}:
		We use a pre-trained BERT model to derive embeddings
		for candidate sentences without using the context sentences.
		The embeddings are used as features that encode the textual information in
		the candidate sentence.
	\item {\bf BERTSentenceChain}:
		The pre-trained BERT model can also be used for deriving embeddings for
		sentence pairs.
		Here, we follow the setting for sentence pair classification tasks
		in \cite{Devlin2018Bert} with a slight adjustment.
		First, we treat the context sentences $(x_1, \dots, x_L)$ as a single
		sentence $z$.
		The BERT model can produce a classification embedding for the sequence
		(\sos, $z$, \sep, $y_n$),
		where \sos\ is the special sequence start token, and \sep\ is the special
		sequence separator token.
		The embedding for \sos\ is used as the classification embedding, since the
		BERT model uses it for the next sentence classification task during
		training.
		The classification embedding is used as the feature that encodes the
		textual information of the sentence chain and the relation between the
		candidate sentence and context sentences.
\end{itemize}

For features derived from BERT, we use the BERT-Large uncased English model
released in \url{https://github.com/google-research/bert}.
Note we do not jointly adapt the pre-trained BERT model when training the
logistic regression model.
The SentenceDistance and TextRank features are z-normalized based on the
training data.
Embeddings derived from BERT are not standardized.

{\bf Evaluation Metric}:
To evaluate the model performance for creating discussion chains,
we make the simplifying assumption that each decision contributes independently
to the overall score.
The annotated data described in \S\ref{chap5:ssec:model_discussion_chain} use
$N=4$ for $L=1$ and $N=3$ for $L=2$.
For each partial discussion chain $(x_1, x_2, \dots, x_L)$ in our data, there
may have 0, 1, or 2 candidate sentences that are labeled as positive for
extending the chain.
In our experiments, partial discussion chains with no candidate sentence labeled
as positive are kept for training the model, but excluded for the final
evaluation. 
Given $M$ distinct partial discussion chains $(x_1, \dots, x_L)$ in the data, we
evaluate the model by calculating accuracy, i.e., $m/M$ where $m$ is the number
of chains for which the top ranked sentence is labeled as positive.
The accuracy on two subsets ($L=1$ and $L=2$) are reported separately.

\begin{table}[!t]
	\centering
	\begin{tabular}{l|c|c|c|c|c}
		\hline\hline
		\multirow{2}{*}{\bf Feature} 
		& \multirow{2}{*}{$|\mathcal{F}|$}
		& \multicolumn{2}{c|}{Validation} 
		& \multicolumn{2}{c}{Test}  \\
		\cline{3-6}
		& & $L=1$ & $L=2$ & $L=1$ & $L=2$ \\
		\hline
		SentenceDistance 							& 1 	 & 45.1 & 59.3 & 54.7 & 62.3 \\
		TextRank 											& 1 	 & 55.0 & 57.5 & 63.2 & 71.9 \\
		BERTSingleSentence    				& 1024 & 53.9 & 64.6 & 62.1 & 69.3 \\
		BERTSentenceChain    					& 1024 & 58.2 & 74.3 & 64.8 & 73.7 \\
		BERTSentenceChain + TextRank 	& 1025 & 53.9 & 70.8 & {\bf 68.4} & {\bf 75.4} \\
		All 													
		& 2050 & {\bf 61.5} & {\bf 75.2} & 66.3 & 70.2 \\
		\hline\hline
	\end{tabular}
	\caption{Results of logistic regression models using different features for
	discussion chain creation. Accuracy values are scaled by 100.}
	\label{chap5:tab:exp_sentence_selection}
\end{table}

\begin{table}[!t]
	\centering
	\begin{tabular}{l|c|c|c|c}
		\hline\hline
		\multirow{2}{*}{\bf Feature} 
		& \multicolumn{2}{c|}{Validation} 
		& \multicolumn{2}{c}{Test}  \\
		\cline{2-5}
		& $L=1$ & $L=2$ & $L=1$ & $L=2$ \\
		\hline
		BERTSingleSentence 						
		& 60.4 (+6.5) & 62.8 (-1.8) & 59.0 (-3.1) & 68.4 (-0.9) \\
		BERTSentenceChain    					
		& 51.7 (-6.5) & 68.1 (-6.2) & 68.4 (+3.6) & 73.7 (+0.0) \\
		BERTSentenceChain + TextRank    					
		& 58.2 (+4.3) & 67.3 (-3.5) & 69.5 (+1.1) & 74.6 (-0.8) \\
		All 													
		& 61.5 (+0.0) & 69.9 (-5.3) & 65.3 (+0.0) & 70.2 (+0.0) \\
		\hline\hline
	\end{tabular}
	\caption{Results of logistic regression models trained and tuned on the joint
		set of $L=1$ and $L=2$. Accuracy values are scaled by 100. Numbers in
		brackets show the absolute difference with respect to corresponding values
		in Table\,\ref{chap5:tab:exp_sentence_selection}.}
	\label{chap5:tab:exp_chain_creation_joint_train}
\end{table}

{\bf Experiment Results}:
Table\,\ref{chap5:tab:exp_sentence_selection} summarizes the validation and test
sets results for logistic regression models using different features.
The corresponding feature dimensions $|\mathcal{F}|$ are also reported.
Note since SentenceDistance and TextRank are both single-value features, the
logistic regression model is equivalent to using the feature value for ranking
sentences.
First, it can be observed that the SentenceDistance performs the worst.
On the validation set, the accuracy values indicate that only in 45.1\% cases the
immediate next sentence in the article is good as a follow-up sentence for the
discussion chain when $L=1$,
since it is common that the first few sentences
in the article provides redundant information when the title is already given.
For $L=2$, the number is slightly higher, i.e., 59.3\%, but there are still
plenty of cases where the next sentence in the article is not a good candidate.
In the validation set, the frequency that at least one of the next two sentences in
the article is positive is 79.1\% for $L=1$ and 85.0\% for $L=2$.
The TextRank feature seems to be more reliable in testing.
The BERTSingleSentence vector has a high dimension, so it can capture features
of generally interesting sentences, but it lacks context so is less useful than
TextRank.
The BERTSentenceChain vector characterizes specific sentence context (vs.\
article context of TextRank), and it gives the best results of the four options
alone on the test data.
Using BERTSentenceChain and TextRank together performs best on the test set.
Merging all features (SentenceDistance, TextRank, BERTSingleSentence,
BERTSentenceChain) leads to the best performance on the validation set but not
on the test set, which might due to the overfitting since there are only 1239
training samples whereas the feature dimension is 2050.

Note that in Table\,\ref{chap5:tab:exp_sentence_selection}, the models for $L=1$
and $L=2$ are trained separately.
We can also train and tune the model on the joint set of $L=1$ and $L=2$.
Table\,\ref{chap5:tab:exp_chain_creation_joint_train} shows the results for this
setting.
While most observations comparing across features are consistent with
Table\,\ref{chap5:tab:exp_sentence_selection}, we notice that training on the
joint set hurts the performance in most cases.
For BERTSingleSentence, we hypothesize that the model may be confused since the
same candidate sentence can have opposite labels in $L=1$ and $L=2$.
For BERTSentenceChain, the performance degradation may be caused by the mismatch
in the length of the context sentence in the two subsets, which may result in
different dynamic range of feature values and confuse the model. 
It can be seen that the degradation is generally less when using all features
including SentenceDistance and TextRank which are not sensitive to the
difference between subsets.

\subsection{Question generation experiments}
\label{chap5:ssec:exp_introductory_clause}

\begin{figure}[!t]
	\centering
	\includegraphics[width=\textwidth,trim=1cm 1cm 1cm 10cm,clip]{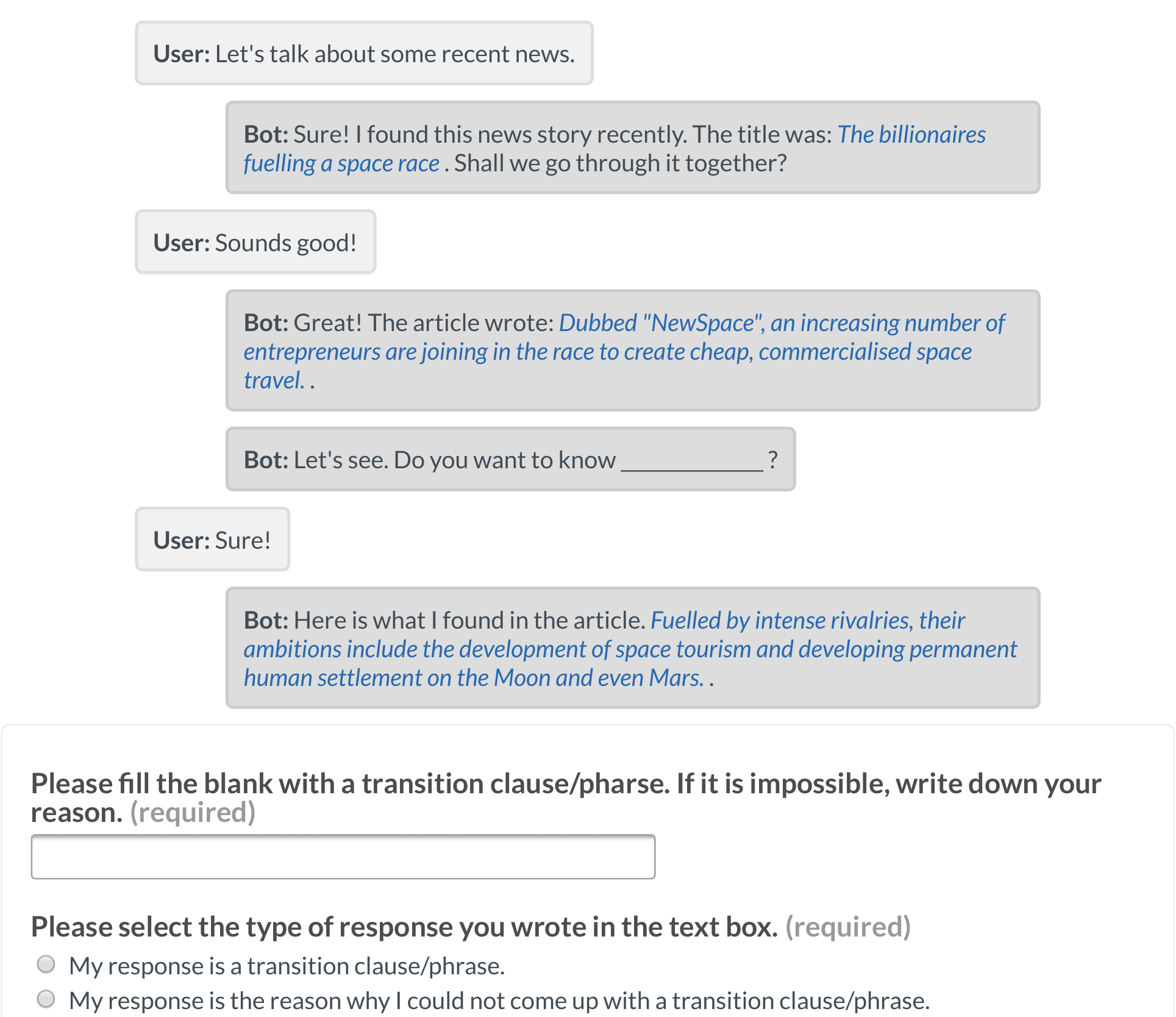}
	\caption{A sample unit for introductory clause collection.}
	\label{chap5:fig:iclause_collection}
\end{figure}

The experiments in this subsection compare several automatic question
generation methods.
Specifically, we are interested in using the generated question for introducing
the sentence in the discussion chain as described in
\S\ref{chap5:ssec:system_init_strategies}.
Thus, the following methods are compared.
\begin{itemize}
	\item {\bf Generic}: This method always uses a generic question, i.e.,  ``do
		you want to hear more about this article?''
	\item {\bf Constituency-Based}:
		In \cite{HeilmanPhDThesis}, an automatic question generation method is
		proposed for reading compression assessment based on constituency parsing,
		and uses a statistical ranker to select the best question.
		We include this method for comparison using their
		implementation.\footnote{\url{http://www.cs.cmu.edu/~ark/mheilman/questions}}
		Since the algorithm was not designed for introducing content, we render the
		question in the form of ``do you want to know \dots'' and adjust the
		auxiliary verb to match the grammar.
	\item {\bf Dependency-Based}: This is the method described in
		\S\ref{chap5:ssec:algo_iclause_generation} based on universal dependencies.
		Questions are selected based on a manually crafted decision tree as
		described in \S\ref{chap5:ssec:system_init_strategies}.
	\item {\bf Human-Written}: Human-written transition clauses are crowdsourced
		from the FigureEight platform using the interface shown in
		Fig.\,\ref{chap5:fig:iclause_collection}.
		We pay 4 cents for each unit.
		A simple validation method is implemented to avoid the worker from
		submitting generic and very short clauses (e.g., ``more'', ``more
		information''), but we allow workers to write down the reason if they cannot
		come up with an appropriate transition clause.
		Multiple clauses are collected for each unit.
		Note the validation method during the data collection cannot eliminate all
		low-quality clauses.
		After manual inspection, we find that clauses that do not start with an
		interrogative word mostly have low quality.
		Thus, in these experiments, we only use human-written clauses that start
		with one of the following words: how, if, what, when, when, whether, which,
		which, who, why.
		When multiple clauses are kept for a sentence, a random one is chosen as
		the human-written reference.
\end{itemize}

To compare different methods, we crowdsource human judgments on the FigureEight
platform to carry out A/B tests.
Specifically, workers are asked to read an incomplete subdialog such as the
example in Fig.\,\ref{chap5:fig:iclause_collection} and rank two clauses
generated from different methods in terms of three aspects, i.e., grammar,
informativeness, and smoothness.
The worker is also asked to select the better clause or ``cannot tell'' with
respect to each of these aspect.
The three aspects are explained in the annotation instruction as described
below.
\begin{itemize}
	\item The clause should be grammatically correct, but please ignore the
		errors in capitalization.
	\item The clause should be informative.
		Ideally, it should provide some information in the next article sentence to
		engage the user.
	\item The clause should provide a smooth transition for the conversation.
		It should connect the previous bot turn and the next bot turn appropriately.
\end{itemize}
The worker is also asked to select the better clause or ``cannot tell'' based on
the overall quality.
We carry out the following four sets of A/B tests, i.e.,
generic vs.\ constituency-based, generic vs.\ dependency-based, 
constituency-based vs.\ human-written,
and dependency-based vs.\ human-written.
For all A/B tests, the same set of 134 sentences are used.
We collect 5 judgments per annotation unit and pay 3 cents per judgments.
The order of presenting A and B in the unit is randomized to avoid
potential bias.

\begin{figure}[t]
	\centering
	\includegraphics[width=0.9\textwidth]{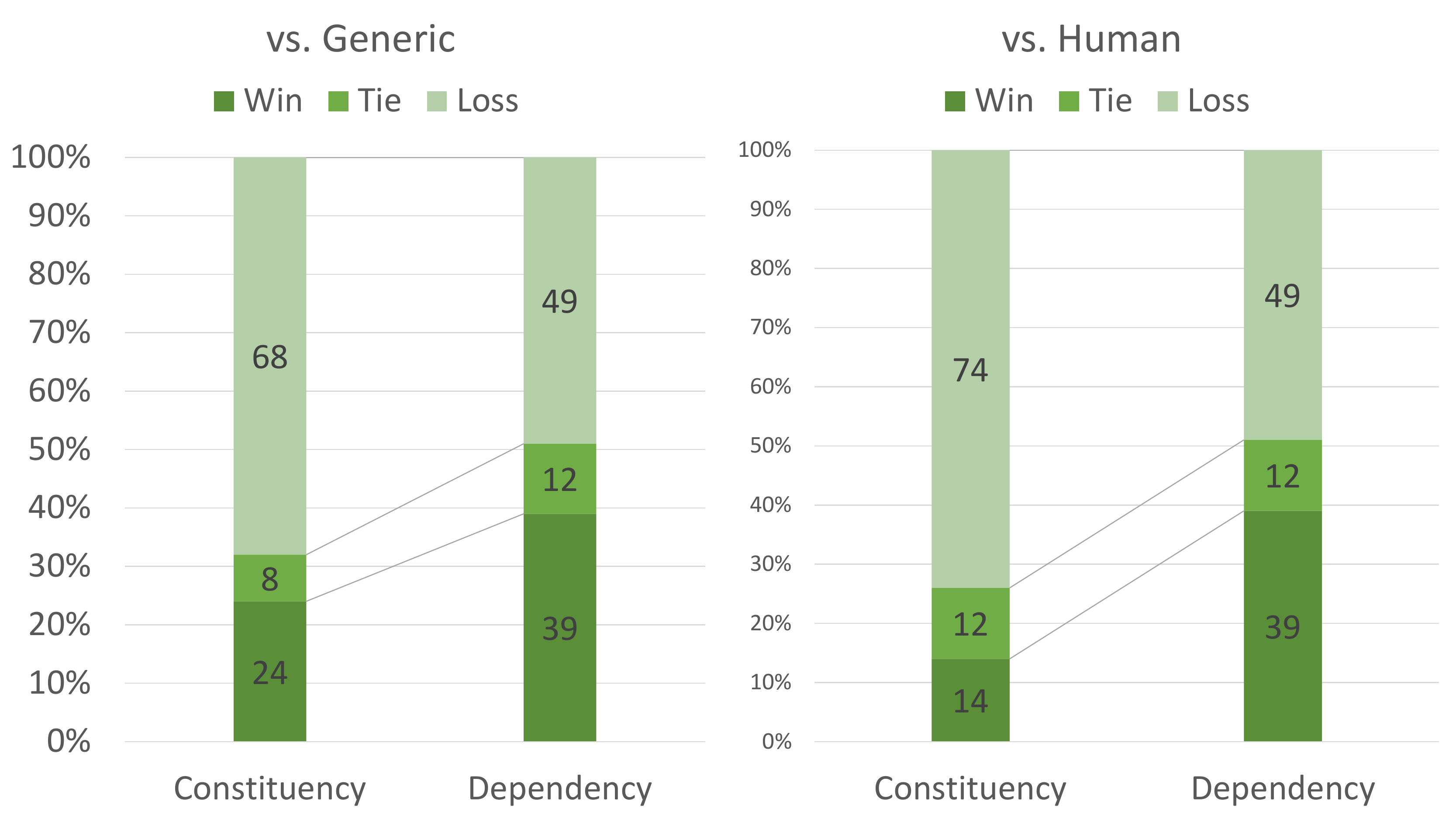}
	\caption{The win-loss-tie records of A/B tests on generated questions in terms
	of grammar.}
	\label{chap5:fig:qgen_grammar}
\end{figure}

\begin{figure}[t]
	\centering
	\includegraphics[width=0.9\textwidth]{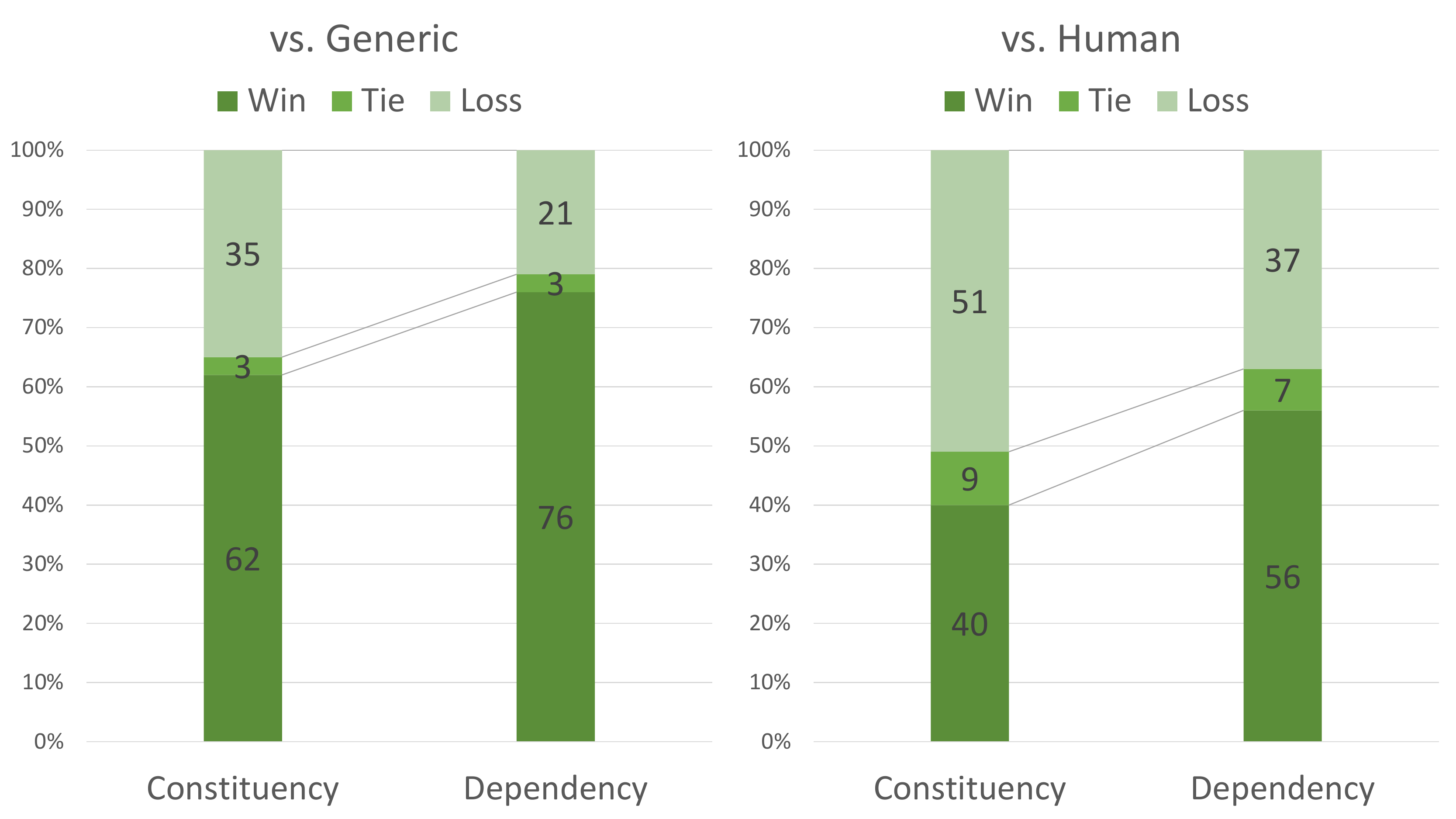}
	\caption{The win-loss-tie records of A/B tests on generated questions in terms
	of informativeness.}
	\label{chap5:fig:qgen_informativeness}
\end{figure}

\begin{figure}[t]
	\centering
	\includegraphics[width=0.9\textwidth]{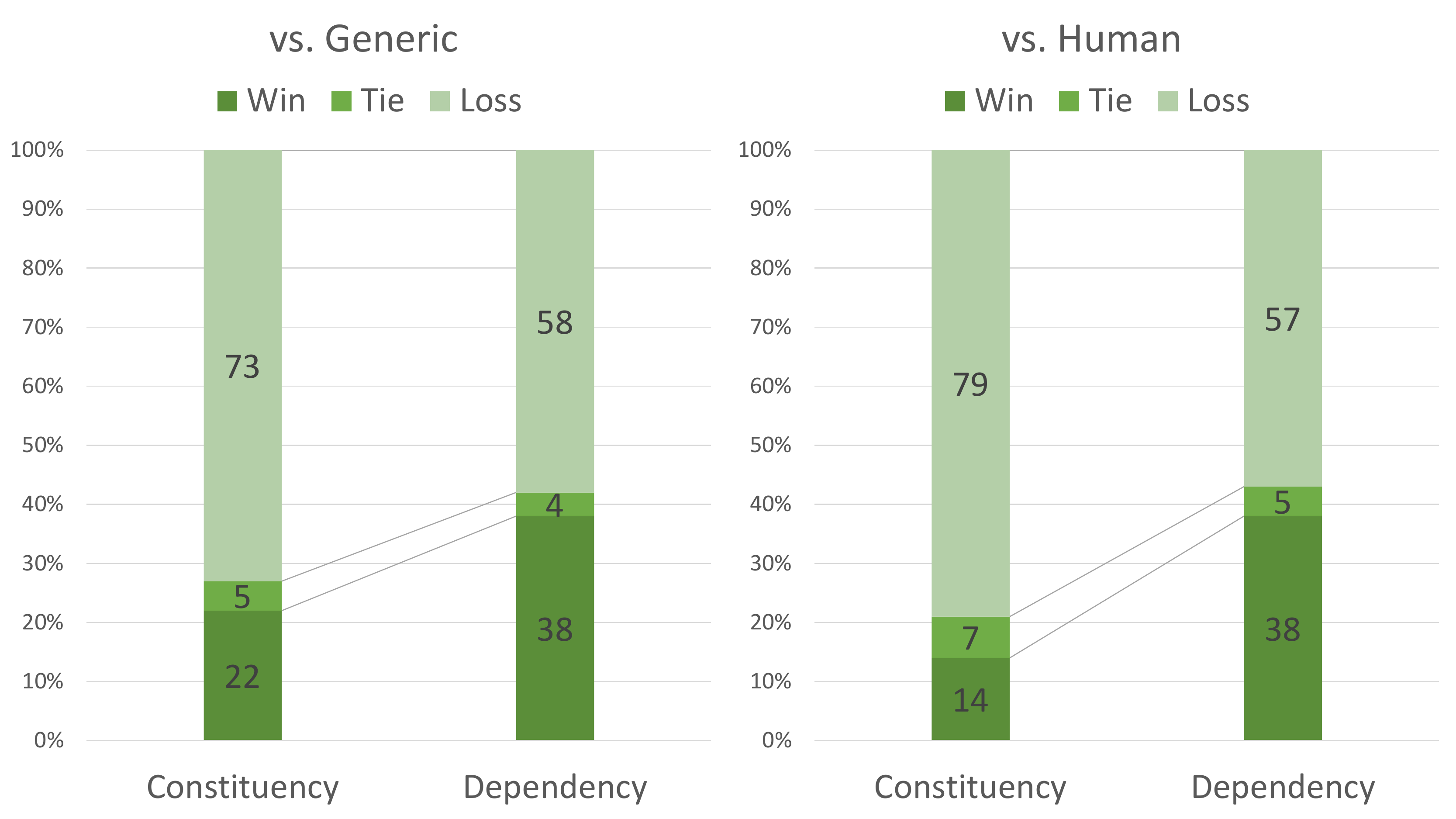}
	\caption{The win-loss-tie records of A/B tests on generated questions in terms
	of transition smoothness.}
	\label{chap5:fig:qgen_smoothness}
\end{figure}

\begin{figure}[t]
	\centering
	\includegraphics[width=0.9\textwidth]{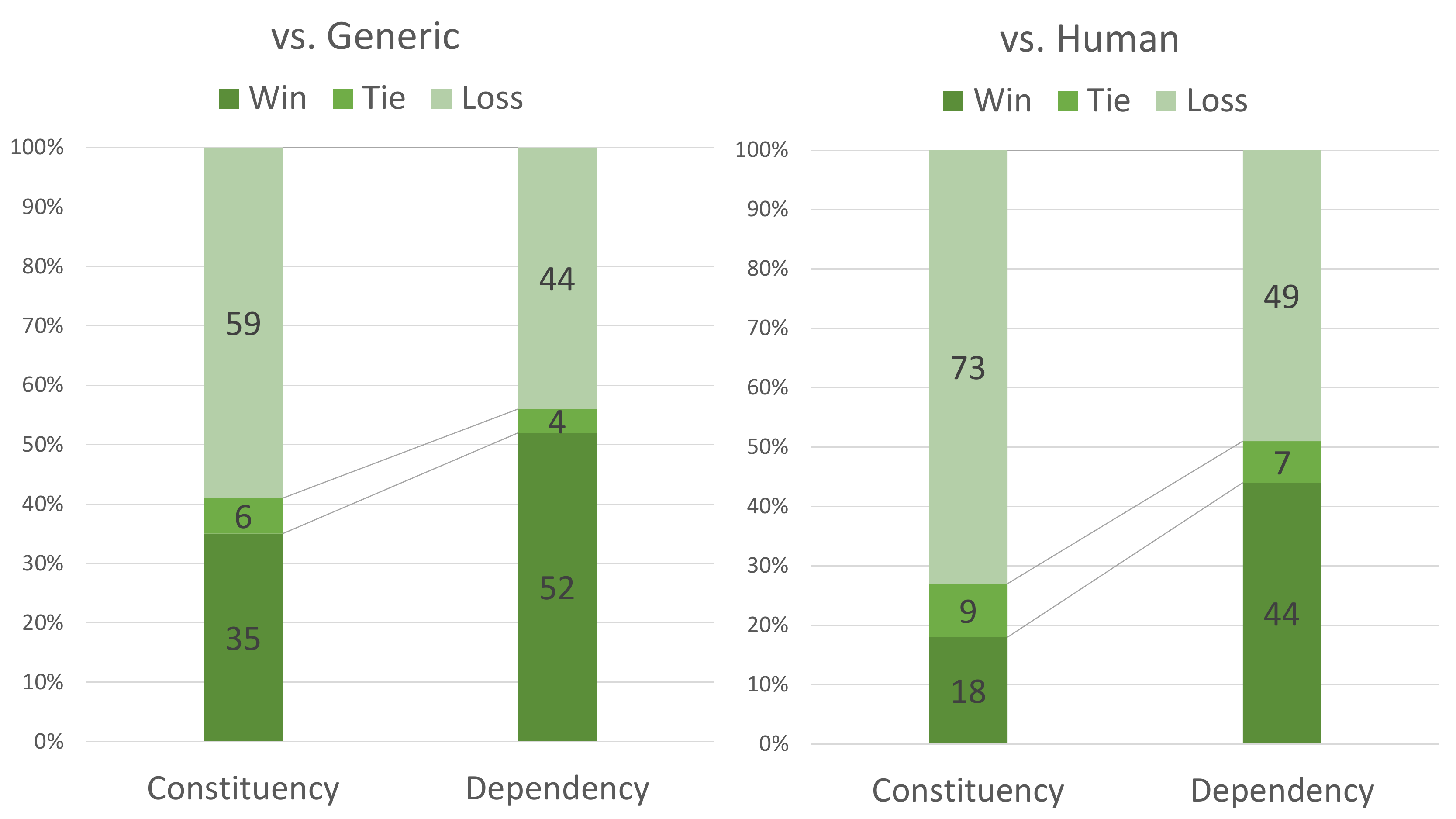}
	\caption{The win-loss-tie records of A/B tests on generated questions in terms
	of overall quality.}
	\label{chap5:fig:qgen_overall}
\end{figure}

Evaluation results are summarized in
Figs.\,\ref{chap5:fig:qgen_grammar}--\ref{chap5:fig:qgen_overall}.
It can be seen that the dependency-based method consistently outperforms the
constituency-based method.
In terms of grammar, the method using constituency parsing is not as good as the
generic or human-written clauses, whereas our dependency-based method tends to
generate grammatically correct clauses.
One would expect that the percentage would be close to 50\% for an A/B test comparing
two human-written clauses.
In terms of informativeness, it can be observed that both
constituency-based and dependency-based methods outperform the generic baseline,
and in 63\% of judgments the dependency-based method is selected as better or
equivalent to human-written clauses.
However, in terms of smoothness, none of the methods outperforms
either the generic baseline or human-written clauses.
This is somewhat expected since both automatic question generation methods do
not take into account the dialog context.
The generic baseline is always an appropriate transition clause although it is
not very informative.
Human workers are mostly good at making good connections between the context and
the next sentence.
Lastly, in terms of the overall quality, we can see that the proposed
dependency-based method is still as good as human-written clauses and slightly
better than the generic baseline.

\subsection{User interaction experiments}
\label{chap5:ssec:exp_dialog_strategies}

To illustrate the usability of the proposed document representation and
mixed-initiative dialog strategies, we implement a socialbot prototype focused
on news article conversations and carry out user studies.
The socialbot is implemented using a similar system architecture as developed in
Chapter~\ref{chap3}.
Specifically, a new miniskill for news article discussion is added to a revised
version of Sounding Board system.
The revisions include a new sentence segmentation model using a bi-directional
recurrent neural network with attention for restoring punctuations
\cite{Tilk2016Interspeech} and a new language understanding model trained using
the RASA library \cite{RASA} on manually annotated user utterances collected
during the Alexa Prize competition.
Since this work focuses on the news article discussion, we temporarily disable
all other miniskills and directly initiate the news article discussion at the
beginning of the conversation.
The conversation ends immediately after the bot finishes the discussion on an
article.

In our controlled studies, recruited users interact with the socialbot
using an Echo device.
Each user interacts with the socialbot for multiple times, and each time the
socialbot uses a different news article and dialog strategies.
The following two dialog configurations are compared.
\begin{itemize}
	\item {\bf Full}: This configuration uses the proposed mixed-initiative dialog
		strategy described in \S\ref{chap5:sec:dialog_design}.
	\item {\bf Baseline}:
		In this configuration, collected comments for the news article are not used.
		Additionally, the configuration does not use the generated questions for
		introducing the next sentence.
		Instead, it uses a generic introductory question, ``{\it do you want
		to know more about this article}''.
		Note the generated questions are still used for question answering in this
		configuration.
\end{itemize}

There are 9 persons involved in the study.
They are divided into three groups, each of which has 3 persons.
Each person in the group takes turns to play the role of interactor and judges.
The user who has interacted with the bot multiple times is asked to rank their
interactions.
While a user is interacting with the bot, the two judges listen to the
interaction and rate the bot performance in the conversation.
From this study, we find that the judges give a higher rating to the full
configuration over the baseline configuration for 9 out of 14 cases.
Interestingly, for interactor-reported rankings, the full configuration only
wins over the baseline configuration 3 out 8 cases, and is tied in 1 case.
Based on our observations, it seems that errors in speech recognition, language
understanding, and failures of finding correct answers for user questions all
impact the interactor ratings.
As was the case with Alexa Prize users, interactors behave quite differently in
terms of their willingness to follow the conversation flow and the frequency of
asking questions.
Lastly, for 8 pairs of conversations with the full and baseline configurations
where each pair is carried out by the same interactor, we find that the full
configuration results in a user vocabulary of 98 distinct words, whereas there
are only 61 distinct words for the baseline configuration.
It seems that the full configuration tends to elicit more diverse user
utterances, suggesting that the user is probably more actively participating in
the conversation.

\chapter{Conclusion}
\label{chap6}
To conclude, we first summarize the work carried out in this thesis and its
contributions in \S\ref{chap6:sec:summary}.
Impacts of this thesis on socialbots and task-oriented systems are discussed in
\S\ref{chap6:sec:impacts}.
Future directions for work on conversational artificial intelligence (AI) are
discussed in \S\ref{chap6:sec:future_work}.

\section{Summary}
\label{chap6:sec:summary}
The focus of this thesis is the socialbot.
While there is a long history of conversational AI research, the socialbot is a
new type born out of the Alexa Prize and has attracted increasing research
interest.
Our work is one of the pioneers in this new area.
This thesis has presented the Sounding Board system in Chapter \ref{chap3} that
won the inaugural Alexa Prize in 2017.
Chapter \ref{chap4} has approached the socialbot evaluation and diagnosis problem
using a large amount of data collected from Sounding Board during the
competition.
In Chapter \ref{chap5}, we have developed new methods for socialbot
conversations grounded on an unstructured document, specifically focused on news
articles which are constantly changing and demanding for socialbots.
More details of the contributions of this thesis are discussed below.

{\bf Sounding Board System}:
To build a mixed-initiative and open-domain socialbot, we have developed
user-centric and content-driven dialog strategies. 
Following these two objectives, several miniskills have been designed for
Sounding Board.
The system uses an architecture that consists of a multi-dimensional language
understanding module,
a novel hierarchical dialog management (DM) module,
a state-independent language generation module,
and a new social chat knowledge graph.
In particular, the hierarchical DM framework decouples the dialog context
tracking and complex dialog control into individual modules for efficient system
development and maintenance.
The framework can be useful for all kinds of dialog systems that need to be
capable of accommodating a variety of domains.
Within this framework, the DM module of Sounding Board implements a dialog
context tracker, a master dialog manager, and several miniskill dialog managers.  
Furthermore, we have designed and compiled a large scale dynamic knowledge base
that contains social chat content collected from multiple sources.
Specifically, we use a graph structure to organize the knowledge base which
allows dialog strategies to be developed as moves on the graph.
In this way, originally isolated content from different knowledge sources can
be used in the conversation coherently.
In this work, we have formulated the dialog flow of a miniskill as a finite
state machine and have manually specified the transitions between states due to 
the fact that the socialbot research is in an early stage that makes
it difficult to create a corpus for learning a statistical dialog manager.
Other challenges to use of a statistical DM are that there is no consensus on
the inventory of state, and new capabilities are frequently added to a
socialbot that interacts with real users.
The proposed hierarchical DM framework alleviates some of these issues as it
allows the system to upgrade a miniskill dialog manager to a statistical version
with minimal impacts on other components of the socialbot.

{\bf Socialbot Conversation Acts and User Rating Analysis}:
We have carried out in-depth analysis on the conversation data and user
ratings collected during the competition.
It is known from prior work on socialbot conversation evaluation
that the conversation length (i.e., the number of turns in total) is positively
correlated with conversation-level user ratings, but the correlation is not very
high and no other automatic metrics have been reported to have a higher
correlation with conversation-level user ratings.
In our study, while we have observed a similar trend that most metrics turn out
to be weakly correlated with user ratings, we still have found several metrics
that have significantly higher correlation than the conversation length.
Using conversation acts we developed for characterizing user and bot turns in
socialbot conversations, our correlation studies have provided valuable new
insights for socialbot evaluation.
Specifically, we have shown that all metrics based on the number of turns for a
specific conversation act are positively correlated with user ratings; only
metrics based on the percentage of turns can reflect negative correlation with
user ratings, which can help us to identify system issues.
We identified both positive and negative conversation acts based on their
correlations with user ratings.

{\bf Multi-Level Evaluation and Socialbot Diagnosis}:
To further pin down system issues for improving the bot, we investigate
approaches to evaluate socialbots at segment levels.
In particular, we explore two multi-level scoring models that are designed based
on two different model assumptions about the user experience (linear vs.\
non-linear contribution of dialog segment experience to the conversation-level
user rating) and a hierarchical strategy for conversation segmentation.
We have compared features based on the proposed conversation acts and previously
proposed features in terms of predicting the conversation score.
Experiments show that user reactions, intents, and language use contain the most
useful information for predicting the final conversation-level user
ratings.
Further, we have shown that the subdialog scores estimated by these two
approaches have much higher correlations with human judges compared to two
baseline methods that use the conversation length and conversation-level user
ratings, respectively.
Lastly, we have provided three examples showing how segment-level scores can be
used to diagnose system performance, i.e., identifying topics that are not handled
well by the bot, comparing topic initiation strategies, and assessing the
quality of content sources.

{\bf Graph-Based Document Representation}:
We have developed methods for socialbots to carry out extended conversations
grounded on a document.
The approach extends the system knowledge representation by developing a
document representation that structures the document as a graph consisting of
sentence, topic, question, and comment nodes.
The proposed graph-structured document representation brings together machine
reading and dialog control techniques, which are two key challenges for enabling
document-grounded conversations.
First, annotations have been collected to build models for automatically
creating the follow-up edges between sentence nodes that form the discussion
chain for a new article.
Using the collected annotations, experiments have been carried out to compare
several features in a model for creating the discussion chain, which show that
the dialog context and document context are both important for the model.
Further, an unsupervised method using universal dependencies has been developed
to automatically generate questions.
These questions are used for both retrieving sentences for user questions and
introducing the next sentence in the conversation.
The generated questions are assessed by crowdsourced workers in terms of
grammar, informativeness, transition smoothness, and the overall quality.
The crowdsourced judgments indicate that the proposed dependency-based method
can generate grammatically correct questions. 
Compared with human-written questions, the generated questions are more
informative, but less coherent in terms providing a smooth transition from the
previous bot turn to the next bot turn in the conversation.
We have also designed a comment crowdsourcing method to augment the document
representation.
Lastly, using the proposed document representation, we have developed a
mixed-initiative dialog strategy for the socialbot to discuss a news article.
A socialbot prototype has been implemented to carry out user studies and assess
the proposed dialog strategy.
The studies show that the proposed dialog strategy is preferred by
judges over a baseline strategy that does not use comments for discussion and
generated
questions for introducing the next sentence.
In addition, the proposed dialog strategy elicit more diverse user utterances,
suggesting the user is more actively participating in the conversation.

\section{Impacts}
\label{chap6:sec:impacts}
This thesis impacts future research on socialbots in several ways.

First, in order to make progress on this area, it is necessary to have a running
system that works reasonably well for the purpose of collecting interaction data
and studying the conversations.
Among the first wave of socialbots competed in 2017 Alexa Prize, Sounding
Board's results led the field
and the original system continued to be competing with the 2018 socialbots.
The system opens up the possibility to study socialbot conversations at scale
such that further progress can be made on this research area, including system
design, system evaluation, dialog management, content management, etc.
The user-centric and content-driven dialog strategy and the hierarchical DM
framework have already influenced several 2018 Alexa Prize socialbots (e.g.,
\cite{UCDavisGunrock2018,CMUTartan2018}).

Second, our user rating analysis and multi-level evaluation approaches provide
valuable insights for socialbot conversations, which can facilitate future
system development and diagnosis.
The conversation acts can be used to augment the language understanding
capability.
Segment-level evaluation helps assess dialog strategies in localized regions of
a conversation which will reduce the number of user conversations needed to
determine impact.
The proposed multi-level scoring models are also useful for learning the dialog
policy for statistical DM.
In particular, in the reinforcement learning setting, reward signals at the end
of the conversation are usually weak and delayed, leading to the issue of slow
learning.
The multi-level scoring methods can be used for reward shaping, which is a
remedy to address this issue by introducing intermediate reward signals to
complement the original sparse reward signal.

Third, the use of graph structure for organizing knowledge can provide a
universal framework for socialbots that bridge machine reading and dialog
control.
It facilitates the combination of techniques in natural language processing and
dialog management.
This idea was initially explored in Sounding Board where we create the knowledge
graph based on shallow understanding of documents and existing structured
knowledge base for movies.
The idea has also influenced several 2018 Alexa Prize socialbots (e.g.,
\cite{UCSCSlugbot2018,HWUAlana2018}).
In Chapter \ref{chap5}, we further address the challenge of discussing
unstructured documents and develop a graph-based document representation to
extend the social chat knowledge graph.
This extension can greatly improve a socialbot's ability to discuss the latest
events.
Furthermore, our studies have formulated several new problems that emphasize
both dialog context and document context, which can trigger interest in both
the natural language processing community and the dialog system community.

{\bf Impacts on Task-Oriented Dialog Systems}:
While this thesis has focused on socialbots, many ideas and findings can be
applied to task-oriented dialog systems.
In particular, there is growing interests in building open-domain and
mixed-initiative task-oriented dialog systems, which inevitably involve complex
dialog control.
The hierarchical DM framework can be very useful for these systems by treating
each task as a miniskill.
Furthermore, the large-scale knowledge base required by the open-domain scenario
makes dialog control very challenging.
The idea of using graph structure to organize the knowledge base and designing
dialog strategies based on moves on the graph can address the challenge.
In addition, in Chapter \ref{chap4}, the hierarchical dialog structure model is
developed primarily based on topical segments.
For task-oriented systems which involve multiple subtasks, a similar dialog
structure model can be developed by treating a subtask as a topic.
It is also straightforward to apply the proposed multi-level evaluation
approaches to such task-oriented systems, which helps to identify tasks that are
not handled successfully by the system.
Lastly, sometimes it is useful and necessary for a dialog system to have both
social chat and task-oriented conversations.
Such hybrid systems can be built upon progress on socialbots made in this
thesis and other socialbot research.

\section{Future Directions} 
\label{chap6:sec:future_work}
Moving forward, open-domain and mixed-initiative are the future trend of
conversational AI systems.
Being open-domain requires the system to make use of a large-scale knowledge
backend.
Being mixed-initiative requires the system to play two roles in a
conversation: the role of a primary speaker who leads the conversation flow
while allowing the user to ask questions and request specific information, and
the role of an active listener who does not only signal listenership to the user
but also contributes to the conversation by asking relevant questions and
bringing in interesting information.
Research efforts on socialbots have made an important step in advancing
conversational AI along this direction.
The work in this thesis is at the frontier of socialbot research.
There is still a long way to go to create a fully fledged conversational AI.
In the remaining part, we discuss several future directions for socialbots and
conversational AI.

Conversational AI is a multi-discipline area which can involve speech
processing, natural language processing, information extraction, data mining,
machine learning, conversation analysis, etc.
The research in conversational AI has benefited a lot from the progress made in
these related areas.
Since each area has its own focus, it is useful to find a way to
facilitate technology transfer from these areas to conversational AI research.
Our work on organizing social chat content and unstructured document using a
graph-based representation can be viewed as an attempt to bridge machine
reading and dialog control.
We expect to see more work along this line as research communities work closer
with each other.

A lot of research problems have been approached using data-driven methods, which
have led to promising results, specifically with the recent surge of interest
in neural network models.
However, to develop data-driven methods for socialbots and other conversational
AI systems, human-bot conversation data needs to be collected, which requires a
working system in place.
This becomes a chicken-and-egg problem.
Efforts have been made to address this problem, mostly using the
human-in-the-loop.
Notably the Wizard-of-Oz setup has been used in data collection for
several task-oriented systems.
Another line of recent work uses crowd-powered conversational AI
and some Alexa Prize teams have explored this idea in their socialbots.
Both attempts can be seen as streamlining the process of data collection, which
will likely become a common strategy for conversational AI in the future.
Our approach in developing methods for building the document representation in
Chapter~\ref{chap5} with carefully designed data collection interface can 
be viewed as one example of this direction.

User modeling and personalization are active research topics for conversational
system.
For socialbots, user characteristics have been shown to correlate with user
ratings and topic preferences.
Sounding Board has used personalized topic suggestion based on self-reported
personality information obtained from the Ask Personality Questions miniskill
described in Chapter~\ref{chap3}.
Other socialbots have also explored personalized topic recommendation using
learned models.
We expect to see more work on user modeling and personalization in both
socialbots and other types of conversational AI systems.

The importance of interesting and suitable content for socialbots has been
emphasized in our content-driven design and also acknowledged by several other
Alexa Prize teams.
Social media platforms have been used for content collection. 
However, they inevitably contain useless, boring, inappropriate, and sometimes
incorrect information.
A family-friendly socialbot should avoid mentioning content about sex, violence,
racism, hate speech, etc.
Therefore, we need to be extremely careful about the social appropriateness of
the content.
It is useful to develop methods that automatically identify inappropriate
content. 
This issue has attracted a lot of interest recently, and there is a rapidly
growing body of work on detecting hate speech and offensive content.
However, the broader question of identifying socially appropriate content is not
yet addressed.

Language understanding is one of the core components of conversational systems.
For conversational speech in socialbots, there are numerous
difficulties in developing a robust language understanding component to handle
the wide variety of user utterances in the open-domain setting.
It becomes more challenging when taking into account issues of speech
recognition errors and unsegmented speech transcripts.
Further, ellipsis and anaphora resolution are frequently observed in
socialbot conversations, which calls for context-dependent language
understanding methods such as the two-stage language understanding used in
Sounding Board.
Another difficulty we experienced is with the use of model-based language
understanding methods.
While model-based methods are promising and dominant in academic research, in
practice, we found it valuable to augment the model with extensive use of
rule-based methods such as regular expressions.  
Ironically, compared to adding regular expression rules to the system, it is
relatively time consuming to change a language understanding model, which involves
collecting data, annotating data, training and tuning the model, and deploying
the model.
A hybrid approach as we have been using in Sounding Board is a middle ground.
It is important to find a good balance point for combining both model-based and
rule-based approaches.

To engage users in the conversation and provide good user experiences, it is
important for the bot to actively signal attention and acknowledge the user.
Sounding Board benefited from extensive use of grounding acts to acknowledge
user reactions, which has since been emphasized by other Alexa Prize teams.
To make the bot to do so, we spent a considerable amount of time 
creating responses for different types of user utterances under different dialog
contexts.
On the other hand, the recent neural response models seem to be able to learn
simple response patterns from large-scale conversation data.
Although the attempt of using neural response models to produce complete
responses for socialbots has not achieved satisfactory results, it may be
possible to use them for signaling attention and acknowledging users, 
training on utterance pairs extracted from existing conversation datasets.

Lastly, it has been known that voice-based interactions are different from
text-based interactions.
For example, voice-based systems need to take into account issues of speech
recognition and synthesis.
During the Alexa Prize competition, we spent a considerable amount of time
trying to handle common speech recognition errors and adjust the prosody of bot
utterances, which are helpful as observed in other socialbots as well.
Working with speech requires more effort to crowdsource data and annotation for
voice-based systems, whereas it may be easier for mechanical turkers to type
responses compared to using their voice. 
On the other hand, voice-based and text-based systems can share many common
components.
Thus, we can imagine bootstrapping a voice-based system using a text-based system
with the help of some adjustment components, e.g., a component for automatically
adapting the prosody of textual bot responses.
This is an interesting future direction for conversational AI and it can
potentially accelerate the advancement in voice-based systems.

To sum up, the future directions discussed above are mostly based on experiences
in the journey of my PhD study.
They are only part of the much broader conversational AI research and we will
see increasing efforts devoted to this exciting area.

\newpage
\bibliographystyle{unsrt}
\bibliography{abrv_long,socialbot_ref,alexaprize_ref,dependency_parsing}

\end{document}